\documentclass[10pt,reqno]{amsart}
\usepackage{amsaddr}
\usepackage[a4paper, total={6.5in, 8.4in}]{geometry}

\usepackage[symbol]{footmisc}

\usepackage{amsmath}
\usepackage{amssymb}
\usepackage{amsthm}
\usepackage[latin1]{inputenc}
\usepackage{eurosym}
\usepackage{graphicx}
\usepackage{epsfig}
\usepackage{hyperref}
\usepackage{dsfont}
\usepackage[space]{cite}
\usepackage{subcaption}
\usepackage{caption}
\usepackage{placeins}
\usepackage{mathtools}
\usepackage{xcolor}
\usepackage{tcolorbox}
\usepackage{comment}
\usepackage{booktabs} 

\usepackage[ruled,vlined]{algorithm2e}
\usepackage[title]{appendix}

\usepackage{amsmath,amssymb,amsfonts}
\usepackage{tikz}

\usepackage[mathlines]{lineno}
\modulolinenumbers[1]

\allowdisplaybreaks
\usepackage{url}
\usepackage{hyperref}

\usepackage{ifthen}
\usepackage{comment}

\makeindex
\usepackage{color}
\usepackage{float}
\usepackage{commath}

\DeclareMathOperator*{\argmin}{arg\,min}

\def\qfa{\quad\text{for all}\quad}

\def\qa{\quad\text{and}\quad}
\def\qf{\quad\text{for}\quad}
\def\qw{\quad\text{where}\quad}

\def\qif{\quad\text{if}\quad}

\def\geq{\geqslant}

\numberwithin{equation}{section}

\newtheoremstyle{thmlemcorr}{10pt}{10pt}{\itshape}{}{\bfseries}{.}{10pt}{{\thmname{#1}\thmnumber{
			#2}\thmnote{ (#3)}}}
\newtheoremstyle{thmlemcorr*}{10pt}{10pt}{\itshape}{}{\bfseries}{.}\newline{{\thmname{#1}\thmnumber{
			#2}\thmnote{ (#3)}}}
\newtheoremstyle{defi}{10pt}{10pt}{\itshape}{}{\bfseries}{.}{10pt}{{\thmname{#1}\thmnumber{
			#2}\thmnote{ (#3)}}}
\newtheoremstyle{remexample}{10pt}{10pt}{}{}{\bfseries}{.}{10pt}{{\thmname{#1}\thmnumber{
			#2}\thmnote{ (#3)}}}
\newtheoremstyle{ass}{10pt}{10pt}{}{}{\bfseries}{.}{10pt}{{\thmname{#1}\thmnumber{
			A#2}\thmnote{ (#3)}}}

\theoremstyle{thmlemcorr}
\newtheorem{theorem}{Theorem}
\numberwithin{theorem}{section}

\theoremstyle{thmlemcorr*}
\newtheorem{theorem*}{Theorem}
\newtheorem{lemma*}[theorem]{Lemma}
\newtheorem{corollary*}[theorem]{Corollary}
\newtheorem{proposition*}[theorem]{Proposition}
\newtheorem{problem*}[theorem]{Problem}
\newtheorem{conjecture*}[theorem]{Conjecture}

\theoremstyle{defi}

\theoremstyle{remexample}
\newtheorem{remark}[theorem]{Remark}

\newcommand{\xqed}[1]{%
        \leavevmode\unskip\penalty9999 \hbox{}\nobreak\hfill
        \quad\hbox{\ensuremath{#1}}}
\newcommand{\Endrmk}{\xqed{\lozenge}} 
\newenvironment{RemarkQED}[1][]%
  {\begin{remark}[#1]\pushQED{\Endrmk}}%
  {\popQED\end{remark}}

\theoremstyle{ass}

\newtheorem*{notations*}{Notations}

\allowdisplaybreaks

\title[Bilevel Optimization for Learning Hyperparameters]{Bilevel Optimization for Learning Hyperparameters: Application to Solving PDEs and Inverse Problems with Gaussian Processes}
\author{\small Nicholas H. Nelsen$^{1,2}$, Houman Owhadi$^2$, Andrew M. Stuart$^2$,\\
Xianjin Yang$^{2,*}$, Zongren Zou$^{2, *}$}

\address{$^1$Department of Mathematics, Cornell University, Ithaca, NY 14853, USA.}

\address{$^2$Department of Computing and Mathematical Sciences, California Institute of Technology,\\ Pasadena, CA 91125, USA.}

\address{$^*$Corresponding authors: yxjmath@caltech.edu, zzou@caltech.edu.}
\email{\{nnelsen,owhadi,astuart,yxjmath,zzou\}@caltech.edu}

\ifpdf
\hypersetup{
	pdftitle={Bilevel Optimization for Learning Hyperparameters: Application to Solving PDEs and Inverse Problems with Gaussian Processes},
	pdfauthor={N. H. Nelsen, H. Owhadi, A. M. Stuart, X. Yang, and Z. Zou}
}
\fi

\begin{document}
\begin{abstract}
Methods for solving scientific computing and inference problems, such as kernel- and neural network-based approaches for partial differential equations (PDEs), inverse problems, and supervised learning tasks, depend crucially on the
choice of hyperparameters. Specifically, the efficacy of such methods, and in
particular their accuracy, stability, and generalization properties, strongly depends on the choice of hyperparameters. While bilevel optimization offers a principled framework for hyperparameter tuning, its nested optimization structure can be computationally demanding, especially in PDE-constrained contexts.
In this paper, we propose an efficient strategy for hyperparameter optimization within the bilevel framework by employing a Gauss-Newton linearization of the inner optimization step. Our approach provides closed-form updates, eliminating the need for repeated costly PDE solves. As a result, each iteration of the outer loop reduces to a single linearized PDE solve, followed by explicit gradient-based hyperparameter updates. We demonstrate the effectiveness of the proposed method through Gaussian process models applied to nonlinear PDEs and to PDE inverse problems. Extensive numerical experiments highlight substantial improvements in accuracy and robustness compared to conventional random hyperparameter initialization. In particular, experiments with additive kernels and neural network-parameterized deep kernels demonstrate the method's scalability and effectiveness for high-dimensional hyperparameter optimization.
\end{abstract}

\maketitle

\section{Introduction}
\label{sec:intro}

Many problems in scientific computing, such as the solution of PDEs \cite{chen2021solving,chen2025sparse,meng2023sparse,raissi2017machine,mora2025gaussian, zou2025learning}, inverse problems~\cite{engl2015regularization,kaipio2005statistical,stuart2010inverse,tarantola2005inverse}, operator learning \cite{li2020fourier, kovachki2023neural, lu2021learning, batlle2024kernel, nelsen2024operator, zou2024neuraluq}, and data-driven ODE or PDE discovery \cite{brunton2016discovering,rudy2017data,jalalian2025data, zou2024correcting}, can be cast as supervised learning. These models are found by enforcing training constraints drawn from data or physics, for example PDE residuals, and by searching for the best hypothesis within a fixed space under a chosen learning algorithm. Hyperparameters define the function space of admissible solutions and, in turn, influence both accuracy and numerical conditioning. We develop an approach to hyperparameter learning which seeks a configuration of the learned model that generalizes well to new data. This perspective leads to a bilevel setup in which the inner problem estimates the model by balancing data fidelity with smoothness, while the outer problem selects hyperparameters to minimize a regularized validation loss.  Our contribution is a simple, memory-efficient algorithm that replaces each full inner solve with a single Gauss--Newton linearization, yielding an explicit state update and enabling hypergradient (the derivatives of the outer loss with respect to the hyperparameters) computation without long unrolling through the inner solver. The result is a scalable method well suited to PDE-constrained and kernel-based learning.

To motivate the importance of hyperparameter learning and to highlight our main contribution, we begin in Subsection \ref{subsec:motivating_example}  with a brief overview of our method in the context of solving nonlinear PDEs using Gaussian processes (GPs), while jointly learning the hyperparameters defining the underlying covariance kernels. In Subsection \ref{ssec:RW} we provide a literature review and context for our contribution. Subsection \ref{ssec:OL} overviews
the contents of the paper.

\subsection{Hyperparameter Learning When Solving PDEs with GPs} 
\label{subsec:motivating_example}
This subsection is devoted to a motivating example, explaining our proposed methodology for hyperparameter learning within the GP-PDE framework \cite{chen2021solving}; the proposed general solution strategy for the bilevel formulation of hyperparameter learning is presented in a more general setting in Section~\ref{sec:general_bilevel}. 
Classical mesh-based PDE solvers---finite difference, finite volume, and finite element---require grid generation, careful treatment of boundary layers, and often lack rigorous uncertainty quantification. In contrast, GP solvers for PDEs~\cite{chen2021solving,chen2025sparse,meng2023sparse,owhadi2022computational,raissi2017machine,mora2025gaussian} are mesh-free, provide built-in probabilistic error estimates, and benefit from well-understood convergence guarantees. In these
aspects GP solvers for PDEs have advantages over classical methods and also over
recently proposed neural-network-based solvers which typically offer fewer theoretical underpinnings. However, existing nonlinear GP solvers \cite{chen2021solving,chen2025sparse,meng2023sparse,owhadi2022computational,raissi2017machine,mora2025gaussian} typically require selecting the solution space \emph{a priori}, a choice that strongly affects both accuracy and convergence. To motivate the role of hyperparameter learning and preview our contribution, we adopt the GP framework of \cite{chen2021solving} and augment it with a bilevel hyperparameter learning scheme. The same mechanism is method agnostic and can be adapted to other solvers and will be developed in a more abstract setting in the remainder of the paper.

We begin by summarizing the approach to PDE solving in \cite{chen2021solving}.
For $d\geq 1$, let $\Omega\subset\mathbb{R}^d$ be a bounded open domain with boundary $\partial \Omega$.  We seek to find a function $u^{\star}$ solving the nonlinear PDE
\begin{align}
\label{eq:pde:general_form}
\begin{cases}
\mathcal{P}(u^{\star})(x) = f(x), \ \forall x\in \Omega,\\
\mathcal{B}(u^{\star})(x) = g(x), \ \forall x\in \partial \Omega.
\end{cases}
\end{align}
Here $\mathcal{P}$ denotes the (possibly nonlinear) interior differential operator; $\mathcal{B}$ the boundary operator (e.g., Dirichlet, Neumann, or Robin); $f$ the source/right-hand side; and $g$ the prescribed boundary data.
Throughout this section we assume that \eqref{eq:pde:general_form} admits a unique strong solution in a
quadratic Banach space $\mathcal{U}$ associated with a positive-definite
covariance operator $\mathcal{K}$, so that all pointwise evaluations and
linear operators appearing in~\eqref{eq:pde:general_form} 
are well defined. For generalization to PDEs with rough coefficients, where pointwise evaluations are unavailable, see \cite{baptista2025solving}.

An example of \eqref{eq:pde:general_form} is the nonlinear elliptic problem posed on a bounded domain \(\Omega \subset \mathbb{R}^d\) with sufficiently smooth boundary:
\begin{equation}
\label{eq:nonlinear_elliptic}
\begin{cases}
-\Delta u(x) + u(x)^3 = f(x), & \forall x \in \Omega,\\[2pt]
u(x) = 0, & \forall x \in \partial\Omega.
\end{cases}
\end{equation}
Assume that the data \(f\) is such that \eqref{eq:nonlinear_elliptic} admits a unique classical solution.
In the notation of \eqref{eq:pde:general_form}, the interior operator and boundary operator are
\[
\mathcal{P}(u)(x) = -\Delta u(x) + u(x)^3,\quad x\in\Omega,
\qquad
\mathcal{B}(u)(x) = u(x),\quad x\in\partial\Omega,
\]
so that \(\mathcal{P}(u)=f\) in \(\Omega\) and \(\mathcal{B}(u)=0\) on \(\partial\Omega\).

The GP method proposed in \cite{chen2021solving} approximates the solution $u^{\star}$ of \eqref{eq:pde:general_form} in $\mathcal{U}_{\theta}$, a reproducing kernel Hilbert space (RKHS) with the covariance kernel
$\kappa_{\theta}$ chosen \textit{a priori}, where the finite-dimensional hyperparameter
$\theta\in\Theta$ controls lengthscales, variances, and smoothness.  Given
$M$ collocation points
$\{x_j\}_{j=1}^M\subset\overline{\Omega}$, partitioned into
$\{x_i\}_{i=1}^{M_\Omega}\subset\Omega$ and
$\{x_j\}_{j=M_\Omega+1}^M\subset\partial\Omega$, the GP approach in \cite{chen2021solving} seeks
the minimal-norm interpolant in $\mathcal{U}_{\theta}$ that  enforces
the PDE and boundary conditions at these collocation points:
\begin{align}
\begin{split}
u_{\theta}
  &=\argmin_{u\in\mathcal{U}_{\theta}}
     \|u\|_{\mathcal{U}_{\theta}}^{2}\\[-2pt]
  &\quad\text{subject to (s.t.)}\quad
    \begin{cases}
      \mathcal{P}(u)(x_i) = f(x_i), 
      &i = 1,\dots,M_\Omega,\\
      \mathcal{B}(u)(x_j) = g(x_j), 
      &j = M_\Omega+1,\dots,M.
    \end{cases}
\end{split}
\label{eq:gp_pde}
\end{align}
By choosing \(u_{\theta}\) to minimize the RKHS norm, one automatically obtains the maximum \textit{a posteriori} (MAP) estimator corresponding to a zero-mean Gaussian prior with covariance function \(\kappa_{\theta}\) and a likelihood that enforces the PDE constraints at the collocation points~\cite{chen2021solving,chen2025gaussian}. The paper~\cite{chen2021solving} also presents a regularized formulation of~\eqref{eq:gp_pde}, replacing the hard constraint at collocation points by
a soft constraint. For clarity, we omit a detailed discussion in the remainder of this subsection. Nevertheless, in both settings, the accuracy and numerical stability of GP-based methods depend on the kernel hyperparameters: poor choices can induce over-smoothing, spurious oscillations, and ill-conditioned linear systems, which in turn slow the convergence of the optimization algorithms.

\begin{figure}[tb]
    \centering
    \includegraphics[width=0.45\linewidth]{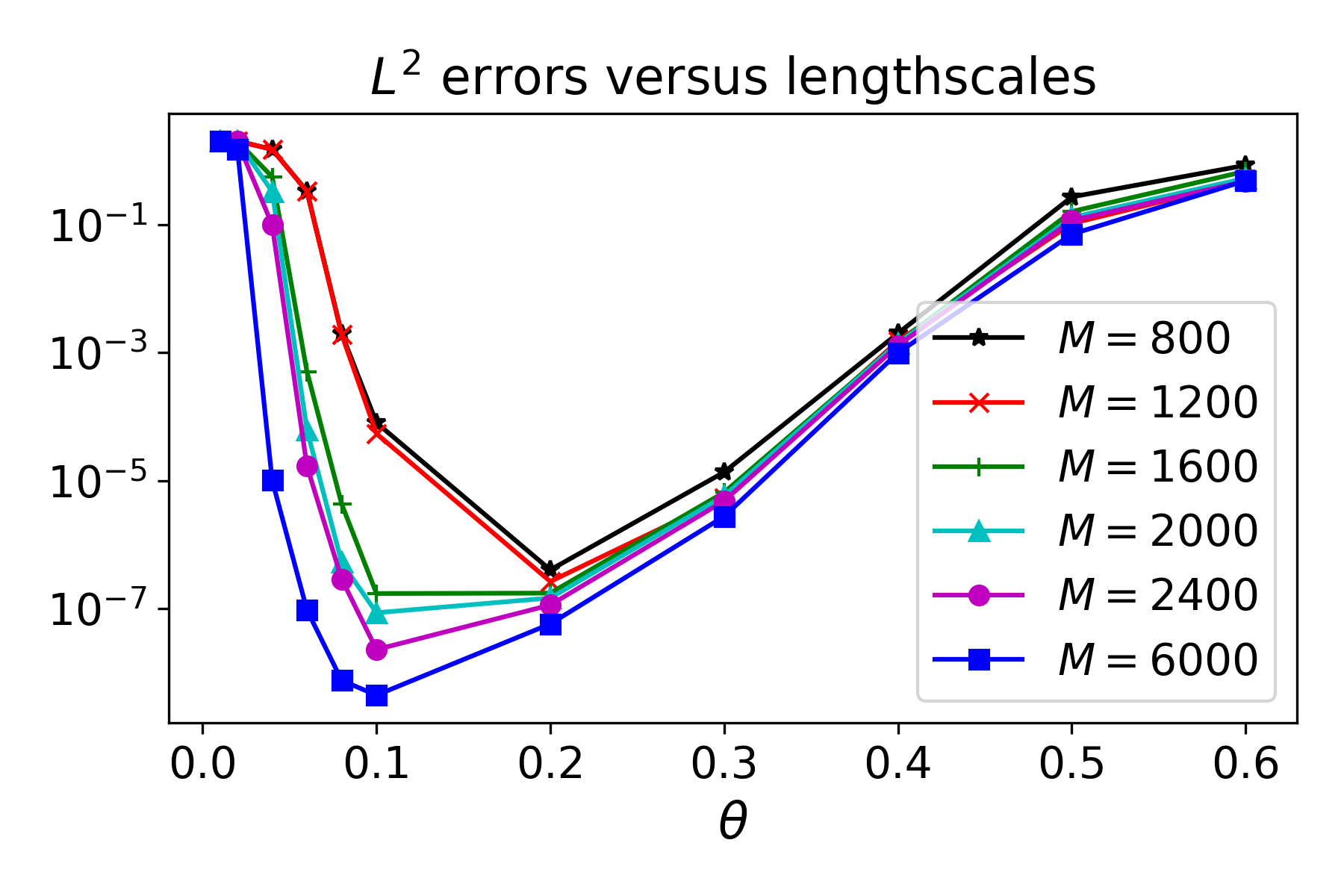}
    \includegraphics[width=0.45\linewidth]{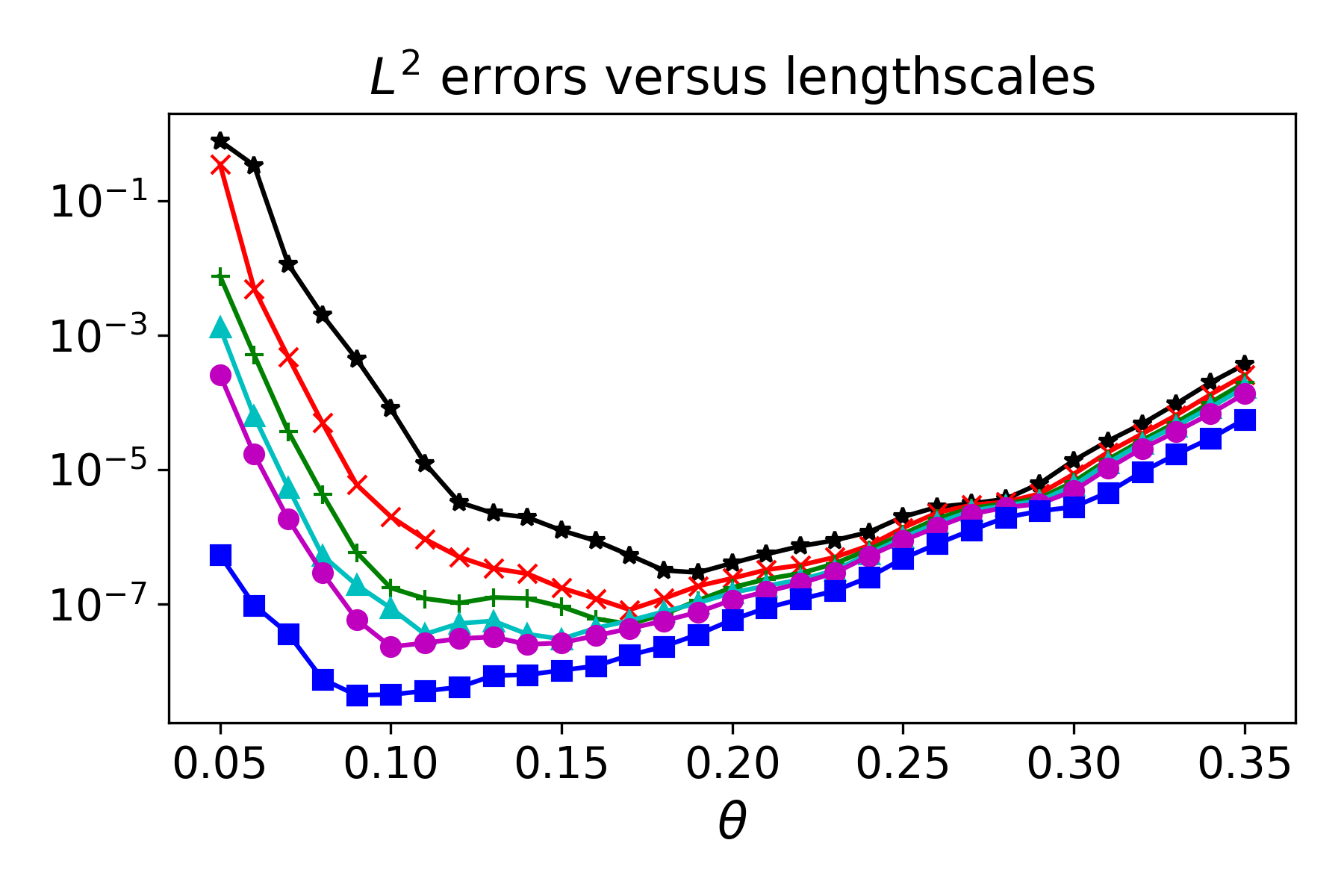}
    \caption{Left: $L^2$ errors of GP solutions \cite{chen2021solving} to \eqref{eq:nonlinear_elliptic} versus kernel lengthscale $\theta$ and training size $M$. Right: zoom for $\theta\in[0.05,0.35]$. Accuracy improves with $M$ but is sensitive to $\theta$; when $\theta$ lies far outside the low-error region, additional increases in $M$ yield  minimal gains.}
    \label{fig:elliptic_0}
\end{figure}

To quantify how kernel hyperparameters affect the GP solver of \cite{chen2021solving}, we solve \eqref{eq:nonlinear_elliptic} using a radial basis function kernel; see Section~\ref{sec:num_ellip}. The kernel has one hyperparameter, a positive scalar lengthscale \(\theta\), which controls the smoothness of the solution. We vary \(\theta\) and the number of training data points \(M\in\mathbb{N}\), and report the resulting \(L^2\) error with respect to a reference solution; see Figure~\ref{fig:elliptic_0}. For the detailed numerical setup of the experiments, see Section~\ref{sec:num_ellip}.  The results show that: (i) the error is highly sensitive to \(\theta\) for all values of \(M\); (ii) the ``optimal'' \(\theta\) depends on \(M\); and (iii) when \(\theta\) is far from the low-error regime, increasing \(M\) alone yields little improvement in accuracy. These findings indicate that automatically selecting \(\theta\) is essential for high accuracy. In principle, when the solution space is correctly specified and the number of points \(M\) tends to infinity, the fill distance (the largest distance from any point in the domain to its nearest training location) tends to zero and the approximate solution converges to the true solution \cite{chen2021solving}. In practice, however, as shown in Figure~\ref{fig:elliptic_0}, increasing \(M\) alone yields only modest gains in accuracy. This motivates the bilevel hyperparameter learning framework developed next.

To determine the optimal hyperparameters \(\theta\), we embed the GP problem~\eqref{eq:gp_pde} within a bilevel optimization framework.  At the outer level, we minimize a weighted \(L^2\) misfit that combines the PDE residual in the interior and the boundary condition residual. We weight the interior by a finite measure \(\mu\) (typically the Lebesgue measure) and the boundary by \(\nu\) (typically the Hausdorff surface measure).
The optimal \(\theta\) and its corresponding approximation \(u_{\theta}\) to the true solution \(u^{\star}\) are then obtained by solving the bilevel problem
\begin{align}
\label{eq:pde:bilevel}
\begin{dcases}
    \min_{\theta \in \Theta} \biggl(\int_\Omega \left| \mathcal{P}(u_\theta)(x)- f(x)\right|^2\dif \mu(x) + \eta \int_{\partial \Omega}\left|\mathcal{B}(u_\theta)(x) - g(x)\right|^2\dif \nu(x)\biggr)\\
    \text{s.t}\quad u_\theta \in  \argmin\limits_{u\in \mathcal{U}_\theta}\|u\|_{\mathcal{U}_\theta}^2   \\   \quad\quad\quad\quad\;\text{s.t. } 
    \begin{cases}
         \mathcal{P}(u)(x_i) = f(x_i), \quad i = 1,\dots, M_\Omega,\\
        \mathcal{B}(u)(x_i) = g({x}_j), \quad j = M_{\Omega}+1, \dots, M. 
    \end{cases}
\end{dcases}
\end{align}
Here, the weighting $\eta>0$ balances interior and boundary fidelity. In this way, the outer loop learns the optimal parameters so that the GP interpolant not only satisfies the
collocation constraints, but also minimizes the global PDE residual.

Solving the bilevel problem in \eqref{eq:pde:bilevel} is challenging because the outer objective depends implicitly on the inner solution \(u_{\theta}\), which itself is defined by a constrained minimization. A poor choice of hyperparameters \(\theta\) can render the inner minimization highly ill-conditioned or slow to converge, thereby stalling or destabilizing the outer optimization as well.  To overcome these interdependencies and ensure robust and efficient convergence, we propose two complementary solution strategies: discretize-then-optimize (DTO) and optimize-then-discretize (OTD).

\subsubsection{Optimize-Then-Discretize}
\label{subsec:OTD:general}
In the OTD approach, we first linearize both the inner PDE solve and the outer hyperparameter objective in the infinite-dimensional function space using 
Gauss-Newton (GN) expansions, guided by functional (Fr\'echet) derivatives. This yields an explicit expression for the inner minimizer in terms of the hyperparameters. Substituting this closed-form solution into the bilevel formulation produces a reduced outer objective defined entirely in function space. We then discretize this outer objective by sampling a finite set of validation points to approximate the integrals, enabling efficient gradient-based updates for \(\theta\).   These steps---functional linearization, inner solution, outer discretization, and hyperparameter update---are repeated iteratively until convergence.

Let \(\theta^k \in \Theta\) be the estimate of the kernel hyperparameters
at the \(k\)-th iteration of the OTD algorithm, and let \(u^k \in \mathcal{U}_{\theta^k}\) denote the corresponding GP approximation of the PDE solution. We linearize the nonlinear operators \(u \mapsto \mathcal{P}(u)\) and \(u \mapsto \mathcal{B}(u)\) around \(u^k\):
\begin{align}
\label{eq:dto:general_pde_disc}
\mathcal{P}(u) &\approx \mathcal{P}(u^k) + D_u \mathcal{P}(u^k)(\,u - u^k\,),\\
\mathcal{B}(u) &\approx \mathcal{B}(u^k) + D_u \mathcal{B}(u^k)(\,u - u^k\,).
\end{align}
Here \(D_u\mathcal{P}(u^k)\) and \(D_u\mathcal{B}(u^k)\) denote the Fr\'echet derivatives of \(\mathcal{P}\) and \(\mathcal{B}\) at \(u^k\). 
For \eqref{eq:nonlinear_elliptic} with \(\mathcal{P}(u) = -\Delta u + u^3\), we first define the Fr\'echet derivative at \(u^k\) as the linear operator acting on a direction \(v\):
\[
D_u \mathcal{P}(u^k)(v) \;=\; -\Delta v(x) \;+\; 3\big(u^k(x)\big)^2 v(x).
\]
Substituting \(v = u - u^k\) in \eqref{eq:dto:general_pde_disc} yields the first-order approximation
\[
\mathcal{P}(u) \;\approx\;  -\Delta u(x) + 3\big(u^k(x)\big)^2 u(x) - 2\big(u^k(x)\big)^3.
\]
An entirely analogous definition and linearization apply to \(\mathcal{B}\).

We recall that  \(\{{x}_i\}_{i=1}^{M_\Omega} \subset \Omega\) and \(\{{x}_j\}_{j=M_\Omega+1}^{M} \subset \partial \Omega\) are the collocation points for the PDE and boundary conditions, respectively. Hence,  at the $k$-th iteration, we obtain $\theta^{k+1}$ as a solution of the minimization problem
\begin{align}
\label{eq:pde:bilevel:opt_disc}
\begin{dcases}
\min\limits_{\theta \in \Theta} \biggl(\int_\Omega \left| \mathcal{P}(u^k) + D_u\mathcal{P}(u^k)(u_\theta-u^k)- f\right|^2\dif \mu + \eta \int_{\partial \Omega}\left|\mathcal{B}(u^k) +D_u\mathcal{B}(u^k)(u_\theta-u^k)- g\right|^2\dif \nu\biggr)\\
\text{s.t}\quad u_\theta \in  \argmin\limits_{u\in \mathcal{U}_\theta}\|u\|_{\mathcal{U}_\theta}^2
\\ \quad\quad\quad\quad\;\text{s.t.} \begin{cases}
 \mathcal{P}(u^k)(x_i) + D_u\mathcal{P}(u^k)(u - u^k)(x_i) = f(x_i)\qf  i = 1,\dots, M_\Omega,\\
\mathcal{B}(u^k)(x_j) + D_u\mathcal{B}(u^k)(u-u^k)(x_j) = g({x}_j) \qf j = M_{\Omega}+1, \dots, M. 
\end{cases}
\end{dcases}
\end{align}
The outer objective in \eqref{eq:pde:bilevel:opt_disc} is linearized to remain consistent with the inner problem. This ensures that at each iteration, the linearized PDE system is solved using the hyperparameters that are optimal. 
 The inner minimization in \eqref{eq:pde:bilevel:opt_disc} is a linearly constrained quadratic problem and therefore admits a unique minimizer.  At the $k$-th iteration,  we assemble the residual vector $\mathbf r^k \in\mathbb{R}^M$ defined entrywise by
 \begin{align*}
     {\mathbf r_i^k}\coloneqq
     \begin{cases}
         f(x_i)-\mathcal P(u^k)(x_i)\qif i\in\{1,\ldots, M_\Omega\},\\
         g(x_i)-\mathcal B(u^k)(x_i)\qif i\in\{M_\Omega+1,\ldots, M\}.
     \end{cases}
 \end{align*}
We also define the linear functionals
\[
\phi_i=\delta_{x_i}\circ D_u\mathcal P(u^k),\quad i=1,\dots,M_\Omega,
\qa
\phi_j=\delta_{x_j}\circ D_u\mathcal B(u^k),\quad j=M_\Omega+1,\dots,M,
\]
and stack them into $\Phi\colon \mathcal U_{\theta}\to\mathbb R^M$ defined by $(\Phi u)_m\coloneqq \phi_m(u)$.  The inner problem is then
\begin{equation}
\label{eq:otd:inner}
u_{\theta}
=\argmin_{u\in\mathcal U_{\theta}}\Bigl\{ \|u\|^2_{\mathcal U_{\theta}}
\quad\text{s.t.}\quad
\Phi\bigl(u - u^k\bigr)=\mathbf r^k\Bigr\}.
\end{equation}
Equivalently, setting $\mathbf b^k\coloneqq \Phi u^k+\mathbf r^k$, one enforces $\Phi u=\mathbf b^k$ directly.  By the representer theorem \cite[Sec. 17.8]{owhadi2019kernel}, the unique minimizer is
\begin{equation}
\label{eq:otd:explicit}
x\mapsto u_{\theta}(x)
= \kappa_{\theta}(x,\Phi)^\top\mathbf K_{\Phi}^{-1}\bigl(\Phi u^k+\mathbf r^k\bigr), 
\end{equation}
where the vector \( \kappa_{\theta}(x,\Phi) \in \mathbb{R}^M \) is defined by applying each functional \(\phi_m \in \Phi\) to the kernel section \(\kappa_{\theta}(x,\cdot)\), i.e., \( (\kappa_{\theta}(x,\Phi))_m = \phi_m(\kappa_{\theta}(x,\cdot)) \).  
The Gram matrix is $\mathbf K_{\Phi}\coloneqq \kappa_\theta(\Phi,\Phi)\in\mathbb R^{M\times M}$.

Substituting \eqref{eq:otd:explicit} into the outer objective, we define
\begin{align}
\label{eq:otd:outer}
\begin{split}
\mathcal J_k(\theta)
&\coloneqq \int_\Omega\Bigl|\mathcal P(u^k)(x)
+ D_u\mathcal P(u^k)[\,u_{\theta}-u^k\,](x) - f(x)\Bigr|^2\dif \mu(x) \\
&\quad\;+\;\eta\int_{\partial\Omega}\Bigl|\mathcal B(u^k)(x)
+ D_u\mathcal B(u^k)[\,u_{\theta}-u^k\,](x) - g(x)\Bigr|^2\dif \nu(x).
\end{split}
\end{align}
To make the outer objective~\eqref{eq:otd:outer} tractable, we replace the domain integrals with averages on a set of \(N_{\rm val}=N_\Omega+N_{\partial\Omega}\) validation points \(\{x_v^{(i)}\}_{i=1}^{N_{\Omega}}\cup \{x_v^{(j)}\}_{j=N_\Omega + 1}^{N_{\rm val}}\subset\Omega\cup\partial\Omega\).  This yields the empirical loss  
\begin{align}\label{eqn:outer_discrete_linearized}
\begin{split}
    \widehat{\mathcal J}_k(\theta)
&\coloneqq \frac{1}{N_\Omega}\sum_{i=1}^{N_\Omega}\Bigl|\mathcal P(u^k)(x_v^{(i)})
+ D_u\mathcal P(u^k)[\,u_\theta - u^k\,](x_v^{(i)}) - f(x_v^{(i)})\Bigr|^2 \\
&\quad\; +\frac{\eta}{N_{\partial\Omega}}\sum_{j=N_\Omega+1}^{N_{\rm val}}\Bigl|\mathcal B(u^k)(x_v^{(j)})
+ D_u\mathcal B(u^k)[\,u_\theta - u^k\,](x_v^{(j)}) - g(x_v^{(j)})\Bigr|^2.
\end{split}
\end{align}
The sums can be interpreted as integrals with respect to the empirical measures defined by the validation points.
We compute the gradient \(\nabla_{\!\theta}\widehat{\mathcal J}_k\) by differentiating through  \(u_\theta\) in \eqref{eq:otd:explicit} using automatic differentiation.  In our implementation, we then update the hyperparameters via a first-order optimizer, e.g., the Adam algorithm~\cite{kingma2014adam}. The new iterate \(u^{k+1}\coloneqq u_{\theta^{k+1}}\) is obtained via \eqref{eq:otd:explicit} with \(\theta=\theta^{k+1}\). These steps are repeated until a chosen scalar-valued convergence metric, 
such as \(\|\theta^{k+1}-\theta^k\|\) or \(\bigl|\widehat{\mathcal J}_k(\theta^{k+1})-\widehat{\mathcal J}_k(\theta^k)\bigr|\), falls below a prescribed tolerance.

\subsubsection{Discretize-Then-Optimize}
\label{subsec:DTO:general}
In contrast to the OTD approach where linearization is carried out in function space before numerical discretization, the DTO strategy begins by \emph{fixing} a finite set of validation points \(\{x_v^{(i)}\}_{i=1}^{N_s} \subset \Omega\) and \(\{x_b^{(j)}\}_{j=1}^{N_b} \subset \partial \Omega\) and proceeds by directly working with a fully discretized loss function. The DTO formulation offers a practical advantage in benchmarking and experimentation: the total number of PDE and boundary evaluations is fixed in advance. This makes it straightforward to compare the performance of different parameter selection strategies under a fixed computational budget, using the same pool of collocation and validation points across all methods.

We begin by replacing the domain and boundary integrals in the outer loss with discrete sums over the validation sets. The resulting outer objective reads:
\begin{equation}
\label{eq:dto:outer-discrete}
\min_{\theta \in \Theta} \biggl(
\frac{1}{N_s} \sum_{i=1}^{N_s} \left| \mathcal{P}(u_\theta)(x_v^{(i)}) - f(x_v^{(i)}) \right|^2
+ \frac{\eta}{N_b} \sum_{j=1}^{N_b} \left| \mathcal{B}(u_\theta)(x_b^{(j)}) - g(x_b^{(j)}) \right|^2\biggr),
\end{equation}
where \(u_\theta\in\mathcal U_\theta\) is the GP interpolant solving the inner problem:
\begin{equation}
\label{eq:dto:inner}
u_\theta \in \argmin_{u \in \mathcal{U}_\theta} \|u\|^2_{\mathcal{U}_\theta}
\quad \text{s.t.} \quad
\begin{cases}
\mathcal{P}(u)(x_i) = f(x_i), & i = 1,\dots,M_\Omega, \\
\mathcal{B}(u)(x_j) = g(x_j), & j = M_\Omega+1,\dots,M.
\end{cases}
\end{equation}

To facilitate gradient-based updates of \(\theta\), we again linearize the nonlinear PDE and boundary operators around the current GP approximation \(u^k\) as in \eqref{eq:dto:general_pde_disc}, which, when inserted into the objective \eqref{eq:dto:outer-discrete}, gives a locally linearized loss:
\begin{equation}
\label{eq:dto:linearized-loss}
\begin{aligned}
\min_{\theta \in \Theta} \biggl(
&\frac{1}{N_s} \sum_{i=1}^{N_s} \left| \mathcal{P}(u^k)(x_v^{(i)}) + D_u \mathcal{P}(u^k)(u_\theta - u^k)(x_v^{(i)}) - f(x_v^{(i)}) \right|^2 \\ 
&\quad\quad\quad+\frac{\eta}{N_b} \sum_{j=1}^{N_b} \left| \mathcal{B}(u^k)(x_b^{(j)}) + D_u \mathcal{B}(u^k)(u_\theta - u^k)(x_b^{(j)}) - g(x_b^{(j)}) \right|^2\biggr)
,
\end{aligned}
\end{equation}
subject to the same interpolation constraints as in \eqref{eq:pde:bilevel:opt_disc}. The solution \(u_\theta\) is again expressed via a representer formula involving the kernel and a set of operator evaluations, as previously defined in \eqref{eq:otd:explicit}. 

In practice, to improve scalability on large datasets or high-dimensional spaces, we employ stochastic approximations to the loss. For each iteration $k$, let \(B_s^{(k)} \subseteq \{1,\dots,N_s\}\) and \(B_b^{(k)} \subseteq \{1,\dots,N_b\}\) denote mini-batches of validation and boundary points, respectively. The corresponding stochastic loss becomes
\begin{equation}
\label{eq:dto:minibatch}
\begin{aligned}
\min_{\theta \in \Theta} \biggl(
&\frac{1}{|B_s^{(k)}|} \sum_{i \in B_s^{(k)}} \left| \mathcal{P}(u^k)(x_v^{(i)}) + D_u \mathcal{P}(u^k)(u_\theta - u^k)(x_v^{(i)}) - f(x_v^{(i)}) \right|^2 \\
&\quad\quad\quad+\frac{\eta}{|B_b^{(k)}|} \sum_{j \in B_b^{(k)}} \left| \mathcal{B}(u^k)(x_b^{(j)}) + D_u \mathcal{B}(u^k)(u_\theta - u^k)(x_b^{(j)}) - g(x_b^{(j)}) \right|^2\biggr).
\end{aligned}
\end{equation}
We compute gradients of the surrogate loss \eqref{eq:dto:minibatch} with respect to \(\theta\) using automatic differentiation. As in OTD, the hyperparameters are updated using a gradient-based optimizer such as Adam. The updated \(\theta^{k+1}\) is then used to resolve the inner problem and calculate \(u^{k+1}\) using \eqref{eq:otd:explicit} with $\theta^{k+1}$.

\begin{RemarkQED}
The DTO strategy offers the advantage of fixing the total number of validation points from the outset, which is particularly useful for benchmarking and ensuring \textit{fair} comparisons across methods---for example, between approaches with learned hyperparameters and those with prescribed, unlearned ones. Throughout this paper, by fair we mean comparability in terms of data usage. Specifically, the total number of training and validation data points used in the hyperparameter learning method should equal the number of training data points used in the fixed hyperparameter method. This ensures that differences in performance can be attributed to the effectiveness of hyperparameter learning itself, rather than disparities in data allocation, thereby demonstrating its value in solving both forward and inverse PDE problems. Thus, in the numerical experiments of Section~\ref{sec:numerical_results}, we include DTO solely to satisfy the fairness criterion defined above when comparing with the GP method using no learned parameters.
However, unlike OTD, the DTO scheme does not exploit the structure of the continuous outer loss prior to discretization. As a result, it may suffer from discretization bias if the number of validation points is too small or poorly chosen. Nevertheless, for well-sampled validation grids, DTO provides a practical and effective alternative to OTD. 
\end{RemarkQED}

\subsection{Related Work} \label{ssec:RW}
Hyperparameter learning is central in machine learning. A common approach is evidence maximization, also known as maximum likelihood estimation, where one selects  parameters by maximizing the marginal likelihood of the observations; see \cite{rasmussen2006gaussian,stein1999interpolation} for textbook treatments and practical guidance. While maximum likelihood is statistically efficient under correct model specification, its behavior can deteriorate when the covariance or noise family is misspecified \cite{white1982maximum, ge2023maximum}. In GP regression, several works show that cross-validation and its variants can yield better predictive performance under misspecification, and can be more robust than maximum likelihood for covariance parameter selection \cite{bachoc2013cross, naslidnyk2025comparing, bachoc2018asymptotic, chen2021consistency}. This motivates validation-based criteria in this paper.

Kernel flows (KF) adapt the kernel by minimizing the relative RKHS error between the interpolants constructed from the full data set and from a random half-subsample, thereby inducing a data-driven flow on features and inputs~\cite{owhadi2019kernel,hamzi2021learning,chen2021consistency}.
KF provides a flexible alternative to maximum likelihood estimation. However, its explicit validation objective is not aligned with a prescribed PDE or inverse task. A different approach proposes recursive feature machines to adapt kernels to data and hence perform model selection~\cite{radhakrishnan2024mechanism}. However, this method is limited to single kernels (e.g., additive kernels with learnable weights are not possible) with a lengthscale matrix as the only hyperparameter. This hyperparameter matrix is heuristically updated with ideas from active subspaces~\cite{beaglehole2024average,constantine2015active}, making the method fast but not grounded in optimization principles.
In the context of nonlinear PDE solving, it is known that the choice of hyperparameters is crucial for both physics-informed neural network~\cite{raissi2019physics, wang2022and, wang2024pinn, zou2025multi, zou2025uncertainty} and GP methods~\cite{chen2021solving, chen2025gaussian}. Recent work combines these methods by proposing a sparse radial basis function network in reproducing kernel Banach spaces~\cite{shao2025solving}. This construction features an adaptive selection of neurons, kernel centers, and kernel bandwidths.

A complementary line of work applies Bayesian optimization (BO) to learn hyperparameters. 
Classical BO algorithms model the objective function with a GP surrogate and choose evaluation points by maximizing an acquisition functional such as expected improvement or upper confidence bound \cite{movckus1974bayesian,jones1998efficient,shahriari2015taking,frazier2018tutorial}. BO is sample-efficient in many applications, but in PDE- and operator-driven learning each function evaluation may require solving a large deterministic optimization problem or a costly simulation. 
In such cases, BO's reliance on repeated full evaluations can be computationally demanding; surrogate misspecification (e.g., kernel choice, noise model) can further degrade performance, particularly in higher-dimensional hyperparameter spaces~\cite{shahriari2015taking,frazier2018tutorial}.

Bilevel formulations such as \eqref{eq:pde:bilevel} cast hyperparameter selection as an outer optimization problem over parameters coupled to an inner training problem that fits the model to data~\cite{franceschi2018bilevel, de2017bilevel}. There have been substantial developments in bilevel approaches for the data-driven solution of inverse problems, many drawing from image processing ideas~\cite[Sec.~4.3]{arridge2019solving}. Here, the outer problem is to learn the optimal regularizer or prior while the inner problem produces a point estimate for the inverse problem solution based on a fixed regularizer.
Another line of work in inverse problems bypasses gradient calculations by employing derivative-free optimization algorithms that evolve an ensemble of particles \cite{cleary2021calibrate,dunbar2025hyperparameter}. The tradeoff is that convergence may be slower than that of derivative-based methods and a large number of particles may be required. Furthermore, these ensemble Kalman-based methodologies are only exact in the infinite particle limit for a limited set of problems \cite{jorgensen2025bayesian}, including Gaussian ones. This issue is studied in the context of nonlinear filtering of Gaussian and near-Gaussian problems in \cite{carrillo2024mean}.

Gradient-based methods for bilevel optimization compute the hypergradient by implicit differentiation or truncated unrolled backpropagation \cite{pedregosa2016hyperparameter,maclaurin2015gradient,franceschi2018bilevel,shaban2019truncated,lorraine2020optimizing}; however, unrolled schemes incur substantial memory costs from reverse-mode propagation through many inner iterations \cite{maclaurin2015gradient,franceschi2018bilevel,shaban2019truncated}, while implicit differentiation demands high-accuracy inner solves at each outer step, which can dominate runtime \cite{pedregosa2016hyperparameter,lorraine2020optimizing}. A complementary approach replaces the lower-level problem with its Karush--Kuhn--Tucker (KKT) conditions, introduces Lagrange multipliers, and formulates a single-level constrained optimization problem by embedding the lower-level stationarity, dual feasibility, and complementarity conditions into the upper-level objective. The resulting large nonlinear system involving both primal and dual variables is then solved using Newton or semismooth Newton methods~\cite{de2017bilevel,calatroni2017bilevel}. This strategy often leads to large-scale systems that must be solved at each iteration, making the resulting optimization problem computationally demanding. Moreover, performance may deteriorate if constraints are violated or only approximately satisfied.

This paper proposes to replace each full inner solve with a single GN linearization of the inner problem. At each iteration, the linearized inner problem admits a closed-form expression, eliminating the need for reverse-mode unrolling. Compared with implicit differentiation, the hypergradient is computed without repeated high-accuracy inner solves. Compared with KKT/Newton formulations, the method avoids large complementarity systems and penalty schedules.
The numerical experiments in this paper demonstrate that hyperparameter learning leads to substantial improvements in solution accuracy compared to randomly initialized baselines. In particular, experiments involving kernels parameterized by neural networks highlight the method's scalability and applicability to high-dimensional hyperparameter learning.

\subsection{Outline} \label{ssec:OL}
In summary, we propose a general algorithm for bilevel hyperparameter learning. This general algorithm is introduced in Section~\ref{sec:general_bilevel}; the key step replaces each full inner solve with a linearization in the state, yielding either a closed-form update or an efficiently solved least-squares system. Section~\ref{sec:pde_system} extends the solution framework of Subsection~\ref{subsec:motivating_example} from single PDEs to PDE systems. In Section~\ref{sec:pde_inverse}, the framework is applied to PDE-constrained inverse problems. In Section~\ref{sec:numerical_results}, numerical experiments demonstrate significantly improved accuracy from learning the hyperparameters compared to their untrained counterparts, as well as the capability of our methods to handle high-dimensional hyperparameter training when learning deep kernels \cite{wilson2016deep} parameterized by neural networks. Section~\ref{sec:conclusion} provides
concluding remarks.

\section{Bilevel Optimization: Formulation and Linearization-Based Algorithms}
\label{sec:general_bilevel}
In this section, we generalize the approach of Sec.~\ref{subsec:motivating_example} to formulate a comprehensive algorithmic framework for bilevel hyperparameter optimization. Let \(\Theta \subseteq \mathbb{R}^p\) denote the admissible hyperparameter domain. For example, $\Theta$ could be a set containing kernel lengthscales and noise variances in GP regression, or (possibly a subset of) the weights and biases in neural network training. For each \(\theta \in \Theta\), let \(\mathcal{U}_\theta\) be a Hilbert space equipped with norm \(\|\cdot\|_{\mathcal{U}_\theta}\). Let \(\mathcal{H}\) be a Hilbert space of residuals with norm \(\|\cdot\|_{\mathcal{H}}\). We define the residual operator \( R_{\mathrm{train}} \colon \mathcal{U}_\theta \to \mathcal{H} \) as a map that evaluates how well a candidate parameter \( u \in \mathcal{U}_\theta \) satisfies the training data or physical constraints. For instance, in regression, \( R_{\mathrm{train}}(u) \) may represent the vector of prediction errors on the training data set, while in PDE-constrained inverse problems it may encode the residual of a discretized differential operator. This section considers learning the hyperparameter \(\theta\in \Theta\) that best supports the solution of the following estimation problem:
\begin{equation}
\label{eq:intro:general_opt_hard_cts}
\min_{u \in \mathcal{U}_\theta}
\|u\|_{\mathcal{U}_\theta}^2
\quad \text{s.t.} \quad R_{\mathrm{train}}(u) = 0.
\end{equation}
We also consider its relaxed counterpart
\begin{equation}
\label{eq:intro:general_opt}
\min_{u \in \mathcal{U}_\theta}\biggl(
\frac{1}{2} \|u\|_{\mathcal{U}_\theta}^2
+
\frac{1}{2} \|R_{\mathrm{train}}(u)\|_{\mathcal{H}}^2\biggr).
\end{equation}
Here, $R_{\mathrm{train}}$ only depends on $\theta$ through its domain $\mathcal{U}_\theta$. For example, in the GP formulation \eqref{eq:gp_pde} for a nonlinear PDE, the training residual \(R_{\mathrm{train}}(u)\) consists of the interior mismatches \(\mathcal{P}(u)(x_i)-f(x_i)\) at points \(x_i\in\Omega\) together with the boundary terms \(\mathcal{B}(u)(x_j)-g(x_j)\) at points \(x_j\in\partial\Omega\).

Problems of this type arise in a wide range of applications, including inverse problems governed by PDEs, kernel-based regression, and neural network training.
In these settings, the choice of \(\theta\) influences both the accuracy of solutions and the conditioning of the problems \eqref{eq:intro:general_opt_hard_cts} and  \eqref{eq:intro:general_opt}, which in turn affects the convergence rate of the solver. To avoid the computational cost of exhaustive grid search, we introduce an automatic hyperparameter tuning algorithm based on bilevel optimization. The algorithm treats problem~\eqref{eq:intro:general_opt_hard_cts} or~\eqref{eq:intro:general_opt} as the inner-level subproblem, with an outer objective defined by validation performance.
We develop an efficient iterative scheme. In each iteration, we linearize \(R_{\mathrm{train}}\) in problem~\eqref{eq:intro:general_opt_hard_cts} or \eqref{eq:intro:general_opt}  so that the inner minimization admits a closed-form solution for \(u=u(\theta)\) as a function of \(\theta\). This solution is then substituted into the outer objective, and the hyperparameters \(\theta\) are updated by solving the resulting optimization problem.
The procedure is repeated until the hyperparameters converge. 

More precisely, to formalize hyperparameter learning, we consider a bilevel optimization framework built upon the standard cross-validation strategy. We define a validation residual operator
\[
R_{\mathrm{val}}\colon \mathcal{U}_\theta \to \mathcal{V}
\]
that evaluates the generalization error of a candidate solution for a given parameter value. We allow $R_{\mathrm{val}}$ to map into a (possibly different) Hilbert space \(\mathcal{V}\). For example, in the GP-based PDE setting of \eqref{eq:pde:bilevel}, we define the validation residual as
\[
R_{\mathrm{val}}(u) \coloneqq \bigl(\mathcal{P}(u)-f,\ \eta(\mathcal{B}(u)-g)\bigr), \qquad u\in\mathcal{U}_\theta,
\]
viewed as an element of
\[
\mathcal{V}\coloneqq L^2(\Omega;\mu)\times L^2(\partial\Omega;\nu)
\]
equipped with the weighted norm
\[
\|(v_1,v_2)\|_{\mathcal{V}}
= \sqrt{\int_{\Omega} |v_1(x)|^{2}\,\mathrm{d}\mu(x)
+  \int_{\partial\Omega} |v_2(x)|^{2}\,\mathrm{d}\nu(x)},
\]
where \(\mu\) and \(\nu\) are measures on \(\Omega\) and \(\partial\Omega\), respectively, and \(\eta>0\) is a weighting parameter.

Hence, for fixed hyperparameter \(\theta\), we solve the inner training problem
\begin{equation}
u^{\star}_\theta
= \argmin_{u \in \mathcal{U}_\theta}\|u\|_{\mathcal{U}_\theta}^2 \quad \text{s.t.} \quad R_{\mathrm{train}}^{}(u)=0,
\label{eq:bilevel-inner-final}
\end{equation}
which yields the learned function \(u^{\star}_\theta\). The outer problem then seeks the hyperparameter vector \(\theta\in\Theta\) that minimizes the validation error, possibly regularized, as follows:
\begin{equation}
\min_{\theta \in \Theta}\biggl(
\frac{1}{2} \|R_{\mathrm{val}}^{}(u^{\star}_\theta)\|_{\mathcal{V}}^2
+ \mathcal{R}(\theta)\biggr);
\label{eq:bilevel-outer-final}
\end{equation}
here \(\mathcal{R}(\theta)\) is the regularizer on \(\theta\), included when prior structure, identifiability, or numerical stability is desired. The regularizer may be omitted by setting \(\mathcal{R}\equiv 0\). Common choices for \(\mathcal{R}\) include an \(L^2\)   penalty, which promotes stability and smoothness, and an \(L^1\) penalty, which encourages sparsity in the coefficients. The coupled system \eqref{eq:bilevel-inner-final}--\eqref{eq:bilevel-outer-final} defines a bilevel optimization problem in which the outer objective reflects generalization performance, and the inner problem ensures solution regularity and feasibility with respect to the constraints. A corresponding bilevel formulation of the regularized inner problem \eqref{eq:intro:general_opt} retains the outer loss \eqref{eq:bilevel-outer-final} and replaces the inner problem \eqref{eq:bilevel-inner-final} with \eqref{eq:intro:general_opt}.

\begin{RemarkQED}[Extension to $K$-Fold Cross-Validation]
\label{rmk:kfold}
The preceding setup, together with the algorithms introduced below, extends directly to $K$-fold cross-validation.
To describe the \(K\)-fold cross-validation procedure, for each fold \(j=1,\dots,K\) of the training data, introduce a training-fold residual operator
\[
R_{\operatorname{train}}^{(j)}\colon \mathcal{U}_{\theta}\to\mathcal{H}^{(j)}
\]
taking values in a Hilbert space \(\mathcal{H}^{(j)}\). This operator agrees with the global training residual \(R_{\mathrm{train}}\) appearing in~\eqref{eq:intro:general_opt_hard_cts}, except it is evaluated only on the complement of the \(j\)-th subset of the training data. For a fixed hyperparameter \(\theta\), the regression problem for fold \(j\) is
\begin{align*}
  u_{\theta}^{(j),\star}
  \;=\;
  \underset{u\in\mathcal{U}_{\theta}}{\argmin}\,
  \biggl\{
      \lVert u\rVert_{\mathcal{U}_{\theta}}^{2}
      \quad\text{s.t.}\quad
      R_{\mathrm{train}}^{(j)}(u)=0
  \biggr\}.
\end{align*}
Solving this problem for each fold yields \(\{u_{\theta}^{(j),\star}\}_{j=1}^{K}\), which we then carry forward to validation.
To assess the out-of-sample performance of a candidate hyperparameter, define a validation operator
\(
R_{\mathrm{val}}^{(j)}\colon \mathcal{U}_{\theta}\to\mathcal{V}^{(j)},
\)
where \(\mathcal{V}^{(j)}\) is a Hilbert space associated with the \(j\)-th validation subset. Aggregating the validation errors across folds and adding a hyperparameter regularizer \(\mathcal{R}(\theta)\) gives the $K$-fold bilevel program
\begin{align*}
  \min_{\theta\in\Theta}\biggl(
  \frac{1}{K}\sum_{j=1}^{K}
      \frac12\,\bigl\lVert R_{\mathrm{val}}^{(j)}\bigl(u_{\theta}^{(j),\star}\bigr)\bigr\rVert_{\mathcal{V}^{(j)}}^{2}
  \;+\;
  \mathcal{R}(\theta)\biggr).
\end{align*}
The approach developed in the remainder of the paper corresponds to the case $K=1$.
\end{RemarkQED}

\begin{figure}[tb]
\includegraphics[width=0.7\linewidth]{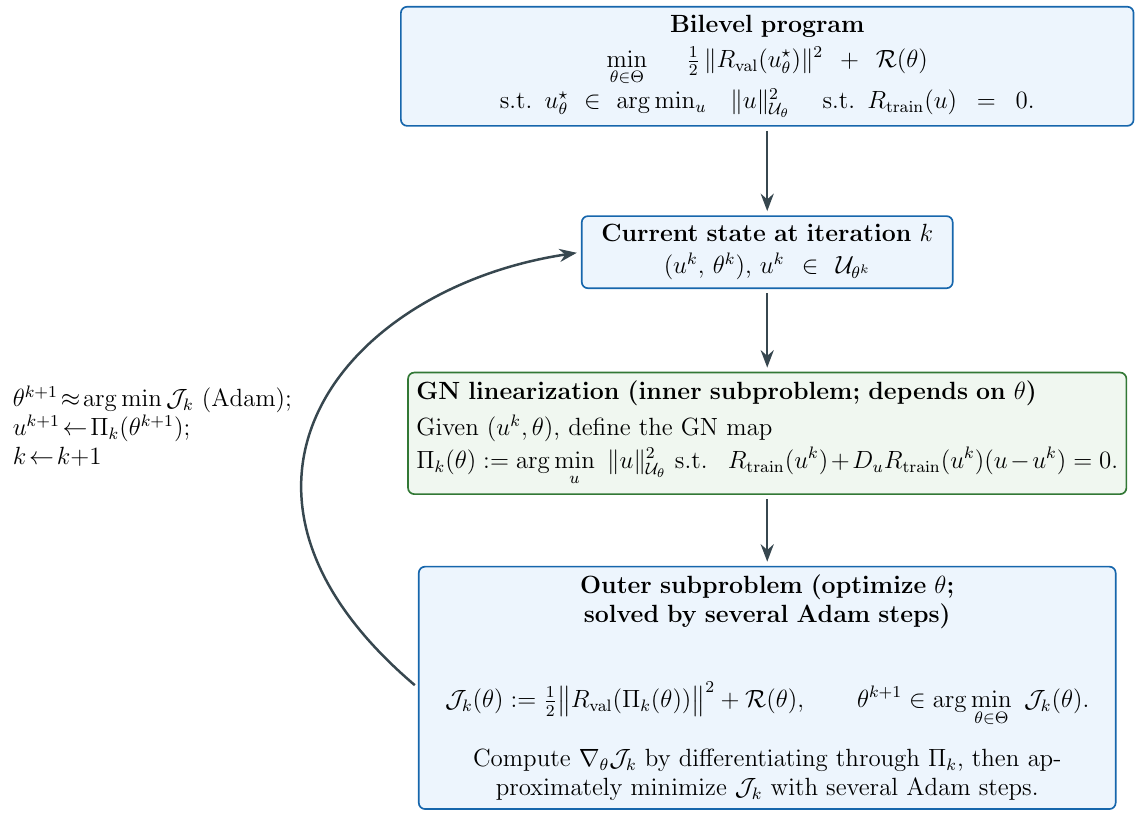}
\caption{Bilevel hyperparameter learning. At iteration $k$, given $(u^k,\theta^k)$ we (i) define the Gauss--Newton map $\Pi_k(\theta)$ by solving the linearized training constraint $R_{\mathrm{train}}(u^k)+D_u R_{\mathrm{train}}(u^k)(u-u^k)=0$ with penalty $\|u\|_{\mathcal{U}_\theta}^2$; (ii) minimize the outer objective $\mathcal{J}_k(\theta)\coloneqq\tfrac12\|R_{\mathrm{val}}(\Pi_k(\theta))\|^2+\mathcal R(\theta)$ to obtain $\theta^{k+1}\approx\arg\min_\theta \mathcal{J}_k(\theta)$ (e.g., Adam); and (iii) update the state via $u^{k+1}=\Pi_k(\theta^{k+1})$. The hypergradient $\nabla_\theta \mathcal{J}_k$ is computed by differentiating through $\Pi_k$. Here $D_u R_{\mathrm{train}}$ denotes the Fr\'echet derivative with respect to $u$.}
\label{fig:illustration_graph}
\end{figure}

The rest of this section, Subsection~\ref{subsec:linearization}, presents a practical linearization-based algorithm that avoids repeated full inner solves, long unrolling of the inner optimization path via backpropagation, and large KKT systems. Each iteration applies a single
GN step in the state variable, yielding an efficiently solvable least-squares update and, in many cases, a closed-form update.
 The resulting state is substituted into the outer objective, and the hyperparameters are updated by minimizing this reduced objective that depends only on the hyperparameters. The GN linearization and hyperparameter updates then alternate until convergence.   Figure~\ref{fig:illustration_graph} provides an overview of the procedure.
 The result is a scalable algorithm with low memory overhead and competitive accuracy that is particularly well suited to PDE-constrained and kernel-based learning problems.

\subsection{Linearization-Based Algorithm}
\label{subsec:linearization}

The fully nonlinear bilevel formulation introduced in \eqref{eq:bilevel-outer-final} and \eqref{eq:bilevel-inner-final} is
conceptually elegant yet computationally burdensome.  At every outer
iteration one must, in principle, solve the inner optimization problem to
numerical convergence, and the inner solves depend nonlinearly on
the state variable \(u\) and on the hyperparameter vector
\(\theta\).  Apart from the  cost of repeatedly invoking a nonlinear
solver, the outer-level optimization requires first- and, for Newton-type
methods, second-order derivatives of the inner solution map
\(\theta\mapsto u^{\star}_{\theta}\).  In the $K$-fold cross-validation setting in Remark \ref{rmk:kfold}, computing these derivatives via implicit
differentiation couples all folds, destroys  parallel
structures, and leads to dense linear systems whose assembly time may rival
the original inner solves themselves.  Consequently, direct implicit differentiation is seldom affordable when the forward operator embedded in the residual
\(R_{\mathrm{train}}^{}\) is expensive to evaluate.

To mitigate these difficulties, we adopt a single-step GN
approximation at the inner level.  More concretely,
let \((u^{k},\theta^{k})\) denote the outer
iterate at iteration $k$.  Around the point \(u^{k}\), we linearize the
training residual with respect to the state variable only. This gives
\begin{align}\label{eq:train-linearized}
  R_{\mathrm{train}}^{}(u)
  \;\approx\;
  r^{k}
  \;+\;
  D_{u}R_{\mathrm{train}}^{}
      \bigl(u - u^{k}\bigr),
\end{align}
where
\(r^{k}\coloneqq R_{\mathrm{train}}^{}\bigl(u^{k}\bigr)\) and the linear operator \(D_{u}R_{\mathrm{train}}^{}\coloneqq D_{u}R_{\mathrm{train}}^{}(u^k)\colon \mathcal{U}_{\theta}\to\mathcal{H}\) is the Fr\'echet derivative of $R_{\mathrm{train}}$ evaluated at $u^{k}$.  Because
both $\mathcal{U}_{\theta}$ and $\mathcal{H}$ are Hilbert spaces, the adjoint operator
\(\bigl[D_{u}R_{\mathrm{train}}^{}\bigr]^{*}\colon \mathcal{H}\to\mathcal{U}_{\theta}\)
is defined via the Riesz pairing
\[
  \bigl\langle
      D_{u}R_{\mathrm{train}}^{}p, q
  \bigr\rangle_{\mathcal{H}}
  \;=\;
  \bigl\langle
      p,
      \bigl[D_{u}R_{\mathrm{train}}^{}\bigr]^{*}q
  \bigr\rangle_{\mathcal{U}_{\theta}}
  \qfa p\in\mathcal{U}_{\theta}\qa q\in\mathcal{H}.
\]

\medskip
\paragraph*{\textbf{Linearized Inner Problem}}
For a fixed hyperparameter \(\theta\), we replace the nonlinear constraint in \eqref{eq:bilevel-inner-final} by its linear surrogate \eqref{eq:train-linearized}, yielding the following optimization problem:
\begin{align}
  \Pi_k(\theta) 
  \; \coloneqq\;
  \underset{u\in\mathcal{U}_{\theta}}{\argmin}
  \biggl\{
      \lVert u\rVert_{\mathcal{U}_{\theta}}^{2}\quad 
      \text{s.t. } r^{k}
          + D_{u}R_{\mathrm{train}}^{}\bigl(u - u^{k}\bigr)=0
  \biggr\}.
  \label{eq:lin-inner}
\end{align}
If the operator \(D_{u}R_{\mathrm{train}}\,[D_{u}R_{\mathrm{train}}]^*\) is invertible, the solution of \eqref{eq:lin-inner} is
\begin{align}
\label{eq:int:explicit_form}
  \Pi_k(\theta)
  &=
  \bigl[D_{u}R_{\mathrm{train}}^{}\bigr]^{*}
  \Bigl(
     D_{u}R_{\mathrm{train}}^{}
     \bigl[D_{u}R_{\mathrm{train}}^{}\bigr]^{*}
  \Bigr)^{-1}
  \Bigl(
     D_{u}R_{\mathrm{train}}^{} u^{k}
     - r^{k}
  \Bigr).
\end{align} 
Otherwise, interpret the inverse as the Moore--Penrose pseudoinverse, or add a small Tikhonov regularization for invertibility. Upon completion, we assemble
the  outer surrogate
\begin{align}
  \mathcal{J}_k(\theta)
  \coloneqq 
      \frac12\,
      \bigl\lVert
R_{\mathrm{val}}^{}\!\bigl(\Pi_k(\theta)\bigr)
      \bigr\rVert_{\mathcal{V}}^{2}
  \;+\;
  \mathcal{R}(\theta)
  \label{eq:outer-surrogate}
\end{align}
and evolve the hyperparameter by solving the reduced optimization
\begin{align}
  \theta^{k+1}
  \coloneqq 
  \underset{\theta\in\Theta}{\argmin}\;
  \mathcal{J}_k(\theta).
  \label{eq:outer-update}
\end{align}
Then, we obtain \( u^{k+1} \) by setting $u^{k+1}=\Pi_k(\theta^{k+1})$ using \eqref{eq:int:explicit_form}. This process is repeated until convergence. Figure~\ref{fig:illustration_graph} depicts the core concept of our algorithm.

\medskip
\paragraph*{\textbf{Regression Case}}
In the bilevel formulation corresponding to the relaxed problem \eqref{eq:intro:general_opt}, Eqn.~\eqref{eq:lin-inner} is replaced by the following constrained minimization
\begin{align}
\label{eq:reg:linear_mini}
\min_{u\in\mathcal{U}_{\theta}}\biggl(
  \frac{1}{2}\lVert u\rVert_{\mathcal{U}_{\theta}}^{2} + \frac{1}{2}\lVert r^{k}+
  D_{u}R_{\mathrm{train}}^{}
  \bigl(u - u^{k}\bigr)\rVert_{\mathcal{H}}^2
  \biggr).
\end{align}
Hence, the solution to \eqref{eq:reg:linear_mini} induces the map \(\theta \mapsto \Pi_k(\theta)\), which is characterized by the linear system
\begin{equation}
\label{eq:normal-system}
H_u^{k}\,\Pi_k(\theta) = -\,g_u^{k},
\end{equation}
where
\begin{align*}
  H_{u}^{k}
  &= I_{\mathcal{U}_{\theta}}
     +
       \bigl[D_{u}R_{\mathrm{train}}^{}\bigr]^{*}
       D_{u}R_{\mathrm{train}}^{}\qa
  \\[4pt]
  g_{u}^{k}
  &= 
     \bigl[D_{u}R_{\mathrm{train}}^{}\bigr]^{*}
     \bigl(
       r^{k}
       - D_{u}R_{\mathrm{train}}^{}u^{k}
     \bigr).
\end{align*}
The operator \(H_u^{k}\) is self-adjoint and coercive. Hence, it is invertible, and \eqref{eq:normal-system} admits a unique solution for any right-hand side. In practice, the system can be solved efficiently using matrix-free Krylov methods or direct solvers. The outer surrogate and the hyperparameter update retain
the form of~\eqref{eq:outer-surrogate}-\eqref{eq:outer-update}.

By eschewing repeated nonlinear solves in favor of a single GN step
per outer iteration, the linearization-based algorithm transforms the nested
bilevel structure into a sequence of linear systems.  These
advantages render the approach well suited to large-scale problems in scientific computing, where the forward model is expensive.

\section{Solving Nonlinear PDE Systems with GPs and Hyperparameter Learning}
\label{sec:pde_system}

In this section we extend the scalar GP-PDE framework, introduced in Section~\ref{sec:intro},
to a general $m$-component PDE system. Minimal new notation is needed to achieve this extension, and this is introduced in Subsection \ref{ssec:gen};
and we illustrate the extension using the Gray-Scott reaction-diffusion equations in Subsection \ref{subsec:grayscott_example}.

\subsection{General Setting} \label{ssec:gen}
Let $\mathbf u^\star=(u_1^\star,\dots,u_m^\star)^\top$ solve
\begin{align}
\label{eq:sys:general_form_repeat}
\mathcal{P}(\mathbf u^\star)=\mathbf f\quad\text{in }\Omega,
\qquad
\mathcal{B}(\mathbf u^\star)=\mathbf g\quad\text{on }\partial\Omega,
\end{align}
where $\mathcal{P}$  and $\mathcal{B}$ may be nonlinear and act componentwise or with cross couplings between the $m$ components of the solution. We assume that \eqref{eq:sys:general_form_repeat} admits a unique strong solution compatible with pointwise evaluations.

We model $\mathbf u=(u_1,\dots,u_m)^\top$ with either independent GPs---corresponding to a block-diagonal kernel---or a multi-output GP---corresponding to a cross-correlated kernel. Denote by $\mathcal{U}_\theta$ the vector-valued RKHS induced by the matrix-valued kernel $\kappa_\theta\colon\Omega\times\Omega\to\mathbb{R}^{m\times m}$ \cite{owhadi2023ideas}. The hyperparameters $\theta$ control marginal lengthscales, variances, and cross-covariances, if used. Given interior collocation points $\{x_i\}_{i=1}^{M_\Omega}\subset\Omega$ and boundary points $\{x_j\}_{j=M_\Omega+1}^{M}\subset\partial\Omega$, the inner GP-PDE solve is the minimal-norm interpolant
\begin{align}
\label{eq:gp_pde_system_vector}
\mathbf u_\theta
=\argmin_{\mathbf u\in\mathcal{U}_\theta}\ \|\mathbf u\|_{\mathcal{U}_\theta}^2
\quad\text{s.t.}\quad
\mathcal{P}(\mathbf u)(x_i)=\mathbf f(x_i),\ i\le M_\Omega\qa
\mathcal{B}(\mathbf u)(x_j)=\mathbf g(x_j),\ j>M_\Omega.
\end{align}
This is the direct vector analog of the scalar problem. As in the scalar case, soft-penalty variants are possible, but we retain hard constraints for clarity.

In the DTO scheme, to select $\theta$, we use a residual-based outer objective over validation sets. Let $\{x_v^{(i)}\}_{i=1}^{N_{s}}\subset\Omega$ and $\{x_b^{(j)}\}_{j=1}^{N_{b}}\subset\partial\Omega$ be fixed. With weight $\eta>0$, we learn $\theta$ by solving the bilevel minimization problem
\begin{align}
\label{eq:sys:bilevel_discrete}
\min_{\theta\in\Theta}\biggl\{ \widehat{\mathcal{J}}(\theta)
\coloneqq \frac{1}{N_{s}}\sum_{i=1}^{N_s}
\big\|\mathcal{P}(\mathbf u_\theta)(x_v^{(i)})-\mathbf f(x_v^{(i)})\big\|^2
+ \eta\,\frac{1}{N_b}\sum_{j=1}^{N_b}
\big\|\mathcal{B}(\mathbf u_\theta)(x_b^{(j)})-\mathbf g(x_b^{(j)})\big\|^2\biggr\}
\quad\text{s.t.}\ \eqref{eq:gp_pde_system_vector}.
\end{align}

To solve \eqref{eq:sys:bilevel_discrete}, we linearize around the current inner solution $\mathbf u^k$ at iteration $k$ using Fr\'echet derivatives as in the scalar derivation:
\begin{align}
\mathcal{P}(\mathbf u_\theta)\approx \mathcal{P}(\mathbf u^k)
+ D_{\mathbf u}\mathcal{P}(\mathbf u^k)\big(\mathbf u_\theta-\mathbf u^k\big),\qquad
\mathcal{B}(\mathbf u)\approx \mathcal{B}(\mathbf u^k)
+ D_{\mathbf u}\mathcal{B}(\mathbf u^k)\big(\mathbf u_\theta-\mathbf u^k\big).
\end{align}
This yields a linearly constrained quadratic inner problem and, by the representer theorem \cite{owhadi2023ideas}, a closed-form finite expansion for the inner solution together with a linearized outer objective (i.e., the residuals are linearized). Substituting the closed form into the outer objective produces a reduced loss in $\theta$; minimizing this yields $\theta^{k+1}$. We then resolve the inner problem at $\theta^{k+1}$ to obtain $\boldsymbol{u}^{k+1}\coloneqq\boldsymbol{u}_{\theta^{k+1}}$ and repeat these steps until convergence.

\subsection{Gray--Scott Reaction-Diffusion System}
\label{subsec:grayscott_example}
We now specialize the DTO scheme to the Gray--Scott model, a prototypical nonlinear reaction-diffusion system, and omit the analogous OTD scheme for brevity. The numerical results are shown in Subsection~\ref{sec:num_gray}. Let $\mathsf{D}_u, \mathsf{D}_v>0$ denote the diffusion coefficients, and let $F>0$ (feed rate) and $k>0$ (kill rate) be given parameters. We consider $(t,x)\in \Omega\coloneqq (0,1)\times(0,1)$ and seek species concentrations $u,v\colon \Omega\to\mathbb{R}$ governed by
\begin{subequations}
    \begin{align}
    \partial_t u &= \mathsf{D}_u\,\partial_{xx}u - u v^2 + F(1-u), \qquad \forall t\in(0,1),\; x\in(0,1), \label{eq:gs:u}\\
    \partial_t v &= \mathsf{D}_v\,\partial_{xx}v + u v^2 - (F+k)\,v, \qquad \forall t\in(0,1),\; x\in(0,1), \label{eq:gs:v}
    \end{align}
\end{subequations}
with homogeneous Neumann boundaries
\[
\partial_x u(0,t)=\partial_x u(1,t)=0,\qquad
\partial_x v(0,t)=\partial_x v(1,t)=0,\qquad \forall t\in(0,1),
\]
and initial conditions
\begin{align}
u(x,0)=-\sin\!\Big(3\pi x+\frac{\pi}{2}\Big),\qquad
v(x,0)=\cos(2\pi x),\qquad \forall x\in(0,1). \label{eq:gs:initial_1}
\end{align}
The system can be cast into the general form \eqref{eq:sys:general_form_repeat}. Define the interior operator and its data by
\[
\mathcal{P}(u,v)\coloneqq\big(\mathcal{P}_u(u,v),\,\mathcal{P}_v(u,v)\big),\qquad
\boldsymbol{f}\coloneqq(0,0),
\]
with
\[
\mathcal{P}_u(u,v)\coloneqq\partial_t u - \mathsf{D}_u\,\partial_{xx}u + u v^2 - F(1-u),\qquad
\mathcal{P}_v(u,v)\coloneqq\partial_t v - \mathsf{D}_v\,\partial_{xx}v - u v^2 + (F+k)\,v.
\]
Meanwhile, we define the boundary operator
\[
\mathcal{B}(u,v)\coloneqq\Big(
\partial_x u(\,\cdot\,,0),\ \partial_x u(\,\cdot\,,1),\ 
\partial_x v(\,\cdot\,,0),\ \partial_x v(\,\cdot\,,1),\ 
u(0,\,\cdot\,),\ v(0,\,\cdot\,)
\Big)
\]
and define the  vector
\[
\boldsymbol{g}\coloneqq\Big(
0,\,0,\,0,\,0,\,u_0(\,\cdot\,),\,v_0(\,\cdot\,)
\Big).
\]
With these choices, the Gray--Scott system is exactly \(\mathcal{P}(u,v)=\boldsymbol{f}\) in \(\Omega\) and \(\mathcal{B}(u,v)=\boldsymbol{g}\) on the boundary.

To solve the system, we assume independent GPs $u \sim \mathsf{GP}\!\big(0,\kappa_\theta^{(u)}\big)$ and $ 
v \sim \mathsf{GP}\!\big(0,\kappa_\theta^{(v)}\big)$
with anisotropic squared-exponential kernels
\begin{align}
\label{eq:gray_scott:kernels}
\begin{split}
\kappa_\theta^{(u)}((t,x),(t',x'))&=\exp\!\Big(-\tfrac{|t-t'|^2}{2(\ell_t^{u})^2}-\tfrac{|x-x'|^2}{2(\ell_x^{u})^2}\Big),\\
\kappa_\theta^{(v)}((t,x),(t',x'))&=\exp\!\Big(-\tfrac{|t-t'|^2}{2(\ell_t^{v})^2}-\tfrac{|x-x'|^2}{2(\ell_x^{v})^2}\Big),
\end{split}
\end{align}
with hyperparameters $\theta=(\ell_t^{u},\ell_x^{u},\ell_t^{v},\ell_x^{v})$. Let $\mathcal{H}_\theta^{(u)}$ and $\mathcal{H}_\theta^{(v)}$ be the induced RKHSs and set
\[
\mathcal{H}_\theta\coloneqq\mathcal{H}_\theta^{(u)}\times\mathcal{H}_\theta^{(v)},\qw
\|(u,v)\|_{\mathcal{H}_\theta}^2\coloneqq\|u\|_{\mathcal{H}_\theta^{(u)}}^2+\|v\|_{\mathcal{H}_\theta^{(v)}}^2.
\]
Choose disjoint collocation point sets
\[
\Xi_\Omega\subset\Omega,\quad
\Xi_{\partial\Omega}\subset(0,1)\times\{0,1\},\qa
\Xi_c\subset\{0\}\times(0,1)
\]
such that $\# \Xi_\Omega=M_\Omega$ and $\# \Xi_{\partial\Omega}+\# \Xi_c =M-M_\Omega$.

Then, given a hyperparameter $\theta$, we obtain the GP solution $(u_\theta,v_\theta)$ as the minimizer of the RKHS norm subject to the Gray-Scott system at collocation points:
\begin{align}
\label{eq:inner_nonlinear_theta}
\min_{(u,v)\in\mathcal{H}_\theta}\ &\tfrac12\|(u,v)\|_{\mathcal{H}_\theta}^2
\\
\text{s.t.}\quad
&\mathcal{P}_u(u,v)(t,x)=0,\quad
 \mathcal{P}_v(u,v)(t,x)=0, &&
\forall (t,x)\in\Xi_\Omega, \notag\\
&\partial_x u(\tau,y)=0,\quad
 \partial_x v(\tau,y)=0, &&
\forall (\tau,y)\in\Xi_{\partial\Omega}, \notag\\
&u(0,z)=u_0(z),\quad
 v(0,z)=v_0(z), &&
\forall (0,z)\in\Xi_c. \notag
\end{align}

Let $\eta>0$ be the boundary condition and initial condition regularization weight. To score a candidate $\theta$, we evaluate  residuals on held-out validation sets
\begin{align}\label{eqn:scott_val_sets}
    \{(t_v^{(i)},x_v^{(i)})\}_{i=1}^{N_s}\subset\Omega,\quad
\{(t_b^{(j)},x_b^{(j)}\}_{j=1}^{N_b}\subset(0,1)\times\{0,1\},\quad
\{(0,x_c^{(k)})\}_{k=1}^{N_c}\subset\{0\}\times(0,1).
\end{align}
To this end, define
\begin{align}
\label{eq:outer_disc_theta_nl}
{\mathcal{J}}(\theta)
&\coloneqq \frac{1}{N_s}\sum_{i=1}^{N_s}
\Bigl(|\mathcal{P}_u(u_\theta,v_\theta)(t_v^{(i)},x_v^{(i)})|^2+|\mathcal{P}_v(u_\theta,v_\theta)(t_v^{(i)},x_v^{(i)})|^2\Bigr)
\\
&\quad
+ \eta\!\left(
\frac{1} {N_b}\sum_{j=1}^{N_b}
\Bigl(|\partial_x u_\theta(t_b^{(j)},0)|^2+|\partial_x u_\theta(t_b^{(j)},1)|^2
+|\partial_x v_\theta(t_b^{(j)},0)|^2+|\partial_x v_\theta(t_b^{(j)},1)|^2\Bigr)\right.
\notag\\
&\qquad\qquad\left.+\;
\frac{1}{N_{c}}\sum_{k=1}^{N_{c}}
\Bigl(|u_\theta(0,x_c^{k})-u_0(x_c^{k})|^2+|v_\theta(0,x_c^{k})-v_0(x_c^{k})|^2\Bigr)
\right).
\notag
\end{align}

The bilevel minimization problem for simultaneously solving the PDE and learning $\theta$ is
\begin{align}
\label{eq:bilevel_nl}
\min_{\theta\in\Theta}\ {\mathcal{J}}(\theta)
\qquad \text{s.t.}\qquad (u_\theta,v_\theta)\ \text{solves \eqref{eq:inner_nonlinear_theta} in }\mathcal{H}_\theta.
\end{align}
Next, we detail the DTO scheme. At iteration $k$, given $\theta^k$ and its inner solution $(u^k,v^k)$, we linearize the interior operators using Fr\'echet derivatives:
\begin{align}
\begin{split}
D_{(u,v)}\mathcal{P}_u(u^k,v^k)(\delta u,\delta v)
&= \partial_t \delta u - \mathsf{D}_u\,\partial_{xx}\delta u
   + (v^k)^2\,\delta u + 2\,u^k v^k\,\delta v + F\,\delta u, \\
D_{(u,v)}\mathcal{P}_v(u^k,v^k)(\delta u,\delta v)
&= \partial_t \delta v - \mathsf{D}_v\,\partial_{xx}\delta v
   - (v^k)^2\,\delta u - 2\,u^k v^k\,\delta v + (F+k)\,\delta v.
   \end{split}
\end{align}

Introduce
\begin{align}
\mathcal{L}_k^{(u)}(u,v)&\coloneqq D_{(u,v)}\mathcal{P}_u(u^k,v^k)(u,v), &
y_k^{(u)}&\coloneqq\mathcal{L}_k^{(u)}(u^k,v^k)-\mathcal{P}_u(u^k,v^k),\\
\mathcal{L}_k^{(v)}(u,v)&\coloneqq D_{(u,v)}\mathcal{P}_v(u^k,v^k)(u,v), &
y_k^{(v)}&\coloneqq\mathcal{L}_k^{(v)}(u^k,v^k)-\mathcal{P}_v(u^k,v^k).
\end{align}
Then, we replace the nonlinear interior constraints and residuals by their linearized counterparts. Hence, the inner problem becomes
\begin{equation}
\label{eq:inner_linear_theta}
\begin{aligned}
\min_{(u,v)\in\mathcal{H}_\theta}\ &\tfrac12\|(u,v)\|_{\mathcal{H}_\theta}^2 \\
\text{s.t.}\quad
&\mathcal{L}_k^{(u)}(u,v)(t,x)=y_k^{(u)}(t,x),
&& \forall(t,x)\in\Xi_\Omega,\\
&\mathcal{L}_k^{(v)}(u,v)(t,x)=y_k^{(v)}(t,x),
&& \forall (t,x)\in\Xi_\Omega, \\
&\partial_x u(\tau,\xi)=0,\quad \partial_x v(\tau,\xi)=0,
&& \forall (\tau,\xi)\in\Xi_{\partial\Omega},\\
&u(0,z)=u_0(z),\quad v(0,z)=v_0(z),
&& \forall (0,z)\in\Xi_c. 
\end{aligned}
\end{equation}

The resulting problem is a linearly constrained quadratic program whose unique minimizer admits a finite-dimensional kernel representer expansion (see \cite[Sec.~17.8]{owhadi2019operator}). 
To optimize the hyperparameter $\theta$, we linearize the outer objective. Recall the validation sets~\eqref{eqn:scott_val_sets}. Then, we compute linearized residuals on the validation sets:
\begin{align}
\label{eq:outer_disc_theta_lin}
\begin{split}
\widehat{\mathcal{J}}_k(\theta)
&\coloneqq \frac{1}{N_s}\sum_{i=1}^{N_{s}}
\Bigl(
|\mathcal{L}_k^{(u)}(u_\theta,v_\theta)(t_v^{(i)},x_v^{(i)})-y_k^{(u)}(t_v^{(i)},x_v^{(i)})|^2
+
|\mathcal{L}_k^{(v)}(u_\theta,v_\theta)(t_v^{(i)},x_v^{(i)})-y_k^{(v)}(t_v^{(i)},x_v^{(i)})|^2
\Bigr)
\\
&\quad
+ \eta\Biggl(
\frac{1}{N_b}\sum_{j=1}^{N_b}
\bigl(
|\partial_x u_\theta(t_b^{(j)},0)|^2+|\partial_x u_\theta(t_b^{(j)},1)|^2
+|\partial_x v_\theta(t_b^{(j)},0)|^2+|\partial_x v_\theta(t_b^{(j)},1)|^2
\bigr)
\\
&\quad+ \frac{1}{N_{c}}\sum_{\ell=1}^{N_c}
\bigl(|u_\theta(0,x_c^{(\ell)})-u_0(x_c^{(\ell)})|^2+|v_\theta(0,x_c^{(\ell)})-v_0(x_c^{(\ell)})|^2\bigr)
\Biggr).
\end{split}
\end{align}

The linearized bilevel problem at iteration $k$ is
\begin{align}
\label{eq:bilevel_theta_lin}
\min_{\theta\in\Theta}\ \widehat{\mathcal{J}}_k(\theta)
\qquad\text{s.t.}\qquad
(u_\theta,v_\theta)\ \text{solves \eqref{eq:inner_linear_theta} in }\mathcal{H}_\theta.
\end{align}
The DTO algorithm proceeds as follows. Given $\theta$, first solve the linear inner problem \eqref{eq:inner_linear_theta} to obtain the closed-form $(u_\theta,v_\theta)$ via the representer expansion. Next, evaluate the linearized outer loss $\widehat{\mathcal{J}}_k(\theta)$ in \eqref{eq:outer_disc_theta_lin} at $(u_\theta,v_\theta)$. Then, differentiate $\widehat{\mathcal{J}}_k$ with respect to $\theta$. Finally, update $\theta$ with the Adam optimizer and resolve the inner problem at the new hyperparameters. We iterate the above procedure until convergence, e.g., until $\widehat{\mathcal{J}}_k$ or $\theta$ stabilizes.
In Section \ref{sec:numerical_results} we turn our attention to numerical results, including for this specific Gray-Scott reaction-diffusion problem.

\section{Solving Inverse Problems with GPs and Hyperparameter Learning}
\label{sec:pde_inverse}
This section illustrates how the bilevel framework introduced in Section~\ref{sec:general_bilevel} can be leveraged to solve inverse problems governed by PDE models, while simultaneously learning hyperparameters to optimize the solution space in which the inversion is carried out.  The inner level adopts the GP method of \cite{chen2021solving}, so that each PDE solve is found from a minimum RKHS norm interpolant that satisfies the governing equation at collocation points. The outer level adjusts the kernel hyperparameters by minimizing the individual residual for each governing PDE, augmented by a data-misfit term. To solve the resulting bilevel problem efficiently, we investigate two complementary strategies: OTD, in which a GN algorithm is first applied in the continuous function space and the outer objective is  discretized only for hyperparameter updates; and DTO, in which validation points are fixed \textit{a priori}, the inner problem is solved via GN iterations, and the hyperparameters are subsequently refined. Together, OTD preserves fidelity to the continuous formulation, while DTO ensures direct control over discretization error and efficient implementation. We first present a general framework for solving inverse problems in Subsection~\ref{subsec:bilevel:inverse}, enabling simultaneous coefficient recovery and data-driven hyperparameter learning. We then instantiate this framework in the context of a Darcy-flow inverse problem in Section~\ref{subsec:darcy_flow_example}.

\subsection{General Setting}
\label{subsec:bilevel:inverse}
We now demonstrate our bilevel framework for solving inverse problems. Let the state \( u^{\star}\colon \Omega \to \mathbb{R} \) and the coefficient \( a^{\star}\colon \Omega \to \mathbb{R} \) satisfy a nonlinear PDE of the form 
\begin{equation}
\label{eq:inv:pde}
\begin{cases}
\mathcal{P}(u^{\star},a^{\star})(x) = f(x), & \forall\, x \in \Omega, \\
\mathcal{B}(u^{\star},a^{\star})(x) = g(x), & \forall\, x \in \partial \Omega. 
\end{cases}
\end{equation}
Here $\mathcal{P}$ denotes the (possibly nonlinear) interior differential operator acting on the state $u$ and coefficient $a$, while $\mathcal{B}$ is the boundary operator. The maps $f$ and $g$ specify the source/right-hand side and the prescribed boundary data, respectively. Suppose we observe noisy pointwise measurements \( u^o \coloneqq \{u_\ell^o\}_{\ell=1}^L \) of the state at given specified locations \( \{x_\ell^o\}_{\ell=1}^L \subset \Omega \). Specifically, for $u^\star=u^\star(\,\cdot\,;a^\star)$ the observations \( u_\ell^o \) are given by
\begin{align}
    u_\ell^o = u^{\star}(x_\ell^o) + \xi_\ell 
\end{align}
where \( \xi_\ell \sim \mathcal{N}(0, \gamma^2) \) are assumed independent and identically distributed (i.i.d.), and where \( \gamma > 0 \). The coefficient \( a \) is  not observed directly. We frame the inverse problem as one of jointly reconstructing the unknown solution \( u^{\star} \) and unknown coefficient \( a^{\star} \) from noisy observations of \( u^{\star} \), while enforcing the underlying PDE and boundary constraints.

To solve the inverse problem with the PDE constraints in \eqref{eq:inv:pde}, we approximate both \( u^{\star} \) and \( a^{\star} \) by independent zero-mean GPs with kernels \( \kappa_u \) and \( \kappa_a \), parameterized by hyperparameters
$\theta_u$ and $\theta_a$  respectively. These parameters define, respectively, the RKHS \( \mathcal{U}_{\theta_u} \) and the RKHS \( \mathcal{A}_{\theta_a} \), in which our GP  approximations takes place. We define \( \theta \coloneqq (\theta_u, \theta_a) \in \Theta \). Our goal is to jointly obtain the pair \( (u_{\theta}, a_{\theta}) \) and optimize the hyperparameters to best fit the observed data and satisfy the PDE constraints.
In this setting, we extend the bilevel framework for solving PDEs introduced in the previous subsection. The inner optimization problem adopts the inverse problem framework in \cite{chen2021solving}, which minimizes a combined RKHS norm regularization on \(u\) and \(a\), while incorporating observed data  at observation locations \(\{x_\ell^o\}_{\ell=1}^L\). Specifically, the inner problem reads
\[
(u_\theta, a_\theta) \in \argmin_{(u,a)\in\mathcal{U}_{\theta_u}\times\mathcal{A}_{\theta_a}} \biggl(\|u\|_{\mathcal{U}_{\theta_u}}^2 + \|a\|_{\mathcal{A}_{\theta_a}}^2 + \frac{1}{\gamma^2} \sum_{\ell=1}^L |u(x_\ell^o) - u_\ell^o|^2\biggr),
\]
subject to the PDE and boundary constraints
\[
\mathcal{P}(u,a)(x_i) = f(x_i), \quad i=1,\dots,M_\Omega, \qa
\mathcal{B}(u,a)(x_j) = g(x_j), \quad j = M_\Omega +1,\dots,M.
\]
Then, we formulate the bilevel optimization problem
\begin{equation}
\label{eq:inv:bilevel}
\begin{dcases}
\min_{\theta \in \Theta} \; \biggl(
\int_\Omega \left| \mathcal{P}(u_\theta, a_\theta)(x)-f(x) \right|^2 \dif \mu(x) 
+ \eta_1 \int_{\partial \Omega} \left| \mathcal{B}(u_\theta, a_\theta)(x)-g(x) \right|^2 \dif\nu(x) 
+ \eta_2 \sum_{\ell=1}^L \left| u_\theta(x_\ell^o) - u_\ell^o \right|^2\biggr), \\[1em]
\text{s.t. } 
(u_\theta, a_\theta) \in \argmin_{(u,a)\in\mathcal{U}_{\theta_u}\times\mathcal{A}_{\theta_a}} \biggl(\|u\|_{\mathcal{U}_{\theta_u}}^2 + \|a\|_{\mathcal{A}_{\theta_a}}^2 + \frac{1}{\gamma^2} \sum_{\ell=1}^L |u(x_\ell^o) - u_\ell^o|^2\biggr)
 \\[0.5em]
\quad\quad\quad\quad\quad\quad\quad\text{s.t. }\  
\mathcal{P}(u,a)(x_i) = 0,\;\; i=1,\dots,M_\Omega,\\ \quad\quad\quad\quad\quad\quad\quad\quad\quad\quad\quad\quad
\mathcal{B}(u,a)(x_j) = 0,\;\; j=M_\Omega+1,\dots,M, 
\end{dcases}
\end{equation}
where \( \eta_1 \) and \( \eta_2 \) are regularization parameters that balance the influence of the boundary constraint and the data misfit, respectively.

\subsubsection{Optimize-Then-Discretize}
We linearize the constraints around the current estimates \( (u^k, a^k) \in \mathcal{U}_{\theta_u^k} \times \mathcal{A}_{\theta_a^k} \) using Fr\'echet derivatives to obtain linearized surrogate operators $\mathcal{P}_k$ and $\mathcal{B}_k$ such that 
\begin{align}
\mathcal{P}(u,a) &\approx \mathcal{P}_k(u, a) \coloneqq \mathcal{P}(u^k,a^k) + D_u\mathcal{P}(u^k,a^k)(u - u^k) + D_a\mathcal{P}(u^k,a^k)(a-a^k), \\
\mathcal{B}(u,a)(x) &\approx \mathcal{B}_k(u, a) \coloneqq  \mathcal{B}(u^k,a^k) + D_u\mathcal{B}(u^k,a^k)(u- u^k) + D_a\mathcal{B}(u^k,a^k)(a - a^k).
\end{align}
Hence, at the $k$-th step, we get $\theta^{k+1}$ as the minimizer of the following surrogate problem
\begin{equation}
\label{eq:inv:surrogate_bilevel}
\begin{dcases}
\min_{\theta \in \Theta} \; \biggl(
\int_\Omega \left| \mathcal{P}_k(u_\theta, a_\theta)(x)-f(x) \right|^2 \dif\mu(x) 
+ \eta_1 \int_{\partial \Omega} \left| \mathcal{B}_k(u_\theta, a_\theta)(x)-g(x) \right|^2 \dif\nu(x) 
+ \eta_2 \sum_{\ell=1}^L \left| u_\theta(x_\ell^o) - u_\ell^o \right|^2\biggr), \\[1em]
\text{s.t. } 
(u_\theta, a_\theta) \in \argmin_{(u,a)\in\mathcal{U}_{\theta_u}\times\mathcal{A}_{\theta_a}} \biggl(\|u\|_{\mathcal{U}_{\theta_u}}^2 + \|a\|_{\mathcal{A}_{\theta_a}}^2 + \frac{1}{\gamma^2} \sum_{\ell=1}^L |u(x_\ell^o) - u_\ell^o|^2\biggr)
 \\[0.5em]
\quad\quad\quad\quad\quad\quad\quad\text{s.t. } \
\mathcal{P}_k(u,a)(x_i) = 0,\;\; i=1,\dots,M_\Omega,\\ \quad\quad\quad\quad\quad\quad\quad\quad\quad\quad\quad\quad
\mathcal{B}_k(u,a)(x_j) = 0,\;\; j=M_\Omega+1,\dots,M.
\end{dcases}
\end{equation}

Define the residual vector \( \mathbf{r}^k \in \mathbb{R}^M \) as the collection of collocation residuals defined entrywise by
\begin{align*}
\mathbf{r}^k_i \coloneqq 
\begin{cases}
    D_u\mathcal{P}(u^k, a^k)(u^k)(x_i) + D_a\mathcal{P}(u^k, a^k)(a^k)(x_i) -\mathcal{P}(u^k,a^k)(x_i)&\qif i\in\{1,\ldots, M_\Omega\},\\
    D_u\mathcal{B}(u^k, a^k)(u^k)(x_i) + D_a\mathcal{B}(u^k, a^k)(a^k)(x_i) -\mathcal{B}(u^k,a^k)(x_i)&\qif i\in \{M_\Omega+1, \ldots, M\}.
\end{cases}
\end{align*}
Let \( \phi_i \) and \( \psi_j \) denote the linear functionals \( \delta_{x_i} \circ D_u\mathcal{P}(u^k,a^k) \) and \( \delta_{x_j} \circ D_a\mathcal{P}(u^k,a^k) \), respectively, and similarly for \( \mathcal{B} \). The linear operators \( \Phi \colon \mathcal{U}_{\theta_u} \to \mathbb{R}^M \) and \( \Psi\colon \mathcal{A}_{\theta_a} \to \mathbb{R}^M \) are constructed by evaluating the Fr\'echet derivatives of the PDE and boundary operators at collocation points; entrywise, they are given by the linear functionals
\[
\Phi_m \coloneqq  \phi_m\qa \Psi_m \coloneqq  \psi_m. 
\]
Thus, the inner problem in \eqref{eq:inv:surrogate_bilevel} becomes the following linearly constrained quadratic minimization problem in the product RKHS \( \mathcal{U}_{\theta_u} \times \mathcal{A}_{\theta_a} \):
\begin{equation}
\label{eq:inv:inner}
\min_{(u,a)\in \mathcal{U}_{\theta_u} \times \mathcal{A}_{\theta_a}} \Bigl\{
\|u\|_{\mathcal{U}_{\theta_u}}^2 + \|a\|_{\mathcal{A}_{\theta_a}}^2
\quad \text{s.t.} \quad \Phi u + \Psi a = \mathbf{r}^k\Bigr\}. 
\end{equation}
By the representer theorem \cite[Sec.~17.8]{owhadi2019operator} for linearly constrained RKHS problems, the minimizers admit the forms
\begin{align}
u_\theta(x) &= \sum_{m=1}^{M} (z_u)_m \, \kappa_u(x, \phi_m) = \kappa_u(x, \Phi)^\top \mathbf{z}_u\qa  \label{eq:inv:rep:u} \\
a_\theta(x) &= \sum_{m=1}^{M} (z_a)_m \, \kappa_a(x, \psi_m) = \kappa_a(x, \Psi)^\top \mathbf{z}_a
\label{eq:inv:rep:a}
\end{align}
for every $x$, where \( \kappa_u(x, \phi) \coloneqq  \phi (\kappa_u(x, \cdot)) \) denotes the action of the linear functional \( \phi \) on the second kernel argument, and similarly for \( \kappa_a \). The vectors \( \mathbf{z}_u\in \mathbb{R}^M \) and \(\mathbf{z}_a \in \mathbb{R}^M \) are unknown coefficients.

Substituting these representations into the constraint \( \Phi u_\theta + \Psi a_\theta = \mathbf{r}^k \), we obtain the finite-dimensional constraint
\begin{equation}
\label{eq:inv:constraint}
\mathbf{z}_u +\mathbf{z}_a = \mathbf{r}^k,
\end{equation}
and the regularized RKHS norms become
\[
\|u_\theta\|_{\mathcal{U}_{\theta_u}}^2 + \|a_\theta\|_{\mathcal{A}_{\theta_a}}^2 = \mathbf{z}_u^\top \mathbf{K}_{\Phi}^{-1} \mathbf{z}_u + \mathbf{z}_a^\top \mathbf{K}_{\Psi}^{-1} \mathbf{z}_a.
\]
In the preceding displays, we used the compact kernel notation from Subsection~\ref{subsec:OTD:general}.
We solve this constrained minimization via Lagrange multipliers and obtain  \( \mathbf{z}_u \) and \( \mathbf{z}_a \), which can be substituted into \eqref{eq:inv:rep:u}-\eqref{eq:inv:rep:a} to recover the updated GP approximations \( u_\theta \) and \( a_\theta \).

To learn the hyperparameters \(\theta=(\theta_u,\theta_a)\), for each iteration $k$ we evaluate the outer loss on a separate, independently sampled validation set consisting of interior points \(\{x_v^{(i,k)}\}_{i=1}^{N_s^{(k)}}\subset\Omega\) and boundary points \(\{x_b^{(j,k)}\}_{j=1}^{N_b^{(k)}}\subset\partial\Omega\), and compute the corresponding empirical loss
\begin{align}
\begin{split}
{\mathcal{J}}_k(\theta) 
\coloneqq \frac{1}{N_s^{(k)}} \sum_{i=1}^{N_s^{(k)}} \left| \mathcal{P}(u_\theta,a_\theta)(x_v^{(i,k)}) -f(x_v^{(i,k)})\right|^2
&+ \frac{\eta_1}{N_b^{(k)}} \sum_{j=1}^{N_b^{(k)}} \left| \mathcal{B}(u_\theta,a_\theta)(x_b^{(j,k)})-g(x_b^{(j,k)}) \right|^2\\
&\qquad\quad + \eta_2 \sum_{\ell=1}^L \left| u_\theta(x_\ell^o) - u_\ell^o \right|^2.
\end{split}
\end{align}

This loss depends on \( \theta \) through kernel evaluations and their derivatives. Its gradient \( \nabla_\theta {\mathcal{J}}_k \) is computed via automatic differentiation, and \( \theta \) is updated using first-order methods such as Adam. The procedure is repeated until convergence of both \( \theta \) and the associated GP solutions \( u_\theta\) and \(a_\theta \).

\subsubsection{Discretize-Then-Optimize}
The DTO method offers an alternative to the OTD strategy and parallels the DTO formulation for forward problems described in Section~\ref{subsec:motivating_example}. While OTD performs a functional linearization of the PDE and boundary operators prior to discretizing the outer loss, DTO first fixes a finite set of validation points and subsequently applies linearization directly at these discrete locations during each optimization iteration.

We begin by selecting  interior validation points \( \{x_v^{(i)}\}_{i=1}^{N_s} \subset \Omega \) and boundary validation points \( \{x_b^{(j)}\}_{j=1}^{N_b} \subset \partial \Omega \). These are used to define a discrete outer loss
\begin{equation}
\label{eq:inv:dto:loss}
\frac{1}{N_s} \sum_{i=1}^{N_s} \left| \mathcal{P}(u_\theta,a_\theta)(x_v^{(i)}) - f(x_v^{(i)})\right|^2
+ \frac{\eta_1}{N_b} \sum_{j=1}^{N_b} \left| \mathcal{B}(u_\theta,a_\theta)(x_b^{(j)})-g(x_b^{(j)}) \right|^2
+ \eta_2 \sum_{\ell=1}^L \left| u_\theta(x_\ell^o) - u_\ell^o \right|^2.
\end{equation}

To evaluate the residuals appearing in this loss, we proceed by linearizing the nonlinear operators $\mathcal{P}$ and $ \mathcal{B}$ at each iteration. Given the current estimate \( (u^k, a^k) \), we apply first-order expansions of the PDE and boundary operators at the chosen validation points. Specifically, for each interior point \( x_v \in \Omega \), we write
\[
\mathcal{P}(u,a)(x_v) \approx \mathcal{P}(u^k,a^k)(x_v) 
+ D_u\mathcal{P}(u^k,a^k)(u - u^k)(x_v)
+ D_a\mathcal{P}(u^k,a^k)(a - a^k)(x_v),
\]
and analogously for the boundary conditions at \( x_b \in \partial \Omega \). This linearization yields a set of affine constraints evaluated at the validation points, which depend linearly on the candidate functions \( u \) and \( a \). These constraints are used to formulate the inner minimization problem, whose solution is computed at each iteration.

The resulting inner problem is a linearly constrained quadratic program over the product RKHS \( \mathcal{U}_{\theta_u} \times \mathcal{A}_{\theta_a} \), identical in structure to that solved in the OTD formulation. It admits closed-form solutions for \( u_\theta \) and \( a_\theta \). These explicit expressions are then substituted into the linearized residual terms derived from~\eqref{eq:inv:dto:loss}, resulting in a fully differentiable surrogate objective that depends on the hyperparameters \( \theta = (\theta_u, \theta_a) \) solely through kernel evaluations and their derivatives. Gradients of this linearized loss are computed via automatic differentiation. Hyperparameters are then updated using a first-order optimization method, such as Adam. This procedure is repeated iteratively until convergence.

\subsection{Darcy Flow}
\label{subsec:darcy_flow_example}
We now instantiate the framework from the preceding subsection in the context of a two-dimensional Darcy flow inverse problem on the unit square. For a given log-permeability field $a$, defined on $\Omega\coloneqq(0,1)^2$, the forward problem is to find pressure $u$ on $\Omega$, which for simplicity we assume satisfies homogeneous Dirichlet boundary conditions on $\partial \Omega$, given by
\begin{align}
\label{eq:intro:darcy:pde}
    \begin{cases}
        \begin{alignedat}{2}
            -\nabla \cdot \bigl(\exp(a(x)) \nabla u(x)\bigr) &=f(x)  && \quad  \forall x \in \Omega,\\
            u(x)&=0 && \quad \forall x \in \partial\Omega.
        \end{alignedat}
    \end{cases}
\end{align}
We let $u^\star$ denote the solution of the forward PDE problem when the log-permeability is $a^\star$.
The inverse problem is to find $u^\star\colon \Omega\to\mathbb{R}$ and $a^\star\colon \Omega\to\mathbb{R}$ given only noisy pointwise measurements of the solution $u^\star$  at fixed locations $\{x_\ell^o\}_{\ell=1}^L\subset\Omega$:
\begin{align}\label{eq:darcy_data}
    u_\ell^o \;=\; u^\star(x_\ell^o)+\xi_\ell,\qquad \xi_\ell\sim\mathcal{N}(0,\gamma^2)\;\text{ i.i.d.}\,.
\end{align}
The forcing $f$ is assumed known. The goal is to reconstruct the pair $(u^\star,a^\star)$ from the data~\eqref{eq:darcy_data} while enforcing the PDE~\eqref{eq:intro:darcy:pde}.

To handle this inverse problem, we must work in the joint solution and coefficient function space. We place independent zero-mean GP priors on $u$ and $a$ with kernels $\kappa_u$ and $\kappa_a$ and hyperparameters $\theta=(\theta_u,\theta_a)\in\Theta$, inducing RKHSs $\mathcal{U}_{\theta_u}$ and $\mathcal{A}_{\theta_a}$. Given $\theta$, we seek $(u_\theta,a_\theta)$ that are both simple (small RKHS norms), fit the $u$-data~\eqref{eq:darcy_data}, and satisfy the PDE and boundary conditions at collocation points $\{x_i\}_{i=1}^{M_\Omega}\subset\Omega$ and $\{x_j\}_{j=M_\Omega+1}^{M}\subset\partial\Omega$:
\begin{equation}
\label{eq:intro:darcy:inner}
(u_\theta,a_\theta)\in\argmin_{(u,a)\in\mathcal{U}_{\theta_u}\times\mathcal{A}_{\theta_a}}
\biggl(\|u\|_{\mathcal{U}_{\theta_u}}^2+\|a\|_{\mathcal{A}_{\theta_a}}^2+\frac{1}{\gamma^2}\sum_{\ell=1}^L|u(x_\ell^o)-u_\ell^o|^2\biggr)
\quad\text{s.t.}\quad
\begin{cases}
-\nabla\!\cdot(\exp(a)\nabla u)(x_i)=f(x_i),\\
u(x_j)=0.
\end{cases}
\end{equation}
Hyperparameters are learned by minimizing a validation loss that penalizes PDE and boundary residuals and the data misfit:
\begin{equation}
\label{eq:intro:darcy:bilevel}
\min_{\theta\in\Theta}\;
\biggl(
\int_\Omega \bigl|-\nabla\!\cdot(\exp(a_\theta)\nabla u_\theta)(x)-f(x)\bigr|^2\dif \mu(x)
+\eta_1\!\int_{\partial\Omega}\!|u_\theta(x)|^2\,\dif\nu(x) 
+\eta_2\sum_{\ell=1}^L |u_\theta(x_\ell^o)-u_\ell^o|^2\biggr),
\end{equation}
where $\eta_1\geq 0$ and $\eta_2\geq 0$ balance boundary enforcement and data fit.

Compared with the bilevel framework for solving PDEs discussed above, the inverse Darcy setting introduces an additional unknown field $a$. Consequently, the inner problem regularizes both $u$ and $a$ and includes a $u$-data misfit to ensure identifiability, while the outer loss balances physics residuals and data fidelity to avoid PDE-consistent yet observation-inconsistent solutions.

For the bilevel problems \eqref{eq:intro:darcy:bilevel}--\eqref{eq:intro:darcy:inner}, we apply OTD and DTO following discussions in the previous subsection. In OTD, we linearize the explicit Darcy equations \eqref{eq:intro:darcy:pde} at $(u^k,a^k)$, yielding linear equality constraints. The inner problem is a linearly constrained quadratic program on $\mathcal{U}_{\theta_u}\times\mathcal{A}_{\theta_a}$ which, by the representer theorem, admits a closed-form solution. We evaluate the linearized surrogate of the outer loss \eqref{eq:intro:darcy:bilevel} at freshly sampled validation points and update $\theta$ with a first-order method via automatic differentiation. DTO mirrors this workflow: we fix interior and boundary validation sets, build a discrete outer loss from the Darcy residual and boundary conditions at those points, and at each iteration linearize these discrete residuals around $(u^k,a^k)$. The resulting inner problem has the same closed-form  as in OTD, enabling efficient solution and differentiation with respect to $\theta$, with mini--batching of validation points and observations to reduce per--iteration cost. In the following section we turn our attention to numerical results, including for this inverse problem.

\section{Numerical Results}
\label{sec:numerical_results}

In this section, we conduct a series of numerical experiments spanning a diverse collection of PDE problems. These examples are selected to systematically test and demonstrate several key aspects of the proposed bilevel method. We begin with the nonlinear elliptic equation, introduced in Subsection
\ref{subsec:motivating_example}, in order to evaluate the method's effectiveness and robustness for both low-dimensional (single-parameter) and moderate-dimensional (multi-parameter) kernel learning tasks (Subsection~\ref{sec:num_ellip}). Next, we consider the complex-valued nonlinear Schr\"odinger equation to highlight the method's capability in handling multi-component PDE systems with multiple hyperparameters, as well as its robustness to different initializations (Subsection~\ref{sec:num_schro}). The Gray-Scott reaction-diffusion system,
introduced in Subsection \ref{subsec:grayscott_example}, serves to illustrate the necessity of hyperparameter learning in coupled PDE systems and further tests the reusability of learned hyperparameters across different initial conditions, suggesting its potential for generalization (Subsection~\ref{sec:num_gray}). To examine scalability in more expressive kernel classes, we solve the Eikonal and Burgers' equations using non-stationary Gibbs kernels whose spatially and/or temporally varying lengthscales are parameterized by neural networks (Subsections~\ref{sec:num_eik} and~\ref{sec:num_burg}). These examples demonstrate the method's ability to handle high-dimensional hyperparameter spaces and yield interpretable kernels that align with the underlying solution structure. Finally, we apply the method to the inverse problem for Darcy flow introduced
in Subsection \ref{subsec:darcy_flow_example}, where the goal is to infer both the solution and an unknown coefficient field from noisy observations (Subsection~\ref{sec:num_darcy}). Together, these examples validate the effectiveness, robustness, scalability, and flexibility of the proposed approach in a wide range of kernel-based PDE learning settings. The implementations of our numerical examples are publicly available online.~\footnote{\url{https://github.com/yangx0e/BilevelHyperparameterLearning.git}}

\subsection{Nonlinear Elliptic Equation}\label{sec:num_ellip}

To illustrate the effectiveness and robustness of the proposed method for hyperparameter learning, we begin with a benchmark nonlinear elliptic PDE defined as
\begin{subequations}\label{eq:elliptic}
    \begin{align}
        -\Delta u(x, y) + \alpha u^m(x, y) =& f(x, y), \ \forall (x, y)\in\Omega,\\
        u(x, y) =& g(x, y), \ \forall(x, y)\in\partial\Omega,
    \end{align}
\end{subequations}
where $\Omega \coloneqq (0, 1)^2$. In this example, we set $\alpha=1, m=3$ and $g(x, y) = 0$ for all $(x, y)\in\partial\Omega$. The true solution is prescribed as $u(x, y)=\sin(\pi x) \sin(\pi y) + 4 \sin(4\pi x) \sin(4\pi y)$ and the corresponding source term $f$ is computed by substituting this prescribed solution into \eqref{eq:elliptic}. We consider the following two cases for learning kernel hyperparameters:
\begin{itemize}
\item \textbf{Case (A)}: An isotropic Gaussian kernel with a single hyperparameter---the shared lengthscale---that requires optimization.
\item \textbf{Case (B)}: An additive kernel composed of both radial basis function (RBF) and second-order polynomial terms, with four hyperparameters to be learned.
\end{itemize}
These two cases are designed to demonstrate both the robustness (i.e., insensitivity to initialization) and the effectiveness (i.e., ability to yield accurate PDE solutions) of the proposed learning strategy, even in more complex, multi-parameter kernel settings where poor initialization may otherwise prevent convergence \cite{shao2025solving}.


\subsubsection{\textbf{Case (A)}: Isotropic Gaussian Kernel}
We begin by evaluating a two-dimensional isotropic Gaussian kernel within the GP-PDE framework \cite{chen2021solving}. Specifically, we use  
\begin{align}
\label{eq:rbf:kernel}
\kappa(\mathbf{x},\mathbf{x}') \coloneqq  \exp\biggl(-\frac{\|\mathbf{x}-\mathbf{x}'\|^2}{2 l^2}\biggr), \qw \mathbf{x}\coloneqq (x, y),
\end{align}  
and the single hyperparameter \(l\) (the lengthscale) governs both spatial dimensions and is estimated with our bilevel procedure.  
Although this is a relatively simple test case---wherein manual tuning (e.g., via grid or random search) is often sufficient---it serves as an effective testbed to evaluate both the robustness and convergence behavior of the proposed method. We adopt the DTO scheme discussed in Subsection~\ref{subsec:DTO:general}. 
At initialization we sample, and then keep fixed for all iterations, a collocation set of $900$ interior points in $\Omega$ and $300$ boundary points on $\partial\Omega$. 
We also draw a disjoint validation set of $900$ interior points for the outer objective and the boundary condition is  excluded from the validation loss. 
At each GN step, the linearized PDE is solved on the fixed collocation set, while the hyperparameter update uses a mini-batch objective formed by uniformly subsampling $200$ points from the fixed validation set.
 The Adam optimizer \cite{kingma2014adam} with learning rate $1\times10^{-2}$ is employed. An identity matrix scaled by factor $10^{-10}$ is added to the Gram matrix to ensure numerical stability (see also Appendix~A of \cite{chen2021solving} for a discussion of this regularization, often referred to as a ``nugget'' in the statistics
 and machine learning community \cite{rasmussen2006gaussian}.)
 We perform $30$ GN iterations, each with $50$ learning iterations.

\begin{figure}[tb]
    \centering
    \includegraphics[width=0.49\linewidth]{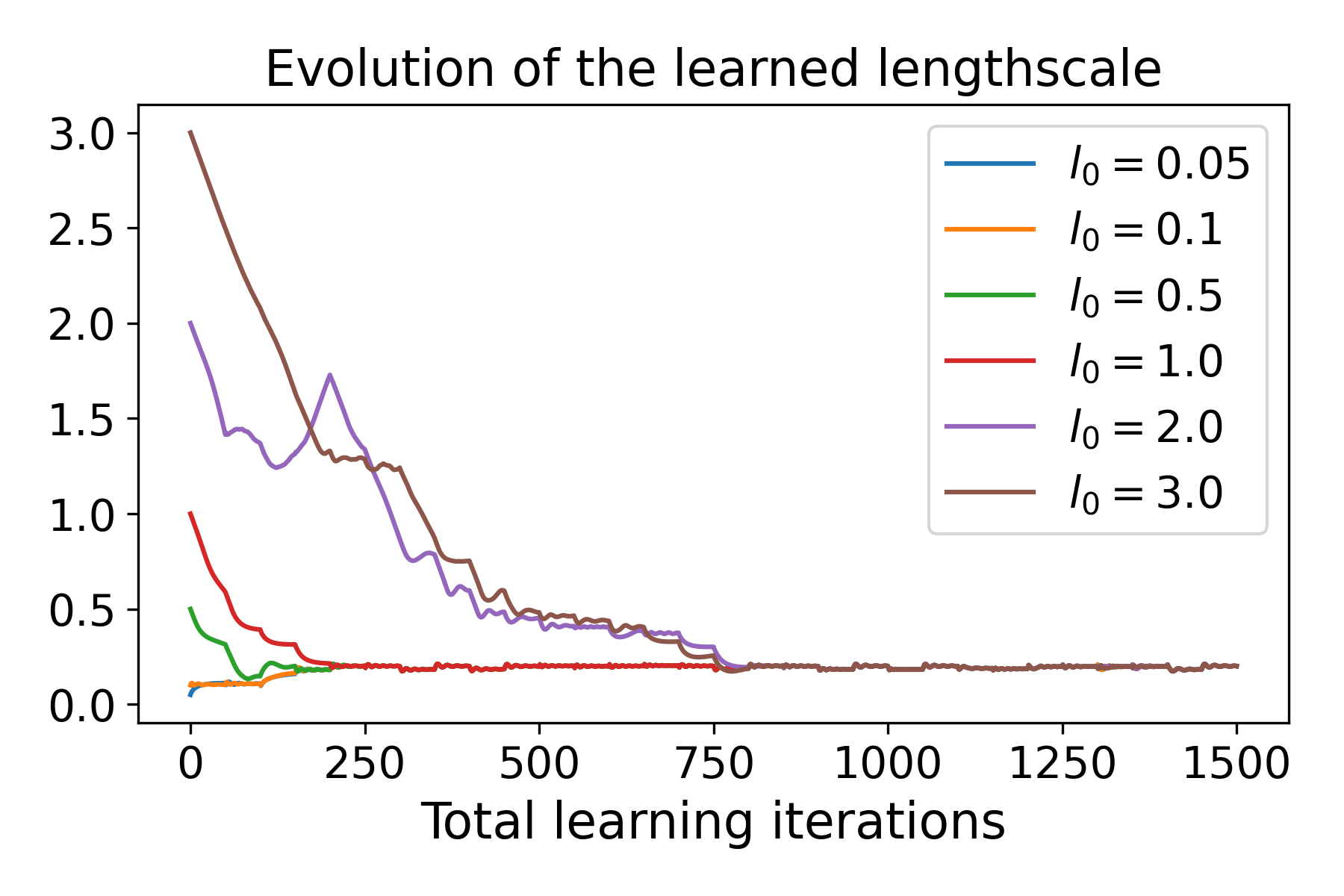}
    \includegraphics[width=0.45\linewidth]{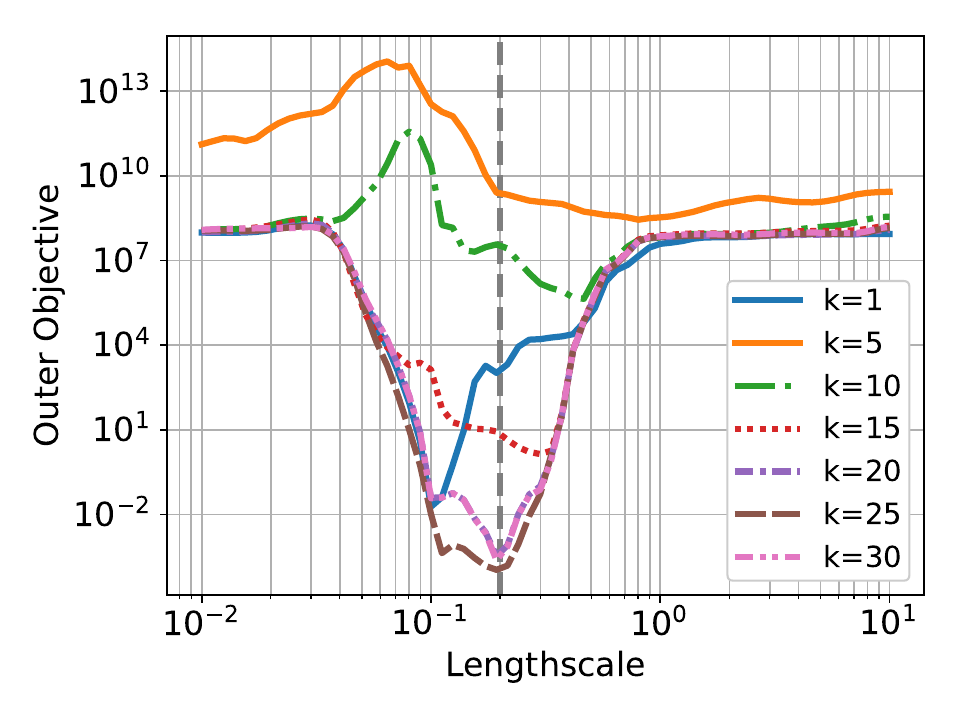}
    \caption{(Left) Evolution of the learned lengthscale in the Gaussian kernel for solving the nonlinear elliptic equation \eqref{eq:elliptic}, initialized from various starting values $l_0$. All trajectories converge to similar final values, indicating robustness of the learning procedure. (Right) Visualization of the landscape of the linearized outer DTO loss versus lengthscale $\theta=l$ of \eqref{eq:rbf:kernel}. As the outer iteration index $k$ increases, the minimum value of the loss begins to stabilize near $l=0.2$, which is represented by the vertical gray dashed line.}
    \label{fig:elliptic_1}
\end{figure}

The left display of Figure~\ref{fig:elliptic_1} shows the evolution of the learned lengthscale initialized from various starting values $l_0 = 0.05, 0.1, 0.5, 1.0, 2.0, 3.0$. Despite differing initializations, the learned lengthscale consistently converges to approximately the same value ($l = 0.2005, 0.2005, 0.2007, 0.2006, 0.2005$, respectively), demonstrating strong robustness. Larger deviations from the optimal value result in slower convergence but ultimately reach the same optimum. Fixing the initialization to be $l_0=2.0$, the right display of Figure~\ref{fig:elliptic_1} plots the linearized outer objective in \eqref{eq:dto:linearized-loss} as a function of $l$ as the outer GN iteration index $k$ ranges from $k=1$ to $k=30$. For each $k$, the loss landscape is relatively well behaved. For sufficiently many iterations (i.e., more than $10$), the minimal value of the loss tends toward $0.2$; this is consistent with the left display of Figure~\ref{fig:elliptic_1}.
To verify effectiveness, we use the learned hyperparameters to solve \eqref{eq:elliptic} using the GP-PDE method \cite{chen2021solving} from scratch, employing all $1,800$ collocation points and $300$ boundary points (which are used in the learning of the lengthscale). We reduce the nugget to $1 \times 10^{-12}$ and perform $10$ GN iterations. The resulting errors are reported in Table~\ref{tab:elliptic}, confirming the accuracy of the learned hyperparameter in solving \eqref{eq:elliptic}.

\begin{table}[tb]
    \footnotesize
    \centering
    \caption{Errors of solutions obtained from using the learned lengthscale initialized from different starting values $l_0$ to solve the nonlinear elliptic equation \eqref{eq:elliptic} with the GP-PDE method \cite{chen2021solving} based on $1,800$ collocation points and $300$ boundary points generated for the learning. $l=0.2005$ is obtained from initial values $l_0=0.05, 0.1, 3.0$, $l=0.2006$ is obtained from initial value $l_0=2.0$, and $l=0.2007$ is obtained from initial value $l=1.0$.} 
    \label{tab:elliptic}
    \begin{tabular}{rccc}
        \toprule
        & $l=0.2005$ & $l=0.2006$ & $l=0.2007$\\ 
        \midrule
        $L^2$ error & $2.20\times10^{-7}$ & $2.21\times 10^{-7}$ & $2.21\times 10^{-7}$\\ 
        $L^\infty$ error & $4.21\times 10^{-6}$  & $4.23\times 10^{-6}$ & $4.24\times 10^{-6}$\\ 
        \bottomrule
    \end{tabular}
\end{table}

\subsubsection{\textbf{Case (B)}: An Additive Kernel with Multiple Hyperparameters}

We now explore a more complex scenario involving an additive kernel with multiple hyperparameters. This kernel is given by
\begin{equation}\nonumber
    \kappa(\mathbf{x}, \mathbf{x}^\prime) \coloneqq \sigma^2 \exp\biggl(-\frac{\|\mathbf{x} - \mathbf{x}^\prime\|^2}{2l^2}\biggr) + \bigl(c + \alpha\mathbf{x}^\top\mathbf{x}^\prime\bigr)^2, \quad \mathbf{x}\coloneqq (x, y), 
\end{equation}
where $\sigma$, $l$, $c$, and $\alpha$ are four learnable hyperparameters of the kernel. This setting reflects practical situations where kernel design involves multiple interacting terms, and hyperparameter tuning becomes significantly more challenging due to the higher-dimensional search space. The same sampling strategy and numerical settings from \textbf{Case (A)} are used here, and we initialize all hyperparameters to $1.0$.

\begin{table}[tb]
    \footnotesize
    \centering
    \caption{Errors of the solution obtained from using the learned additive kernel to solve the nonlinear elliptic equation \eqref{eq:elliptic} with the GP-PDE method \cite{chen2021solving} based on $1,800$ collocation points and $300$ boundary points generated in the hyperparameter learning process. We set nugget to $1\times10^{-10}$ for numerical stability. When the additive kernel with the learned hyperparameters is employed, the GP-PDE method converges within 10 GN iterations. In contrast, it fails to converge even after 200 GN iterations when fixing the hyperparameters at the values used as an initial guess in the learned
    hyperparameter approach (and hence we write ``N/A'' for the errors).} 
    \label{tab:elliptic_2}
        \begin{tabular}{rcc}
    \toprule
      & The learned additive kernel & $\sigma=l=c=\alpha=1.0$ \\ 
     \midrule
     $L^2$ error & $7.49\times 10^{-7}$ & N/A\\ 
     $L^\infty$ error & $1.27\times 10^{-5}$  & N/A\\ 
     \bottomrule
    \end{tabular}
\end{table}

Table~\ref{tab:elliptic_2} compares the solution errors obtained when using the learned additive kernel with errors obtained when the hyperparameters are simply fixed
at an arbitrary point. When the GP-PDE method \cite{chen2021solving} is applied with the learned hyperparameters, it converges within $10$ GN iterations and achieves high accuracy. In contrast, using the fixed hyperparameter values $\sigma = l = c = \alpha = 1.0$ leads to a failure to converge even after $200$ GN iterations. While such divergence can sometimes be mitigated through alternative strategies---such as better preconditioning of the initial guess \cite{chen2021solving} or adopting more robust optimization algorithms like the Levenberg-Marquardt method \cite{jalalian2025data}---our intention here is not to rule out those techniques. Rather, this example underscores a key message: when hyperparameters are poorly chosen or left untrained, the GP-PDE method 
\cite{chen2021solving} may struggle to converge or yield inaccurate results. Hence, systematic hyperparameter learning is not only beneficial but often necessary to ensure the reliability and performance of kernel-based PDE solvers.

\subsection{Complex-Valued Schr\"{o}dinger Equation}\label{sec:num_schro}
In this example, we consider the complex-valued nonlinear Schr\"{o}dinger equation on unit interval $(0,1)$ in time and on the circle $\mathbb{S}\coloneqq[-5,5)$ (that is with periodic boundary conditions in space) given by
\begin{equation}\label{eq:schrodinger}
    \mathrm{i} \frac{\partial h}{\partial t} + \frac{1}{2} \frac{\partial^2 h}{\partial x^2} + g |h|^2 h = 0, \ \forall (t, x) \in (0, 1) \times \mathbb{S}.
\end{equation}
Here $g\equiv 1$ defines a focusing nonlinearity. The initial condition is specified as 
\begin{equation}
    h(0, x) \coloneqq \frac{2}{\cosh(x)}, \ \forall x\in (-5, 5).
\end{equation}
To approximate the time-dependent, complex-valued solution $h$ by GP-PDE methods it is possible use a time-stepper such as backward Euler and then use GP-PDE to solve the
resulting PDE at each time-step; alternatively, as introduced in \cite{chen2021solving},
it is possible to use space-time GPs and, although this does not enforce causality
in time, it can nonetheless be effective. We choose this second, non-causal, approach;
we represent the real and imaginary components of $h$ using two independent GPs, each equipped with a periodic RBF kernel
\begin{equation}\nonumber
    \kappa(t, x, t^\prime, x^\prime) = \exp\left(-\frac{|t - t^\prime|^2}{2l_t^2}\right) \exp\left(-\frac{2\sin^2\bigl(\frac{\pi}{p}(x-x^\prime)\bigr)}{l_x^2}\right)
\end{equation}
with $p=10$. Since each GP has its own kernel, this setup introduces four kernel hyperparameters: the temporal and spatial lengthscales for the real part ($l_t^u$, $l_x^u$) and the imaginary part ($l_t^v$, $l_x^v$). 

\begin{figure}[tb]
    \centering
    \includegraphics[width=1.0\linewidth]{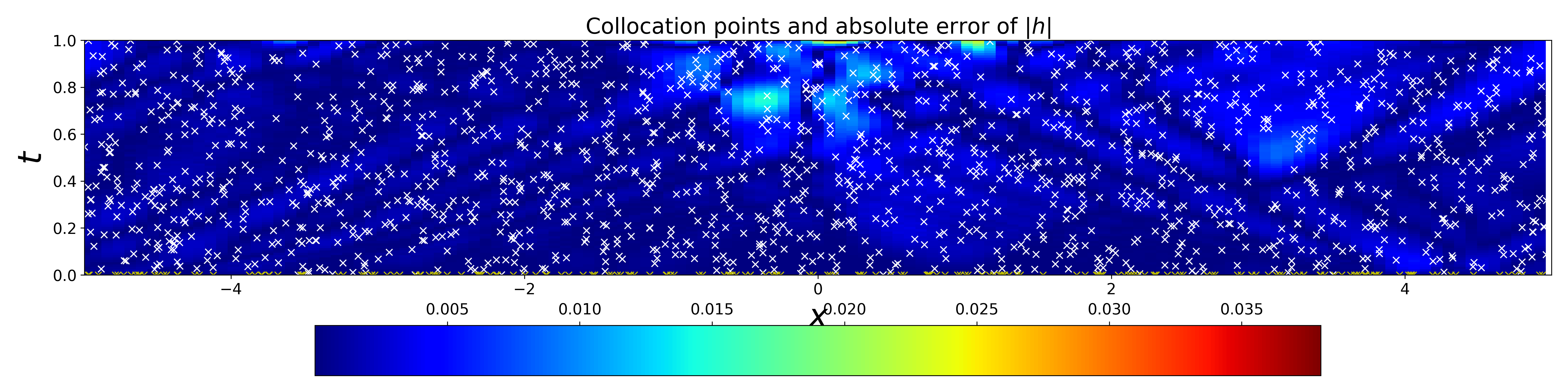}
    \includegraphics[width=0.3\linewidth]{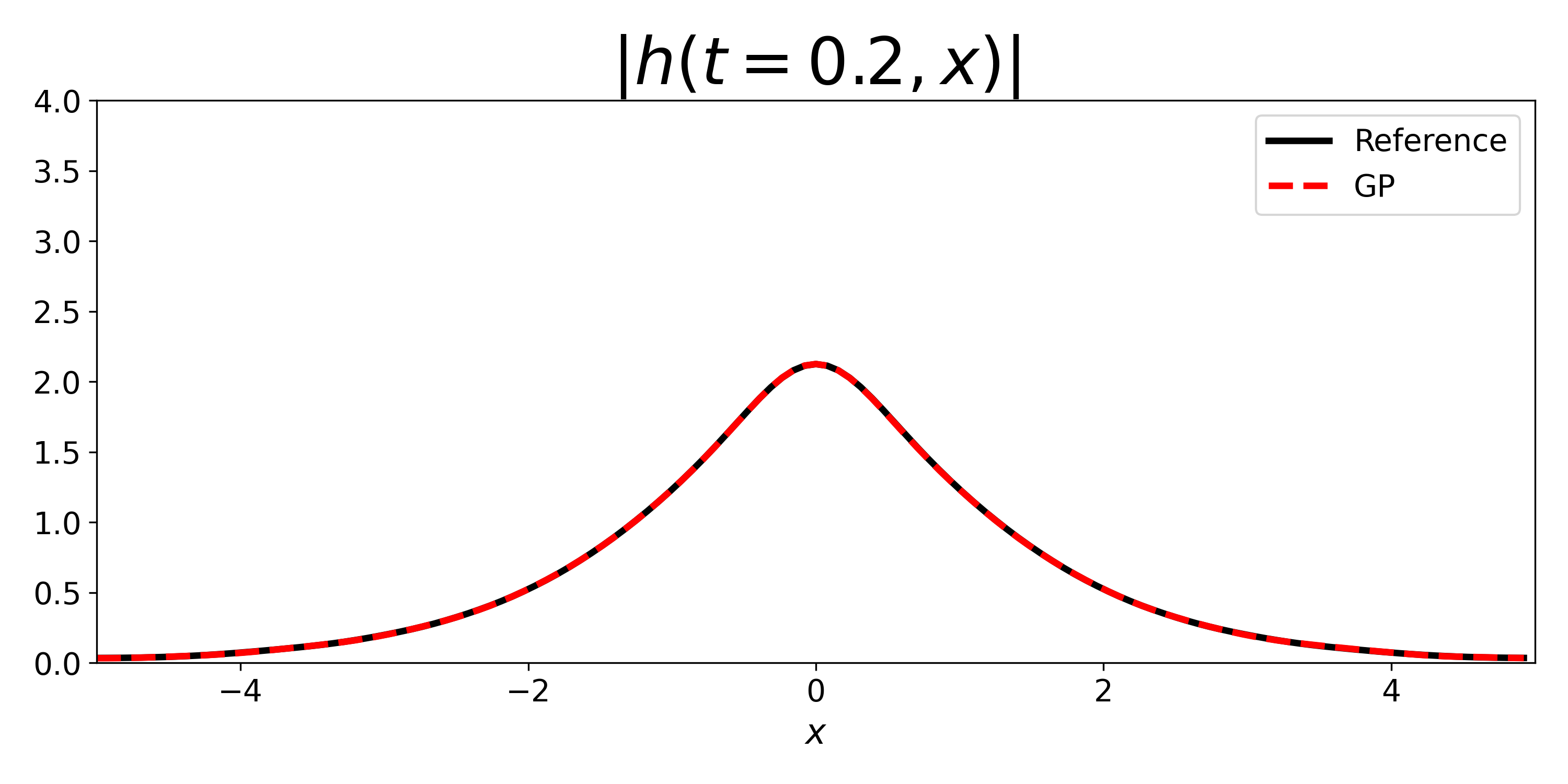}
    \includegraphics[width=0.3\linewidth]{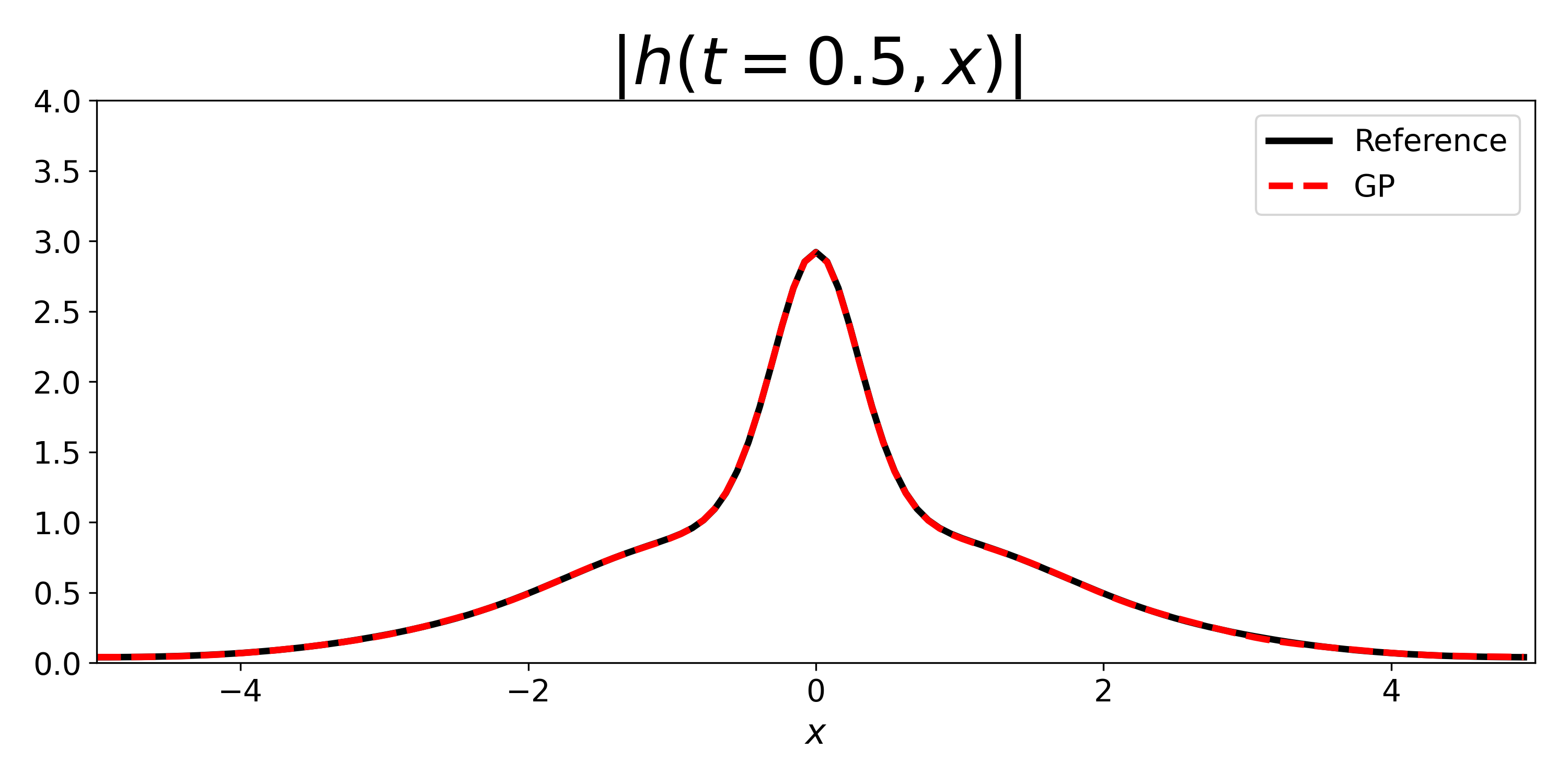}
    \includegraphics[width=0.3\linewidth]{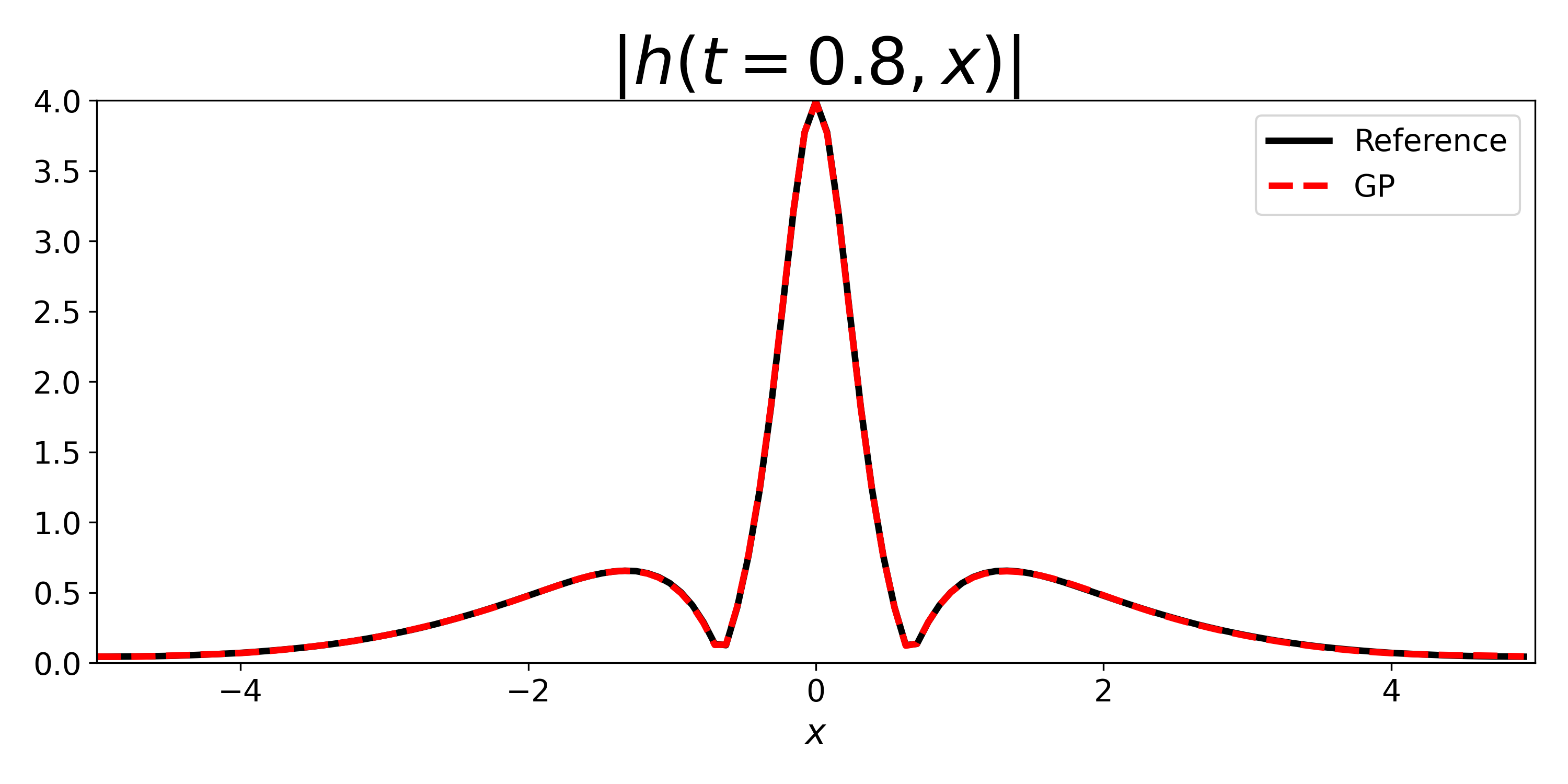}
    \caption{Solving the complex-valued nonlinear Schr\"{o}dinger equation with the GP-PDE method using the learned hyperparameters. The white and yellow ``x'' in the top figure represent the locations of the collocation and boundary points, respectively.}
    \label{fig:schrodinger}
\end{figure}

We use the GP-PDE method to solve \eqref{eq:schrodinger}, while simultaneously optimizing the hyperparameters via the DTO scheme outlined in Subsection~\ref{subsec:DTO:general}. Specifically, we initialize all four lengthscales with the same values---$1.0$, $0.5$, or $0.2$---and apply the learning algorithm in these three cases. We use the DTO scheme for this experiment. At initialization, we sample and then keep fixed a collocation set of 
$1,500$ interior points in $\Omega=(0,1)\times[-5,5)$ and 
$200$ boundary points along the initial time slice $t=0$ (since the spatial kernel is periodic, only the initial condition must be enforced). 
We also draw, once and for all, a separate validation set of $1,500$ interior points. 
Then, in each GN iteration the linearized PDE is solved on the fixed collocation set, 
while the hyperparameter update is driven by a mini-batch of $200$ points 
uniformly subsampled from the fixed validation set within each GN iteration. We use the Adam optimizer with learning rate $1\times10^{-2}$. A nugget term of $1\times 10^{-10}$ is added for numerical stability. The learning process consists of $30$ GN iterations, each with $50$ learning iterations.

\begin{table}[tb]
    \footnotesize
    \centering
    \caption{Errors of the solutions obtained from using the learned and initial hyperparameters to solve the complex-valued nonlinear Schr\"{o}dinger equation with the GP-PDE method.} 
    \label{tab:schrodinger}
        \begin{tabular}{ rcccc }
    \toprule
      & Learned hyperparameters & $l^u_t = l^u_x =l^v_t = l^v_x = 1.0$ & $l^u_t = l^u_x =l^v_t = l^v_x = 0.5$ & $l^u_t = l^u_x =l^v_t = l^v_x = 0.2$\\ 
     \midrule
     $L^2$ error & $0.0027$ & $0.2531$ & $0.1301$ & $0.0297$ \\ 
     $L^\infty$ error & $0.0380$ & $1.7438$  & $0.8957$ & $0.1820$ \\ 
     \bottomrule
    \end{tabular}
\end{table}

The resulting learned hyperparameters from the three different initializations ($1.0/0.5/0.2$) are:
\begin{subequations}\nonumber
    \begin{align}
        l^u_t &= 0.2461/0.2451/0.2493,\  l^u_x= 0.1455/0.1457/0.1450, \\
        l^v_t &= 0.2476/0.2470/0.2494, \ l^v_x = 0.1340/0.1342/0.1338,
    \end{align}
\end{subequations}
demonstrating the robustness of the proposed method with respect to initialization. We further test the learned hyperparameters by solving \eqref{eq:schrodinger} with the GP-PDE method from scratch using the learned hyperparameters and compare with using the initialized ones. Here, the learned hyperparameters are chosen to be $l^u_t = 0.2461$, $l^u_x = 0.1455$, $l^v_t = 0.2476$, and $l^v_x = 0.1340$. To evaluate the model, we solve the PDE using a new set of randomly sampled  $1,500$ interior collocation points and $200$ initial condition points (different from the ones used in the learning stage). We carry out 30 GN iterations in each experiment. Figure \ref{fig:schrodinger} visualizes the solution and the error, including the distributions of collocation and boundary points. In Table \ref{tab:schrodinger}, we compare the performance of the learned hyperparameters with several fixed (non-optimized) choices. The results highlight a clear trend: unlearned hyperparameters lead to substantial degradation in accuracy. In contrast, the learned values yield significantly lower $L^2$ and $L^\infty$ errors.

\subsection{Gray--Scott Reaction-Diffusion System}\label{sec:num_gray}
We consider the Gray--Scott model from Subsection~\ref{subsec:grayscott_example} 
with diffusion and reaction parameters set to $\mathsf{D}_u = 0.001$, $\mathsf{D}_v = 0.002$, $F = 0.04$, and $k = 0.06$. To solve this PDE system by GP-PDE,  we model $u$ and $v$ using two independent GPs, each equipped with the anisotropic RBF kernels in \eqref{eq:gray_scott:kernels}. 
This results in four hyperparameters to be learned: the temporal and spatial lengthscales for $u$, namely ($l_t^u$, $l_x^u$), and for $v$, namely ($l_t^v$, $l_x^v$). Solving systems of PDEs with multiple dependent variables naturally involves optimizing multiple GP kernels, each with its own set of hyperparameters.

To learn these hyperparameters, we initialize all lengthscales to $1.0$ and apply the DTO scheme in Subsection~\ref{subsec:DTO:general}.
At initialization we sample and then keep fixed a collocation set of 
$600$ interior points in $\Omega=(0,1)\times(0,1)$ and 
$400$ boundary points along $x=0$, $x=1$ (to enforce boundary conditions) and $t=0$ (to enforce the initial condition).  
We also draw once and keep fixed a separate validation set of $600$ interior points for hyperparameter learning.  
In each GN iteration, the linearized PDE is solved on the fixed collocation set,  
while the outer update subsamples a mini-batch of $200$ points uniformly from the fixed validation set.  The Adam optimizer is used with a learning rate of $1 \times 10^{-2}$, and a nugget term of $1 \times 10^{-10}$ ensures numerical stability. We perform $20$ GN iterations, each with $50$ learning steps.

\begin{figure}[tb]
    \centering
    \includegraphics[width=0.23\linewidth]{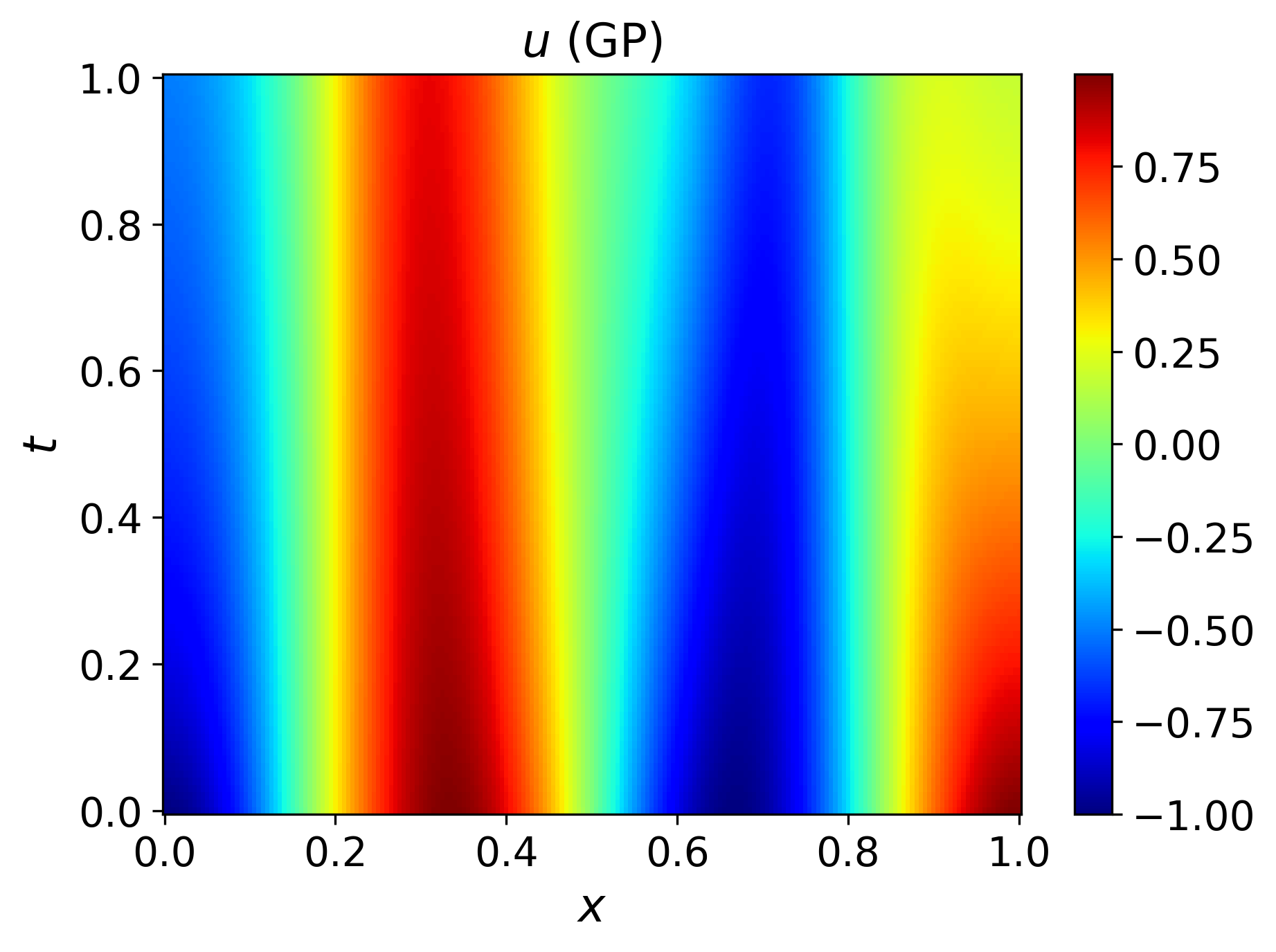}
    \includegraphics[width=0.23\linewidth]{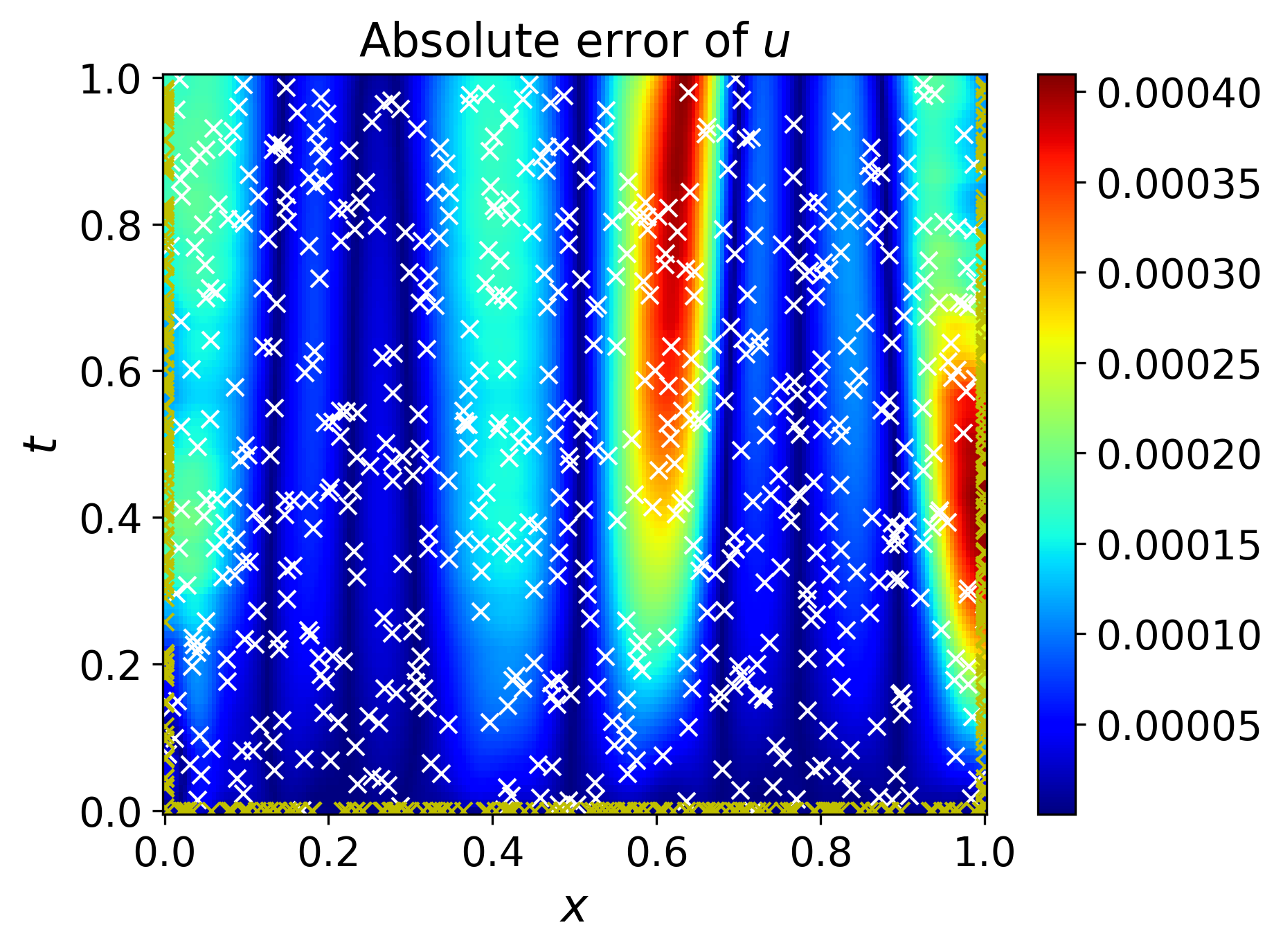}
    \includegraphics[width=0.23\linewidth]{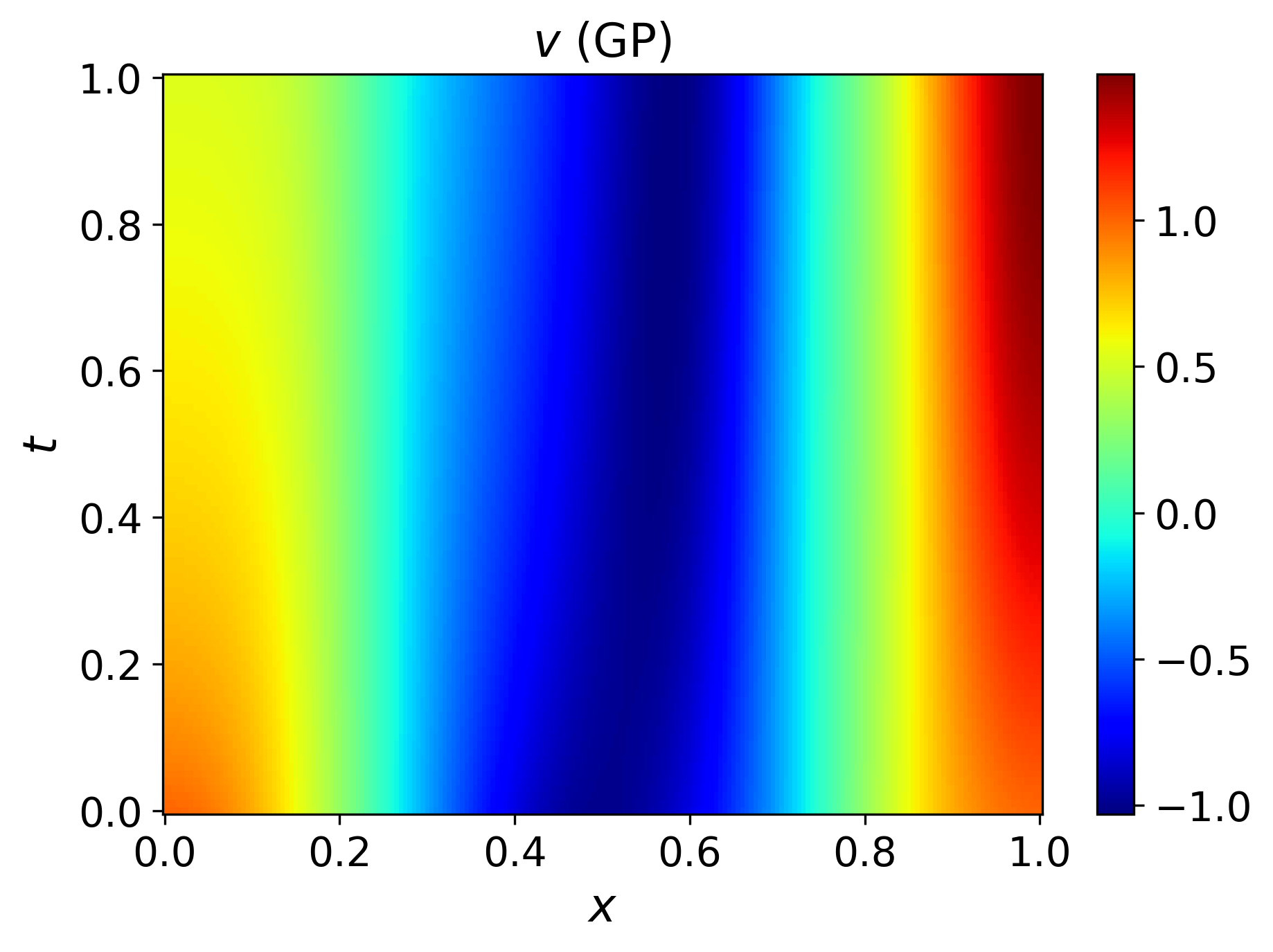}
    \includegraphics[width=0.23\linewidth]{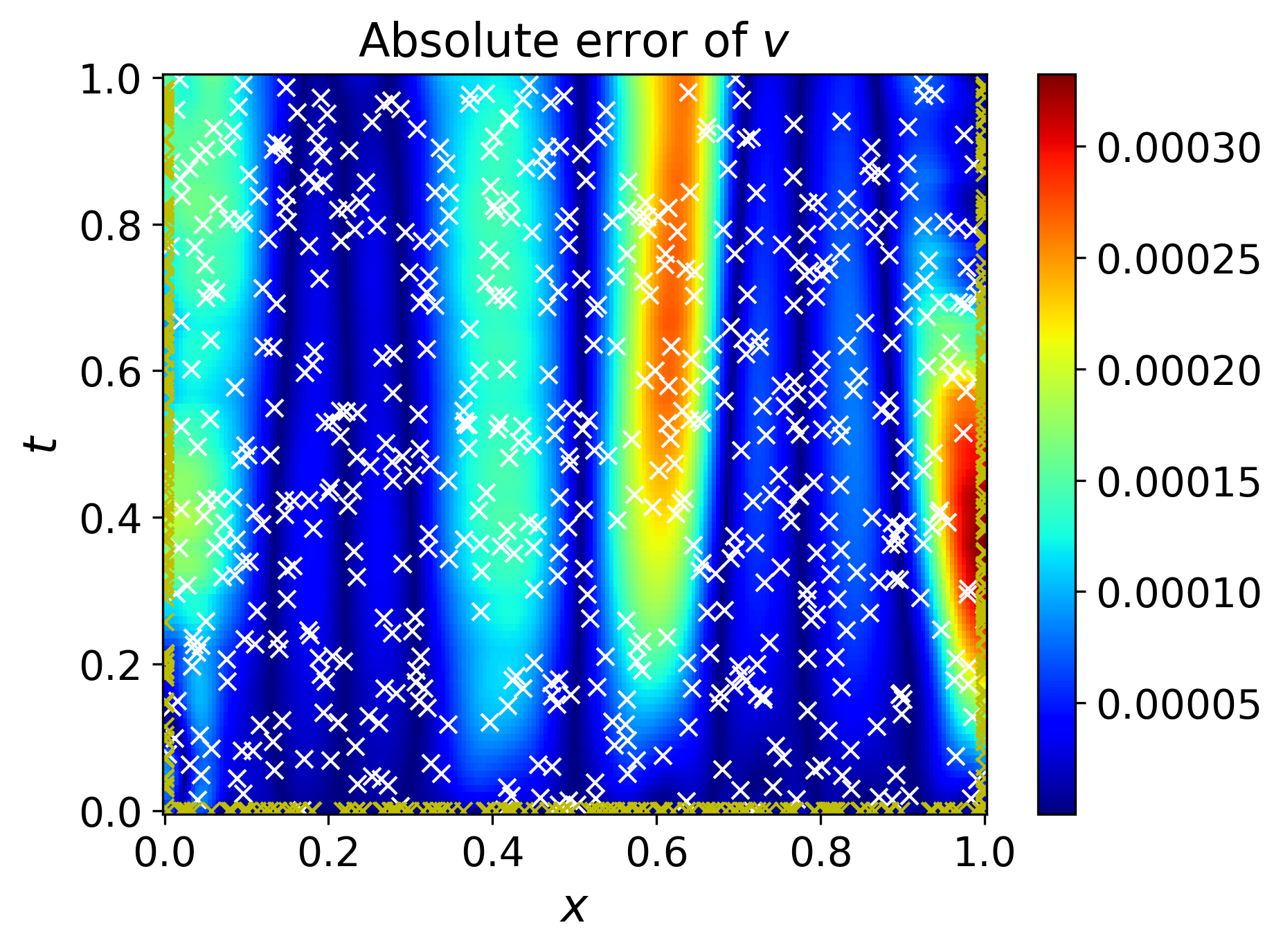}
    \caption{Solving the Gray--Scott equation using the GP-PDE method with learned hyperparameters. The white and yellow ``x'' markers indicate collocation and boundary point locations, respectively.}
    \label{fig:grayscott}
\end{figure}

\begin{table}[tb]
    \footnotesize
    \centering
    \caption{Errors of the solutions obtained using the GP-PDE method with learned versus unlearned hyperparameters for solving the Gray--Scott model \eqref{eq:gs:u}-\eqref{eq:gs:v} with initial conditions \eqref{eq:gs:initial_1}.} 
    \label{tab:grayscott}
    \begin{tabular}{ rcc }
    \toprule
    & Learned hyperparameters & $l^u_t = l^u_x =l^v_t = l^v_x = 1.0$ \\ 
    \midrule
    $L^2$ error of $u, v$ & $1.3123\times10^{-4}, 1.0177\times 10^{-4}$ & $5.2802\times 10^{-2}, 9.8506\times 10^{-2}$  \\ 
    $L^\infty$ error of $u, v$ & $4.0991\times10^{-4}, 3.3251\times 10^{-4}$ & $1.8409\times 10^{-1}, 2.9585\times 10^{-1}$ \\ 
    \bottomrule
    \end{tabular}
\end{table}

The resulting learned hyperparameters are
\begin{align*}
    l^u_t = 1.0542, \ l^u_x = 0.1173, \  l^v_t = 1.5027, \ \text{and} \ l^v_x = 0.1123.
\end{align*}
To evaluate their effectiveness, we solve \eqref{eq:gs:u}-\eqref{eq:gs:v} from scratch using the GP-PDE method \cite{chen2021solving} with the learned hyperparameters and a new set of $600$ collocation and $400$ boundary points. As shown in Figure \ref{fig:grayscott} and Table \ref{tab:grayscott}, the solutions obtained using the learned hyperparameters are significantly more accurate than those computed with the initial values. This confirms the importance of hyperparameter learning in solving PDE systems with multiple GPs.

As a preliminary assessment of the utility of the learned hyperparameters, we test their generalizability and reusability on the same PDE system but with different initial conditions. Specifically, we reuse the hyperparameters learned from initial conditions \eqref{eq:gs:initial_1} to solve \eqref{eq:gs:u}-\eqref{eq:gs:v} under the following new initial conditions:
\begin{enumerate}
    \item Case (A): $u(x, 0) = \sin(7\pi x+\frac{\pi}{2})\qa v(x, 0) = -\cos(2\pi x),$
    \item Case (B): $u(x, 0) = -\cos(4\pi x)\qa v(x, 0) = \sin(5\pi x+\frac{\pi}{2}),$
    \item Case (C): $u(x, 0) = \cos(8\pi x)\qa v(x, 0) = \cos(5\pi x),$
\end{enumerate}
for $x \in (0, 1)$. The corresponding results are reported in Table \ref{tab:grayscott2}. As shown, the GP-PDE method achieves consistently high accuracy when reusing the learned hyperparameters---even under different initial conditions.  This generalizability further enhances the practical utility of the proposed learning strategy in real-world applications where multiple simulations of the same PDE system are required.

\begin{table}[tb]
    \footnotesize
    \centering
    \caption{$L^2$ errors of the solutions ($u$, $v$) obtained using the learned hyperparameters from \eqref{eq:gs:initial_1} to solve the Gray--Scott model \eqref{eq:gs:u}-\eqref{eq:gs:v} with different initial conditions.} 
    \label{tab:grayscott2}
    \begin{tabular}{ ccc }
    \toprule
    & Learned hyperparameters & $l^u_t = l^u_x =l^v_t = l^v_x = 1.0$ \\ 
    \midrule
    Case (A) & $5.0970\times10^{-4}, 3.1712\times10^{-4}$ & $4.6469\times 10^{-1}, 9.8150\times10^{-2}$  \\ 
    Case (B) & $4.4294\times10^{-4}, 3.4435\times10^{-4}$ & $5.3386\times 10^{-2}, 1.8805\times 10^{-1}$ \\ 
    Case (C) & $7.8320\times 10^{-4}, 7.9105\times 10^{-4}$ & $4.6179\times 10^{-1}, 1.7078\times 10^{-1}$  \\ 
    \bottomrule
    \end{tabular}
\end{table}

\subsection{Eikonal Equation with Non-Stationary Kernel Parameterized by Neural Networks}\label{sec:num_eik}
In this example, we solve the following regularized Eikonal equation
\begin{align}
\label{eq:eikonal}
    \begin{cases}
        \begin{alignedat}{2}
            |\nabla u(x, y)|^2&=f(x, y) + \epsilon \Delta u(x, y)  && \quad  \forall (x,y) \in \Omega,\\
            u(x,y)&=0 && \quad \forall (x,y) \in \partial\Omega,
        \end{alignedat}
    \end{cases}
\end{align}
where $\Omega \coloneqq  (0, 1)^2$, $f(x, y) \equiv 1$, and $\epsilon = 0.01$. The reference solution is computed using the transformation $u = -\epsilon \log(v)$, which yields the linear PDE $-\epsilon^2 \Delta v + f v = 0$. This transformed equation is solved using a second-order finite difference scheme on a uniform mesh with grid size $1/1000$, following the approach of \cite{chen2021solving}, to provide an exact solution.

To test the applicability and scalability of our method in the presence of more complex hyperparameter structures, we consider a non-stationary kernel: the Gibbs kernel \cite{gibbs1998bayesian, rasmussen2006gaussian}, which allows spatially varying lengthscales. Specifically, we model the solution using a GP equipped with a Gibbs kernel
\begin{equation}
    \kappa(\mathbf{x}, \mathbf{x}^\prime) = \left(\prod_{i=1}^d\sqrt{\frac{2l_i(\mathbf{x})l_i(\mathbf{x}^\prime)}{l^2_i(\mathbf{x})+l^2_i(\mathbf{x}^\prime)}}\right)\exp\left(-\sum_{i=1}^d\frac{(x_i - x^\prime_i)^2}{l^2_i(\mathbf{x})+l^2_i(\mathbf{x}^\prime)}\right), 
\end{equation}
where $\mathbf{l}(\mathbf{x}) = [l_1(\mathbf{x}), \ldots, l_d(\mathbf{x})]$ specifies a separate length scale for each dimension, $\mathbf{x} = [x_1, x_2, \ldots, x_d]$, and $d$ denotes the dimension. Due to the symmetry of the Eikonal equation, we use an isotropic form: $l_1(\mathbf{x}) = l_2(\mathbf{x}) = l(\mathbf{x})$. The spatially varying lengthscale function $l(\mathbf{x})$ is parameterized by a neural network with two hidden layers, each containing $50$ neurons and using the hyperbolic tangent activation function. This yields a total of $2,751$ trainable hyperparameters. The network takes $(x, y)$ as input and outputs a single scalar lengthscale shared across both spatial dimensions. 

We adopt the DTO scheme in Subsection~\ref{subsec:DTO:general}. At initialization, we sample and fix a collocation set consisting of $900$ interior points in the domain $\Omega$ and $300$ boundary points. These are used for solving the PDE at every GN step. We also sample, once and for all, an additional validation set of $900$ interior points for the outer optimization. In each GN iteration, we compute the hyperparameter update by uniformly subsampling a mini-batch of $200$ points from this fixed validation set.  We employ the Adam optimizer with a learning rate of $1\times10^{-3}$ to train the neural network and a nugget term of $1\times10^{-8}$ is used. We perform $30$ GN iterations, each followed by $50$ learning steps for hyperparameter optimization.

\begin{figure}[tb]
    \centering
    \includegraphics[width=0.24\linewidth]{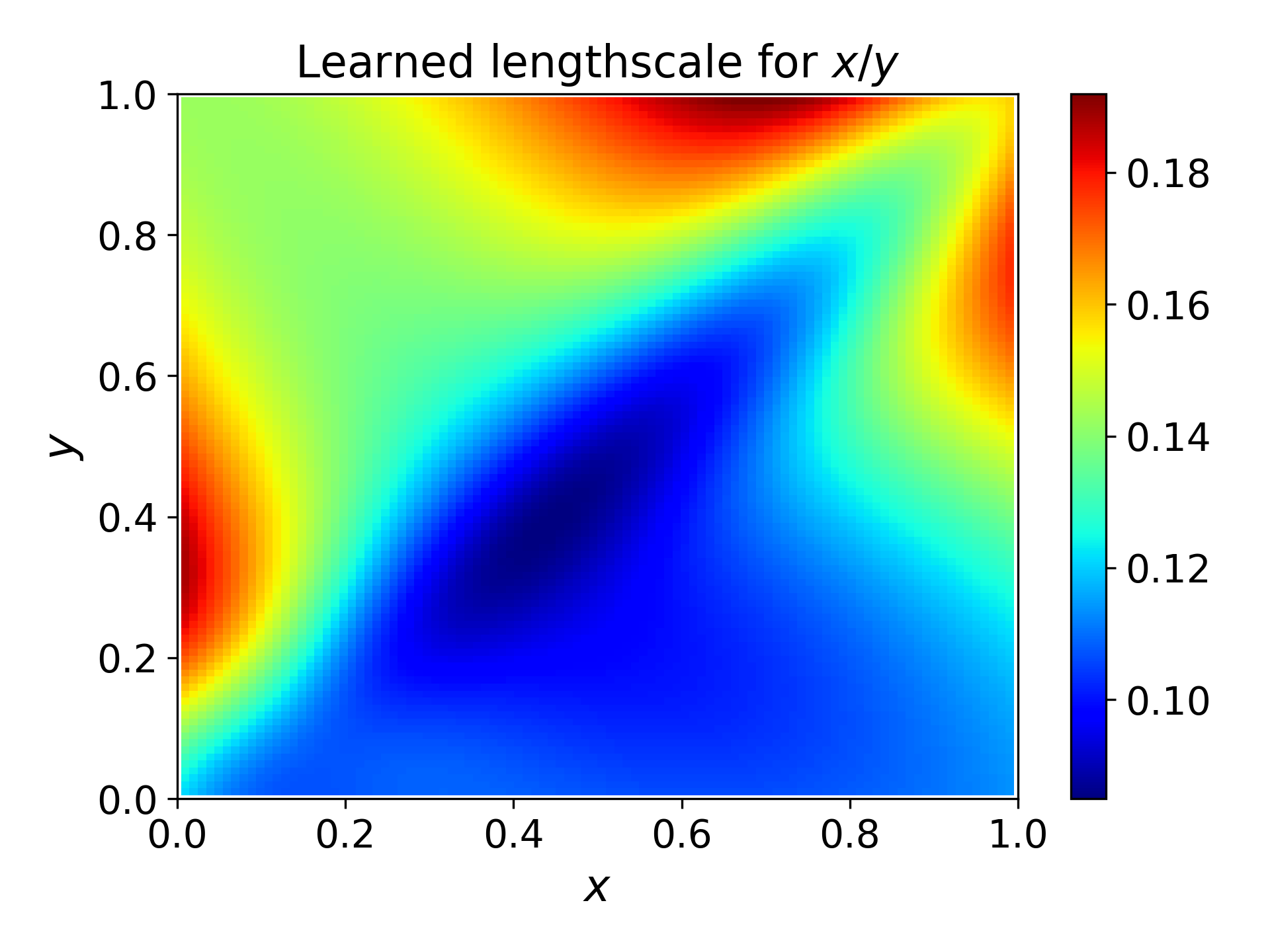}
    \includegraphics[width=0.24\linewidth]{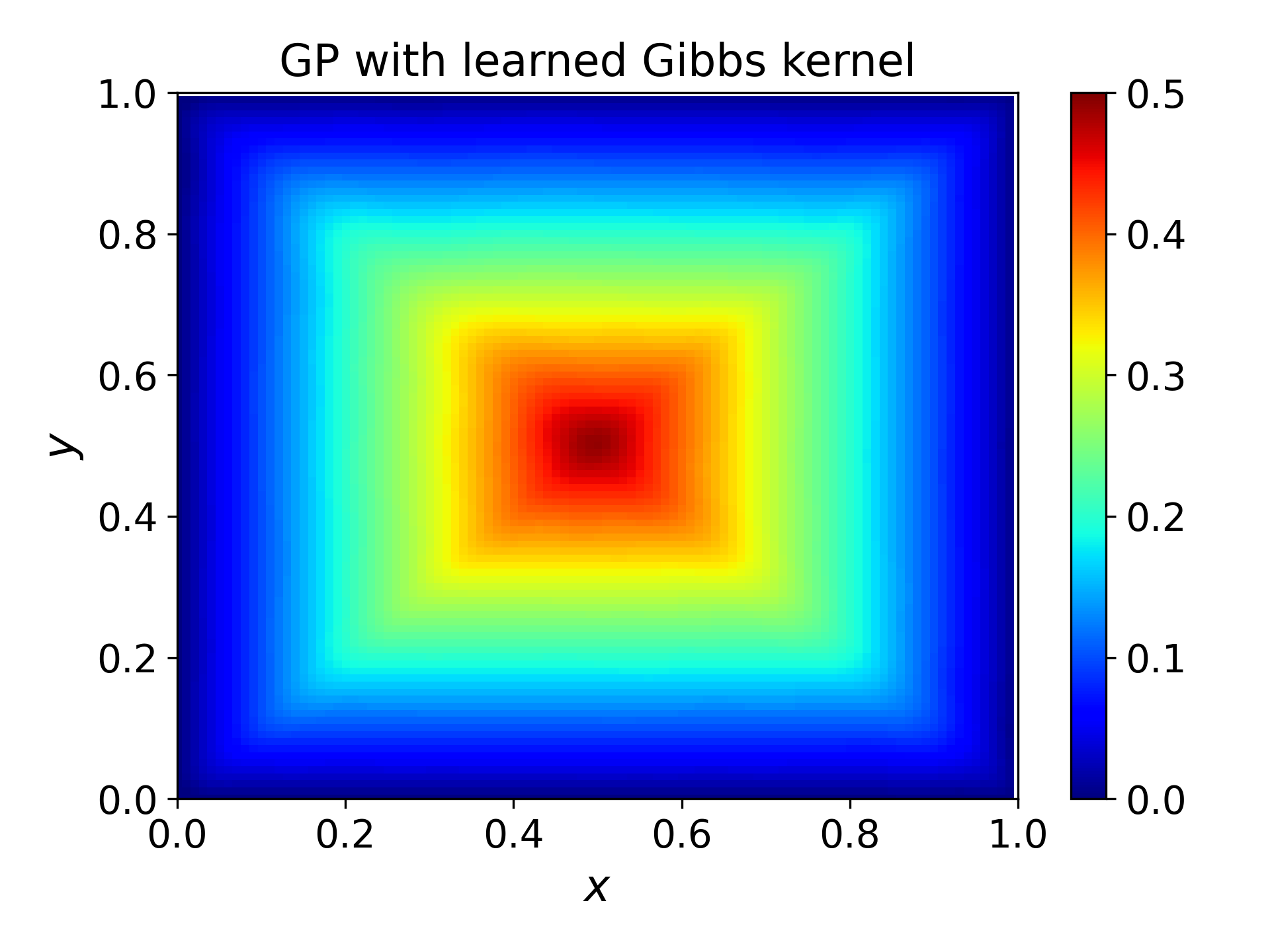}
    \includegraphics[width=0.24\linewidth]{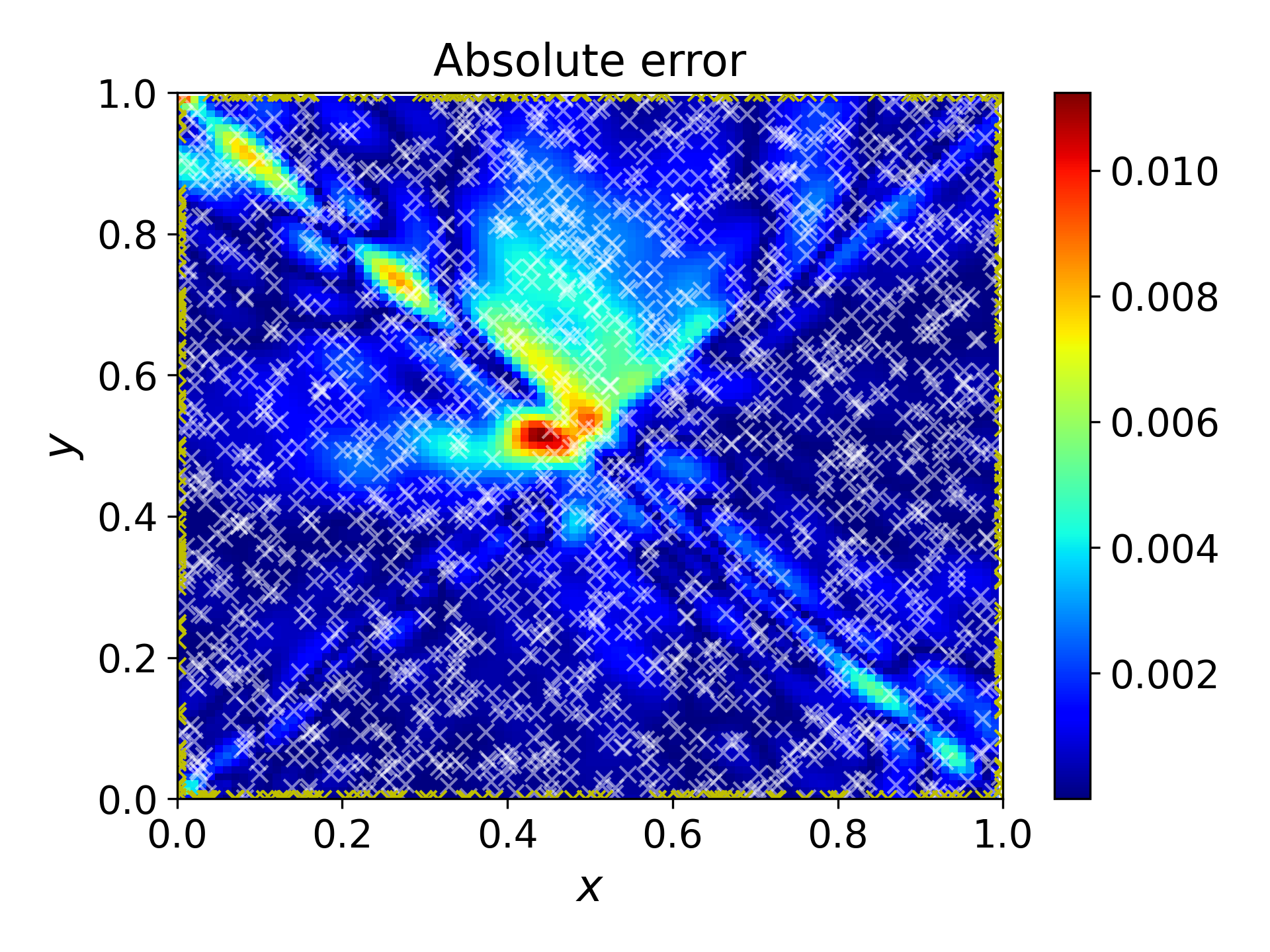}
    \includegraphics[width=0.24\linewidth]{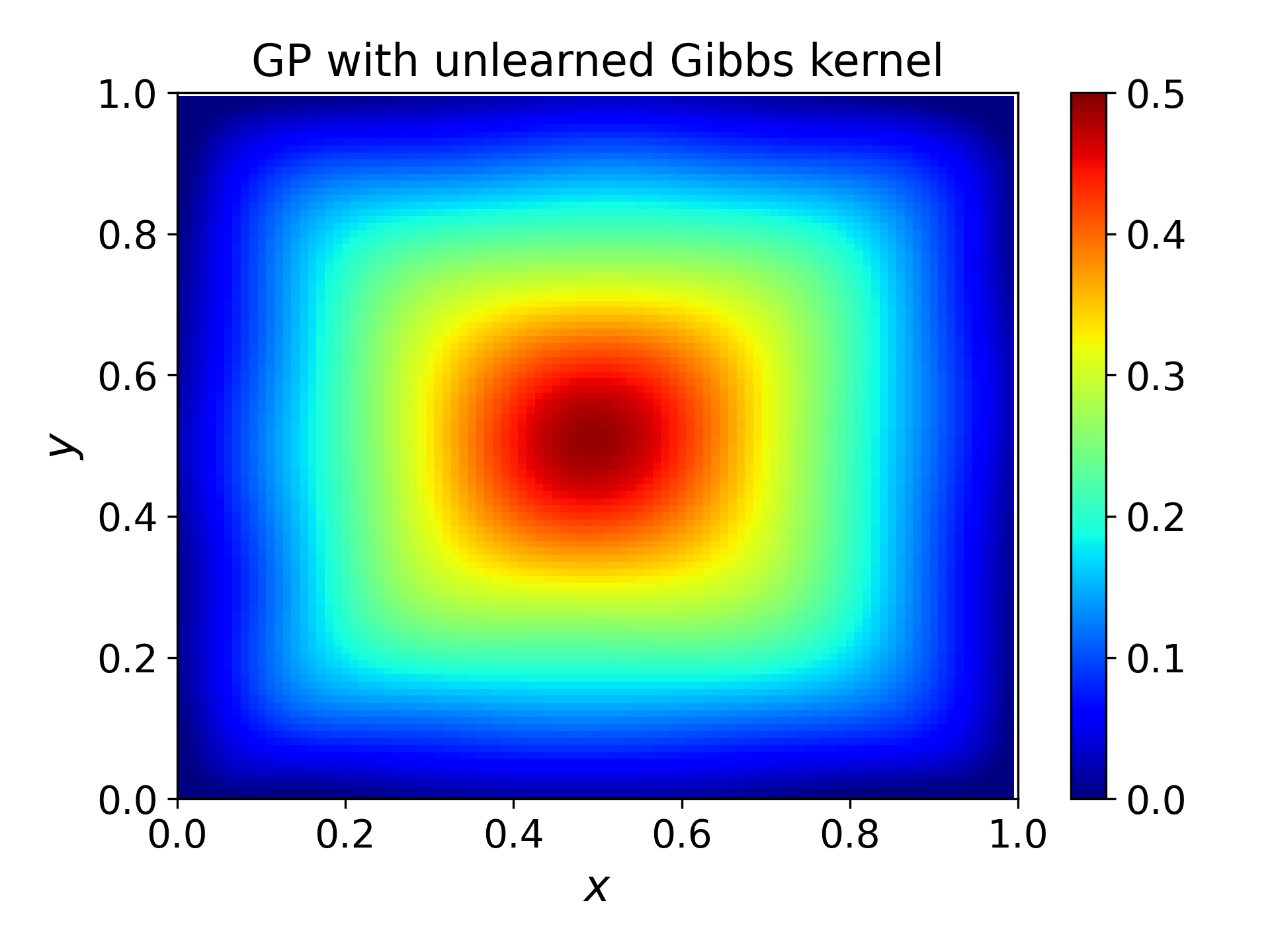}
    \caption{Solving the Eikonal equation using the GP-PDE method with an isotropic Gibbs kernel. Left to right: learned spatially-varying lengthscale, solution obtained with learned kernel, absolute error, and solution obtained with unlearned (initialized) kernel.}
    \label{fig:eikonal}
\end{figure}

The learned lengthscale function is visualized in the left panel of Figure \ref{fig:eikonal}. Notably, the lengthscale is significantly smaller (less than $0.10$) near the domain center $(x, y) = (0.5, 0.5)$, reflecting the reduced smoothness of the solution in this region-consistent with the behavior of the Eikonal equation. This indicates that the learned lengthscale is not only effective but also interpretable in terms of local solution regularity.

We further evaluate the effectiveness of the learned lengthscale by solving \eqref{eq:eikonal} with the GP-PDE method from scratch using the Gibbs kernel and the learned lengthscale and all $1,800$ collocation points and $300$ boundary points (which are used in the learning stage). For comparison, we also solve the problem using the untrained kernel, where the neural network retains its initial weights. We perform $30$ GN iterations. Figure \ref{fig:eikonal} and Table \ref{tab:eikonal} present the corresponding solutions, absolute errors, and quantitative error metrics. The results demonstrate that the learned kernel yields significantly improved accuracy in both $L^2$ and $L^\infty$ norms, highlighting the method's effectiveness for high-dimensional hyperparameter spaces and non-stationary kernel structures.

\begin{table}[tb]
    \footnotesize
    \centering
    \caption{Errors of the solutions obtained using the GP-PDE method with learned versus unlearned isotropic Gibbs kernels for solving the Eikonal equation.} 
    \label{tab:eikonal}
        \begin{tabular}{ rcc }
    \toprule
      & Learned Gibbs kernel & Unlearned Gibbs kernel \\ 
     \midrule
     $L^2$ error& $1.8295\times10^{-3}$ & $1.5798\times 10^{-2}$   \\ 
     $L^\infty$ error & $1.1240\times10^{-2}$ & $4.7330\times 10^{-2}$ \\ 
     \bottomrule
    \end{tabular}
\end{table}

This example demonstrates the scalability of the proposed method to settings involving non-stationary kernels and high-dimensional hyperparameter spaces, such as those induced by neural network parameterizations. Moreover, it shows that the learned lengthscales reflect meaningful spatial variation in the solution's regularity, further validating the interpretability and utility of the approach.

\subsection{Burgers' Equation}\label{sec:num_burg}
We solve the following Burgers' equation:
\begin{align}
\label{eq:burgers}
    \begin{cases}
        \begin{alignedat}{2}
            \frac{\partial u}{\partial t} + u\frac{\partial u}{\partial x} - \nu \frac{\partial^2 u}{\partial x^2}&=0  && \quad  \forall (x, t) \in (-1, 1)\times(0, 1],\\[2mm]
            u(x,0)&=-\sin(\pi x) && \quad \forall x \in (-1, 1),\\[2mm]
            u(-1, t) &= u(1, t) = 0 && \quad \forall t \in (0, 1], 
        \end{alignedat}
    \end{cases}
\end{align}
with $\nu=0.02$. The solution develops a steep gradient (shock-like structure) over time, making this an ideal testbed for modeling non-uniform smoothness via non-stationary kernels. 

In this setting, we employ an anisotropic Gibbs kernel, where the lengthscales are spatially dependent and learned through a neural network. Specifically, the input to the kernel is $\mathbf{x} = (t, x)$, and the neural network outputs two separate lengthscales: $l_t(t, x)$ for time and $l_x(t, x)$ for space. The network architecture comprises two hidden layers with $50$ neurons each and hyperbolic tangent activation, resulting in $2,802$ trainable parameters.

\begin{figure}[tb]
    \centering
    \includegraphics[width=0.24\linewidth]{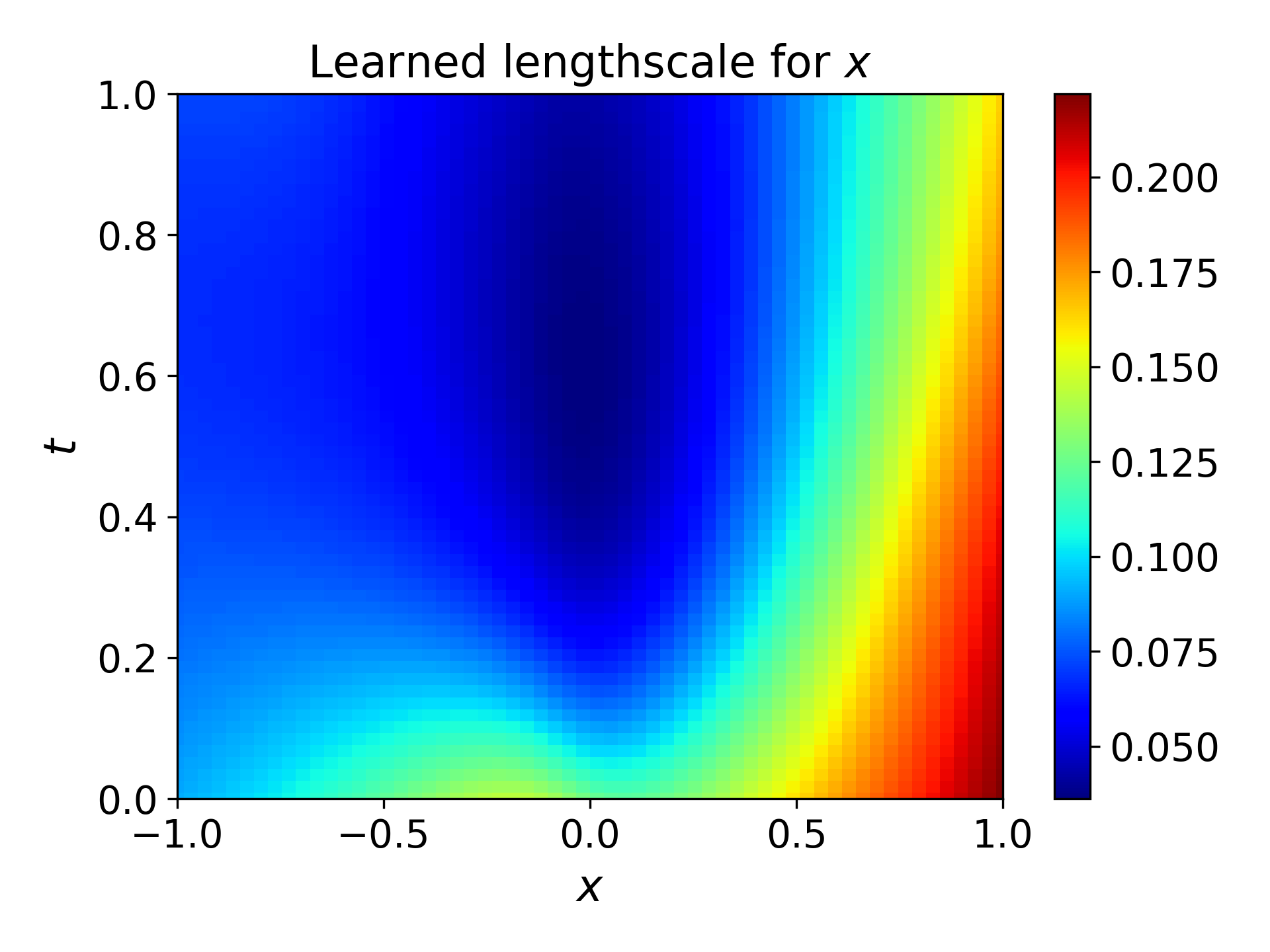}
    \includegraphics[width=0.24\linewidth]{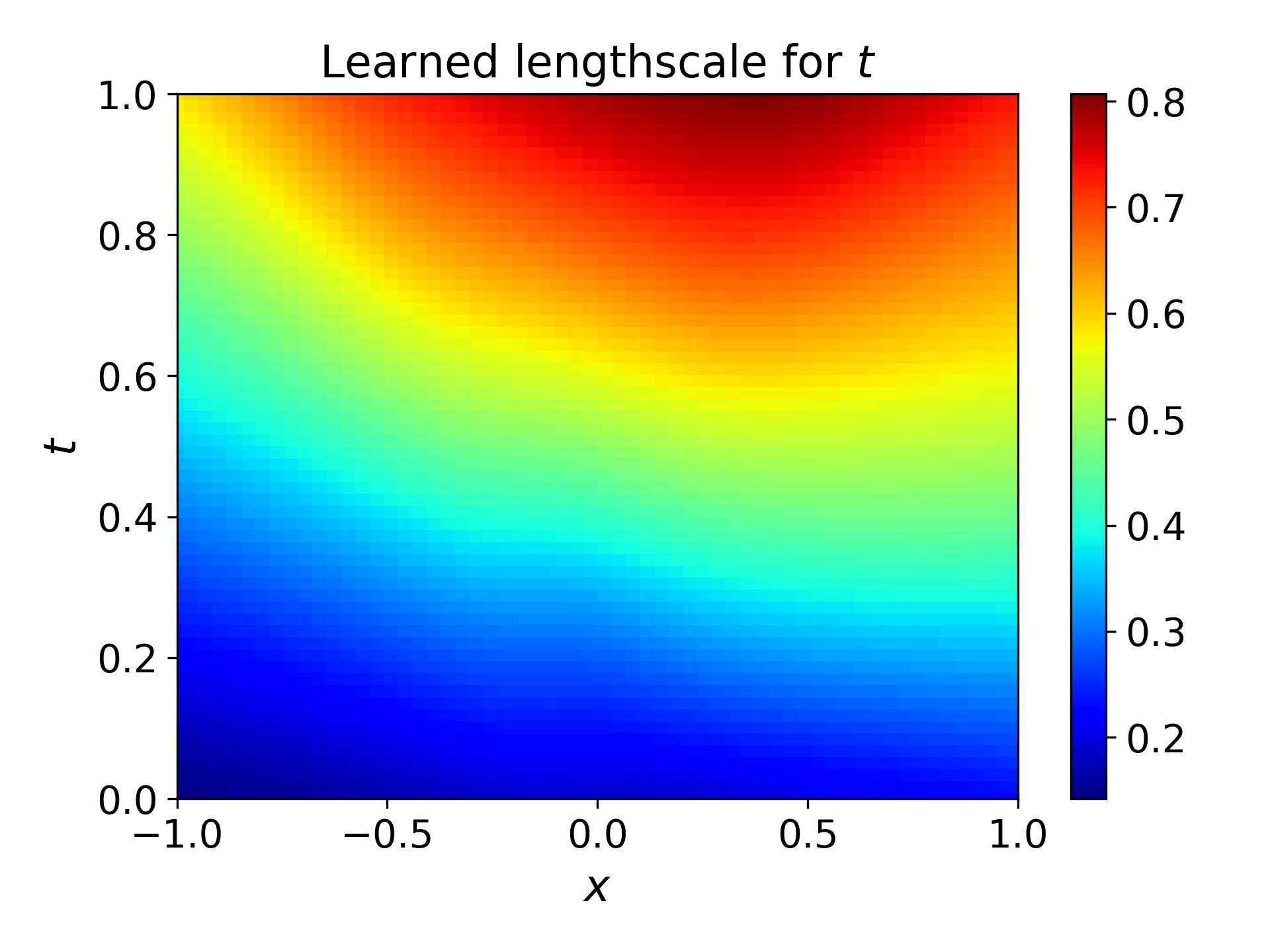}
    \includegraphics[width=0.24\linewidth]{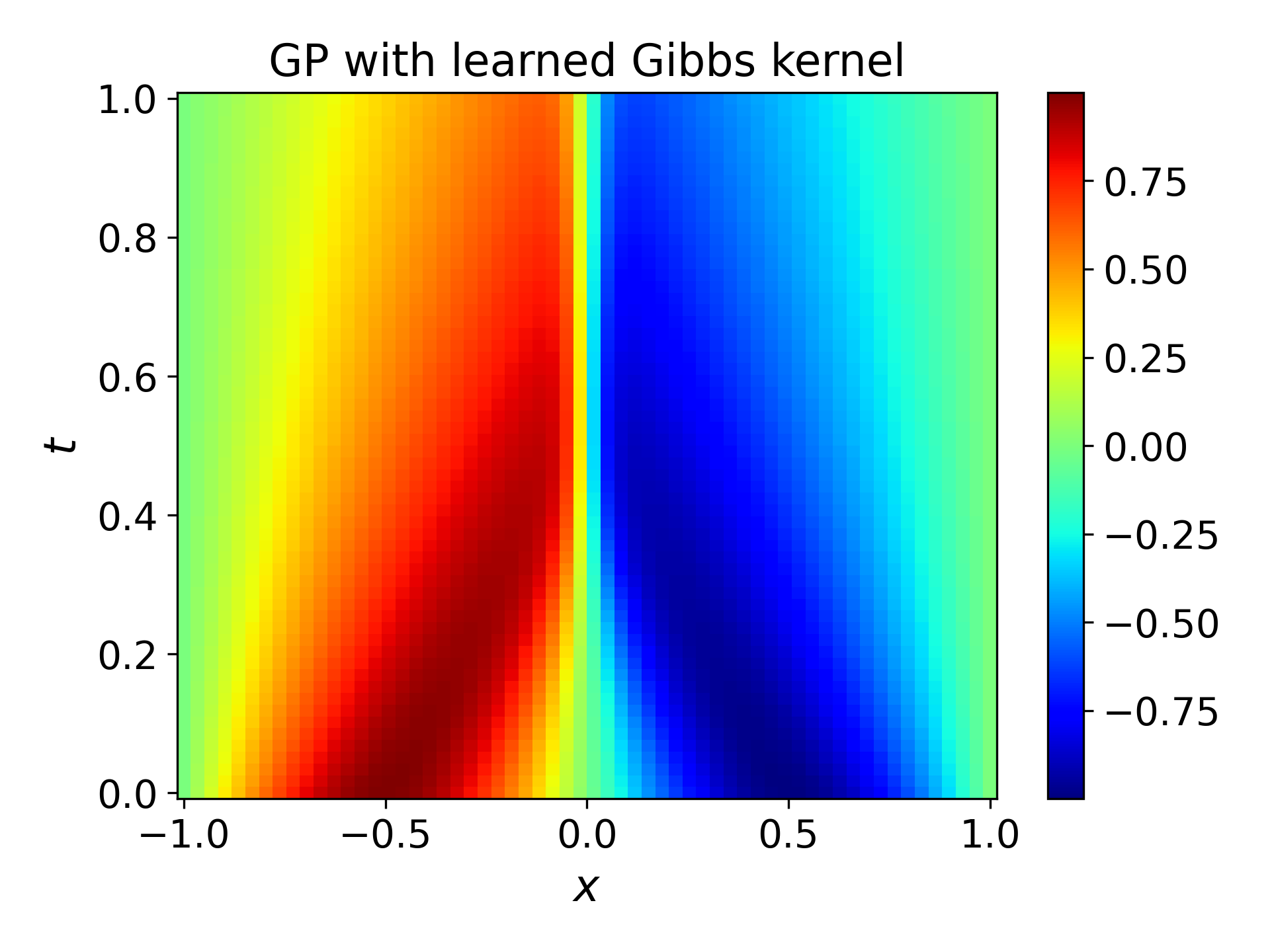}
    \includegraphics[width=0.24\linewidth]{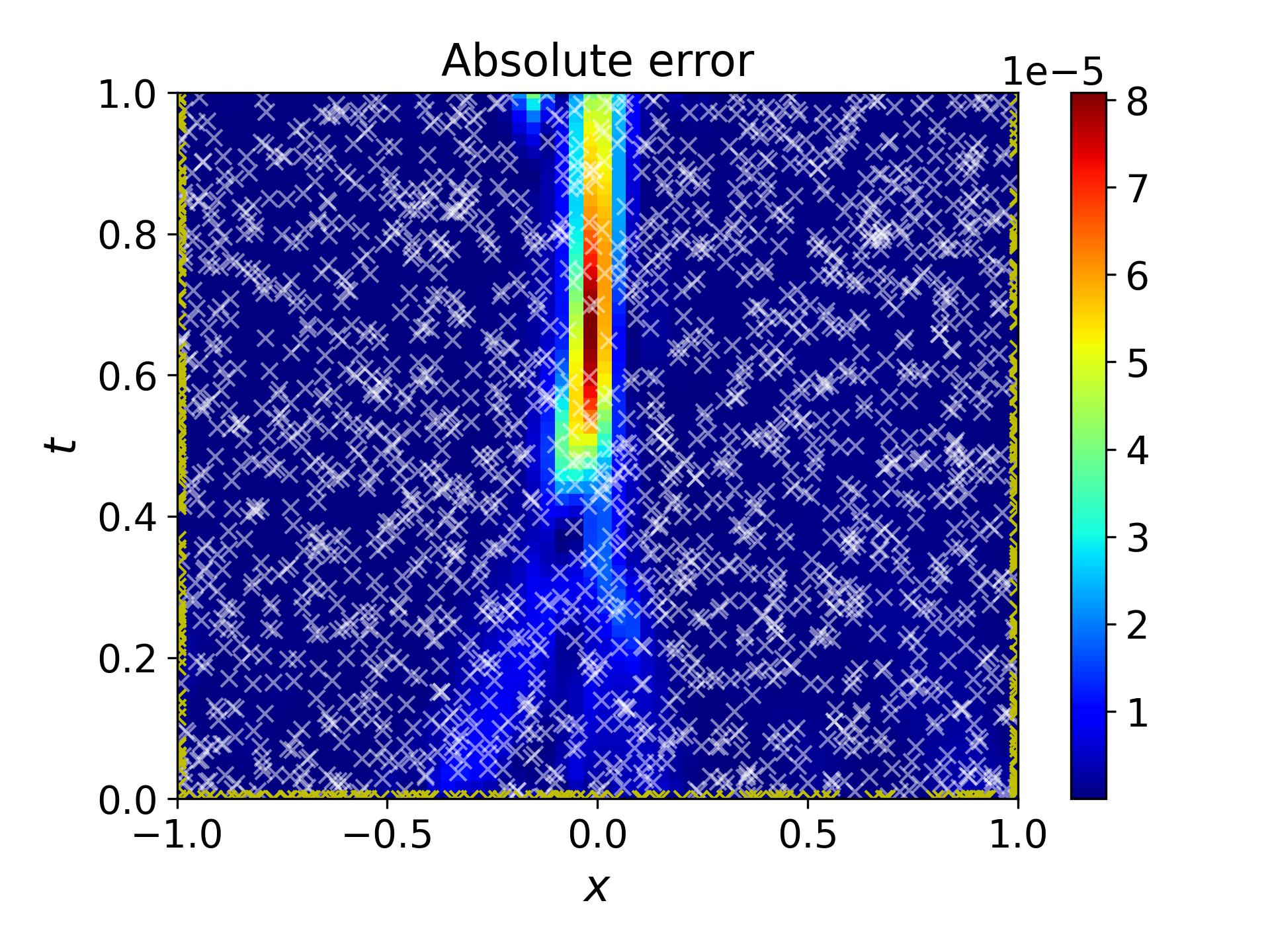}
    \caption{Solving Burgers' equation using the GP-PDE method with an anisotropic Gibbs kernel. Left to right: spatial lengthscale $l_x(t,x)$, temporal lengthscale $l_t(t,x)$, predicted solution with learned kernel, and absolute error.}
    \label{fig:burgers}
\end{figure}

To train the neural network, we first draw (and then keep fixed for all iterations) a collocation set of $900$ interior points and $300$ boundary points. For the hyperparameter optimization, we draw a separate validation set of $900$ interior points. At each GN step, the PDE is solved on the fixed collocation grid, and the hyperparameters are updated using a mini-batch of $200$ points uniformly subsampled from the fixed validation set. We employ the Adam optimizer with a learning rate of $1\times10^{-3}$ to train the neural network and a nugget term of $1\times10^{-10}$ is used. We perform $20$ GN iterations, each followed by $50$ learning steps. The learned lengthscale fields are shown in the first two panels of Figure \ref{fig:burgers}. The spatial lengthscale $l_x(t, x)$ is notably reduced near $x = 0$ as $t$ approaches $1.0$, indicating sharper features in that region -- a trend consistent with the expected shock formation in the Burgers' solution (see third panel of Figure~\ref{fig:burgers} for an approximated solution to \eqref{eq:burgers}). In contrast, the temporal lengthscale $l_t(t, x)$ exhibits less pronounced variation, suggesting relatively uniform smoothness in time. These patterns reflect the ability of the learned kernel to adaptively capture local variations in solution regularity.

To assess the impact of the learned lengthscales, we solve \eqref{eq:burgers} from scratch using the GP-PDE method with the trained Gibbs kernel and the full set of $1,800$ collocation and $300$ boundary points used during training. For comparison, we also evaluate the GP-PDE method using an untrained kernel in which the neural network remains at its initial state. We perform $10$ GN iterations in case each case. Figure~\ref{fig:burgers} and Table~\ref{tab:burgers} show the predicted solutions, error distributions, and error norms. The results clearly demonstrate a substantial improvement in accuracy when the learned kernel is used in both $L^2$ and $L^\infty$ errors.

\begin{table}[tb]
    \footnotesize
    \centering
    \caption{Errors in solving Burgers' equation \eqref{eq:burgers} using the GP-PDE method with learned and unlearned anisotropic Gibbs kernels.} 
    \label{tab:burgers}
    \begin{tabular}{ rcc }
    \toprule
    & Learned Gibbs kernel & Unlearned Gibbs kernel \\ 
    \midrule
    $L^2$ error& $9.5012\times10^{-6}$ & $1.5085\times 10^{-1}$   \\ 
    $L^\infty$ error & $8.0836\times10^{-5}$ & $9.5425\times 10^{-1}$ \\ 
    \bottomrule
    \end{tabular}
\end{table}

\subsection{Darcy Flow Inverse Problem}\label{sec:num_darcy}
We consider the two-dimensional Darcy flow problem described by
\begin{align}
\label{eq:darcy_flow_eq}
    \begin{cases}
        \begin{alignedat}{2}
            -\nabla\cdot(\exp(a)\nabla u)(\mathbf{x})&=f(\mathbf{x})  && \quad  \forall \mathbf{x}\in\Omega,\\[2mm]
            u(\mathbf{x})&=0 && \quad \forall \mathbf{x}\in\partial\Omega,
        \end{alignedat}
    \end{cases}
\end{align}
where $\Omega=(0, 1)^2$. 
We consider the inverse problem with the true coefficient
\(a^{\star}(\mathbf{x})\) satisfying
\begin{equation}
  \exp\bigl(a^{\star}(\mathbf{x})\bigr)
  =
  \exp\bigl(\sin(2\pi x_{1}) + \sin(2\pi x_{2})\bigr)
  +
  \exp\bigl(-\sin(2\pi x_{1}) - \sin(2\pi x_{2})\bigr).
  \label{eq:true-coeff}
\end{equation}
The right-hand-side source term is \(f \equiv 1\).
For the inverse problem, we randomly select \(L = 60\) locations
\(\{\mathbf{x}_\ell\}_{\ell=1}^{L}\subset\Omega\) and observe the corresponding
values of the state \(u(\mathbf{x}_\ell)\).
Reference values \(u^{\star}(\mathbf{x}_\ell)\) are generated by first solving
\eqref{eq:darcy_flow_eq} with the true coefficient \(a^{\star}\) on a uniform
grid using a finite-difference scheme and then interpolating the resulting
solution to the observation points.
Independent Gaussian noise
\(\mathcal{N}\!\bigl(0,\gamma^{2}I\bigr)\) with
standard deviation \(\gamma = 10^{-3}\) is added to these observations. Both the coefficient \(a\) and the state \(u\) are modeled as zero-mean GPs with RBF kernels
\eqref{eq:rbf:kernel}; their lengthscales \(l_a\) and \(l_u\) are learned. A nugget of \(10^{-5}\) is appended to the diagonal of each Gram matrix. 

We follow the DTO scheme in
Subsection~\ref{subsec:bilevel:inverse}.
At initialization we draw, and then keep fixed, a collocation set of
\(400\) interior points and \(100\) boundary points for solving the PDE at
every GN step. A separate validation set of \(400\) interior points  is likewise fixed for the outer objective.
At each GN iteration, we approximate the hypergradient using a mini-batch consisting of \(60\) uniformly subsampled validation points together with the \(60\) observation points. In \eqref{eq:inv:dto:loss}, we set  $N_s=400$, $\eta_1=0$, and $\eta_2 = \tfrac{1}{\gamma^2N_s}$. 
 The Adam optimizer with learning rate \(10^{-3}\) updates the lengthscales,
starting from the common initial value \(l_a=l_u=1\).
We perform \(30\) GN steps, each followed by \(100\) hyperparameter updates.

\begin{figure}[tb]
      \centering
      \begin{subfigure}[b]{0.24\linewidth}
        \includegraphics[width=\linewidth]{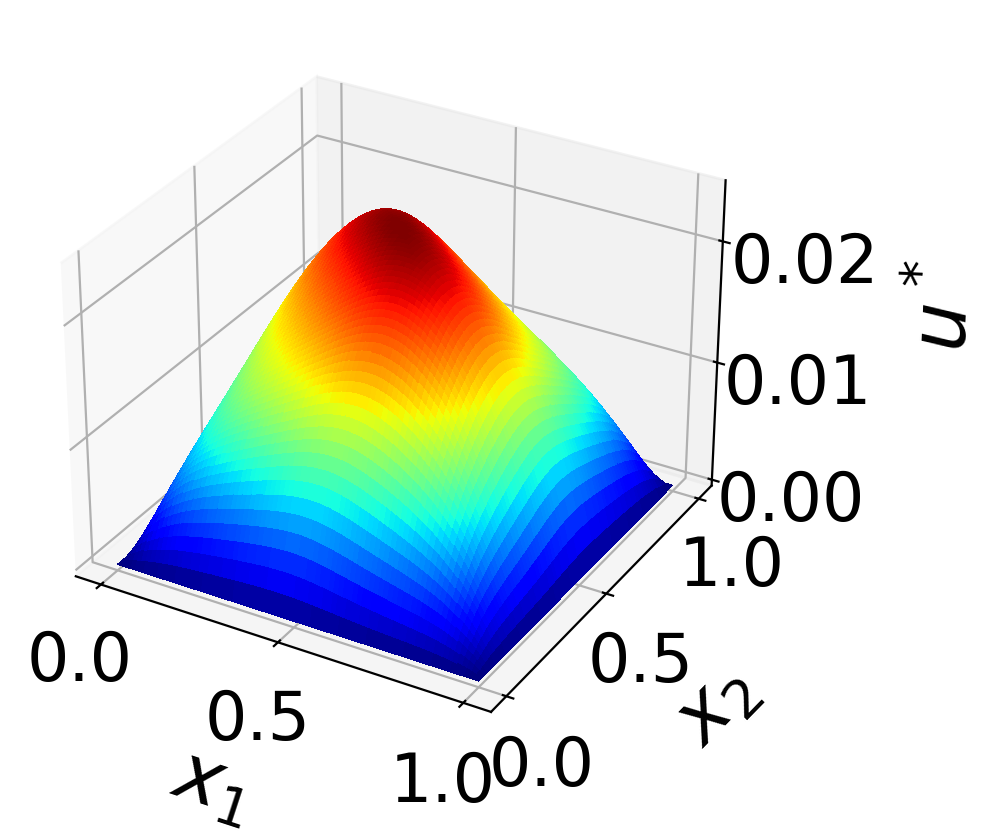}
        \caption{Reference state $u^{\star}$}
      \end{subfigure}
      \begin{subfigure}[b]{0.24\linewidth}
        \includegraphics[width=\linewidth]{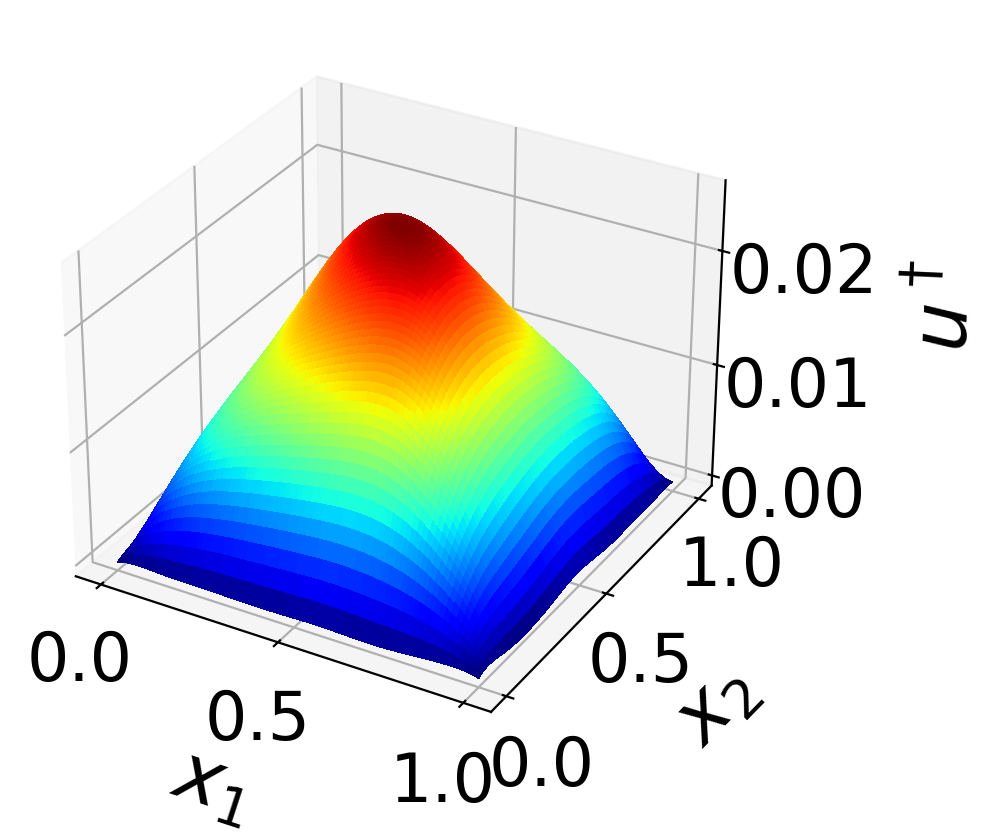}
        \caption{GP estimate $u^{\dagger}$}
      \end{subfigure}
      \begin{subfigure}[b]{0.24\linewidth}
        \includegraphics[width=\linewidth]{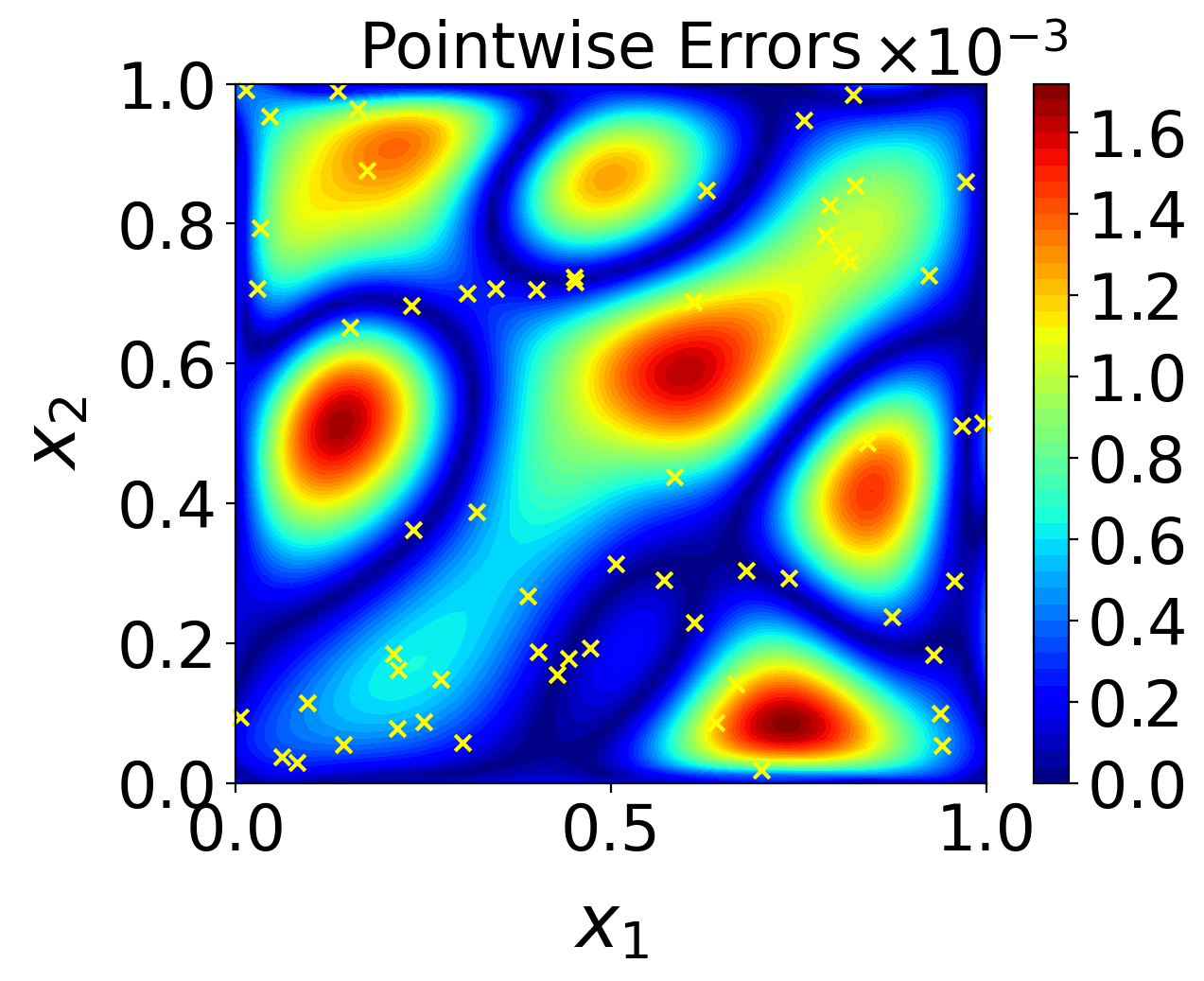}
        \caption{Pointwise error $|u^{\dagger}\!-\!u^{\star}|$}
      \end{subfigure}
    
      \vspace{4mm}
    
      \begin{subfigure}[b]{0.24\linewidth}
        \includegraphics[width=\linewidth]{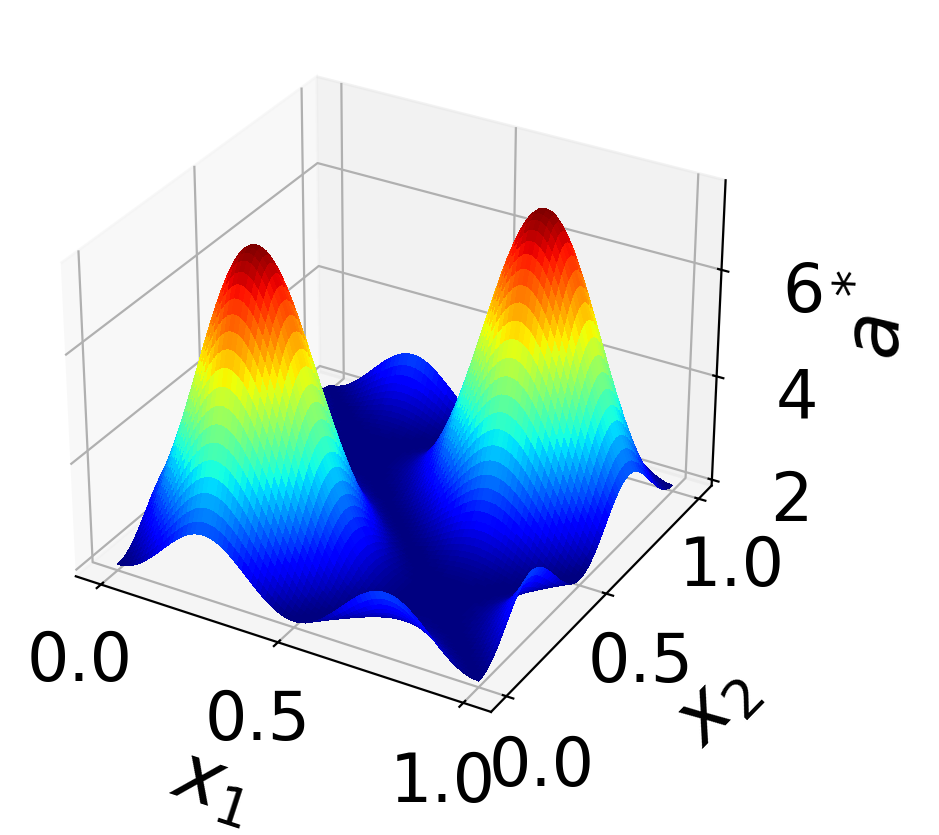}
        \caption{True coefficient $a^{\star}$}
      \end{subfigure}
      \begin{subfigure}[b]{0.24\linewidth}
        \includegraphics[width=\linewidth]{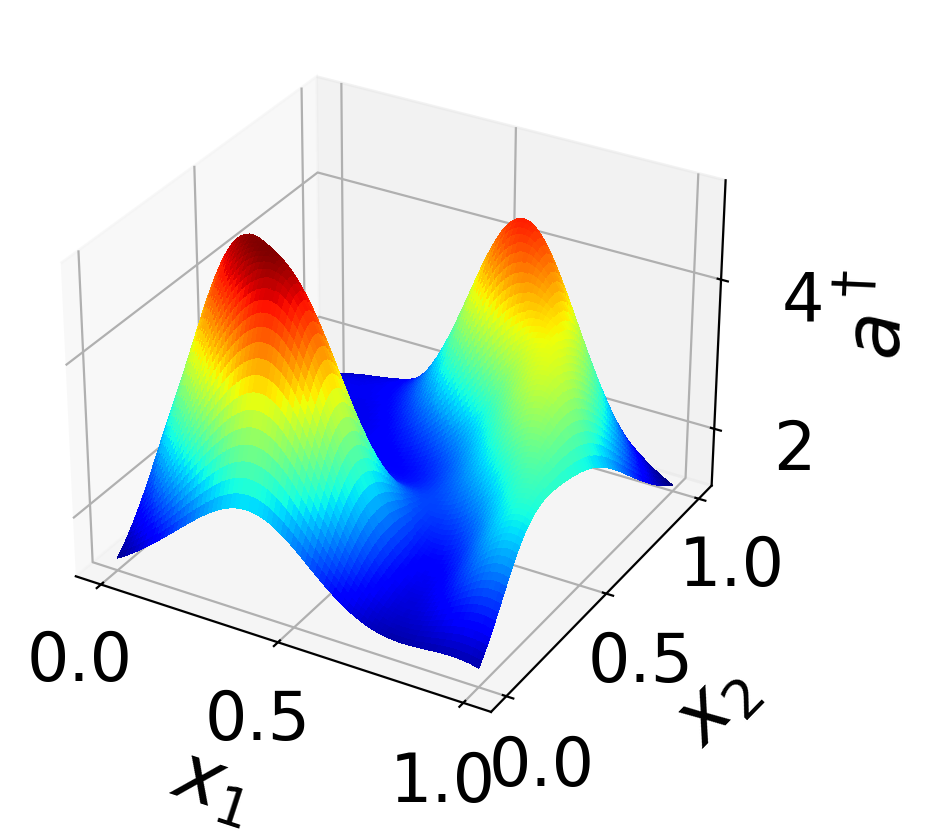}
        \caption{Recovered $a^{\dagger}$}
      \end{subfigure}
    
    \caption{
    Solution of the inverse Darcy flow problem using learned  kernels.
    The recovered coefficient~$a^\dagger$ and forward state~$u^\dagger$ closely match the true fields~$a^\star$ and~$u^\star$.
    Each panel shows the reference fields, GP reconstructions, or the pointwise error.
    }
    
      \label{fig:darcy_flow}
\end{figure}

Figure~\ref{fig:darcy_flow} compares the reference state \(u^{\star}\),
its GP approximation \(u^{\dagger}\), and their pointwise error; the true
coefficient \(a^{\star}\) and its reconstruction \(a^{\dagger}\) are also
shown. The results demonstrate reasonably accurate recovery of both \(a\) and the forward
solution \(u\). In contrast, keeping the initial lengthscales fixed at \(l_a=l_u=1\)
causes the GN iterations to diverge, underscoring the necessity of kernel learning and the effectiveness of the proposed bilevel method.

\section{Conclusion and Future Work}\label{sec:conclusion}
This work proposes a bilevel algorithm for hyperparameter learning and applies it to PDE and inverse problem solvers. The paper casts hyperparameter selection as a bilevel optimization problem, performs a single Gauss--Newton linearization of the inner problem, and exploits the closed-form expression for the linearized state update. As a result, each outer iteration reduces to one linear solve and a hyperparameter optimization, avoiding both full inner convergence and lengthy reverse-mode unrolling. A series of numerical experiments on PDEs and inverse problems show significant gains in accuracy and robustness over random and grid-search initialization.

There are several promising directions for future work. Firstly, it is natural to extend the approach to operator learning and PDE discovery problems \cite{jalalian2025data}. 
Secondly, on the computational side, it is natural to seek ways to accelerate the
computations; the method as described in this paper requires assembling and inverting or factorizing Gram matrices whose dimension equals the number of collocation points, which limits scalability to very large-scale or high-dimensional settings. One may accelerate the linearized inner solve via low-rank and randomized sketching techniques, sparse Cholesky factorizations \cite{chen2025sparse}, Hessian-free iterative methods, or hierarchical kernel approximations, thereby alleviating the Gram-matrix bottleneck. 
A third interesting direction is to extend the proposed framework to train neural networks by integrating kernel-based outer optimization with neural network-based inner representations \cite{cyr2020robust,shao2025solving}. For instance, one could optimize the nonlinear hidden layers of a neural network in the outer loop, while treating the final linear layer as part of the inner optimization governed by the PDE constraints. This hybrid formulation would bridge kernel methods and neural network training, potentially combining the interpretability and structure-awareness of kernel-based models with the expressive power of deep learning, enabling improved accuracy and enhanced insight into learned representations.
Fourthly, on the analysis side, developing convergence guarantees, adaptive trust-region or line-search strategies for the Gauss--Newton step, and rigorous bounds on the linearization error would strengthen the theoretical foundations. One may also explore alternative linearization schemes for the inner subproblem, such as the Levenberg--Marquardt method, which can yield closed-form updates and improved convergence. Finally, integrating the bilevel Gauss--Newton scheme with Bayesian uncertainty quantification or multi-fidelity models offers a path toward scalable and uncertainty-aware hyperparameter learning.

\appendix

\section*{Acknowledgments}
All authors acknowledge support from the Air Force Office of Scientific Research through the MURI award FA9550-20-1-0358 (Machine Learning and Physics-Based Modeling and Simulation).
HO and AMS are grateful for support through their respective
Department of Defense Vannevar Bush Faculty Fellowships.
NHN is partially supported by a Klarman Fellowship through Cornell University's College of Arts \& Sciences, the U.S. National Science Foundation Graduate Research Fellowship Program under award DGE-1745301, the Amazon/Caltech AI4Science Fellowship, and the Department of Defense Vannevar Bush Faculty Fellowship held by AMS.
HO, XY and ZZ  acknowledge support from the Air Force Office of Scientific Research under MURI award number
FOA-AFRL-AFOSR-2023-0004 (Mathematics of Digital Twins), the Department of Energy under award number DE-SC0023163 (SEA-CROGS: Scalable, Efficient, and Accelerated Causal Reasoning Operators, Graphs and Spikes for Earth and Embedded Systems),  the National Science Foundation under award number 2425909 (Discovering the Law of Stress Transfer and Earthquake Dynamics in a Fault Network using a Computational Graph Discovery Approach) and the VBFF under ONR-N000142512035.

\bibliographystyle{siam}
\bibliography{sample}

\begin{thebibliography}{10}

\bibitem{arridge2019solving}
{\sc S.~Arridge, P.~Maass, O.~{\"O}ktem, and C.-B. Sch{\"o}nlieb}, {\em Solving inverse problems using data-driven models}, Acta Numerica, 28 (2019), pp.~1--174.

\bibitem{bachoc2013cross}
{\sc F.~Bachoc}, {\em Cross validation and maximum likelihood estimations of hyper-parameters of {G}aussian processes with model misspecification}, Computational Statistics \& Data Analysis, 66 (2013), pp.~55--69.

\bibitem{bachoc2018asymptotic}
\leavevmode\vrule height 2pt depth -1.6pt width 23pt, {\em Asymptotic analysis of covariance parameter estimation for {G}aussian processes in the misspecified case}, Bernoulli, 24 (2018), pp.~1531--1575.

\bibitem{baptista2025solving}
{\sc R.~Baptista, E.~Calvello, M.~Darcy, H.~Owhadi, A.~M. Stuart, and X.~Yang}, {\em Solving roughly forced nonlinear {PDE}s via misspecified kernel methods and neural networks}, preprint arXiv:2501.17110,  (2025).

\bibitem{batlle2024kernel}
{\sc P.~Batlle, M.~Darcy, B.~Hosseini, and H.~Owhadi}, {\em Kernel methods are competitive for operator learning}, Journal of Computational Physics, 496 (2024), p.~112549.

\bibitem{beaglehole2024average}
{\sc D.~Beaglehole, P.~S{\'u}ken{\'\i}k, M.~Mondelli, and M.~Belkin}, {\em Average gradient outer product as a mechanism for deep neural collapse}, in Advances in Neural Information Processing Systems, vol.~37, 2024, pp.~130764--130796.

\bibitem{brunton2016discovering}
{\sc S.~L. Brunton, J.~L. Proctor, and J.~N. Kutz}, {\em Discovering governing equations from data by sparse identification of nonlinear dynamical systems}, Proceedings of the national academy of sciences, 113 (2016), pp.~3932--3937.

\bibitem{calatroni2017bilevel}
{\sc L.~Calatroni, C.~Cao, J.~C. De~Los~Reyes, C.-B. Sch{\"o}nlieb, and T.~Valkonen}, {\em Bilevel approaches for learning of variational imaging models}, Variational methods: In imaging and geometric control, 18 (2017), p.~2.

\bibitem{carrillo2024mean}
{\sc J.~A. Carrillo, F.~Hoffmann, A.~M. Stuart, and U.~Vaes}, {\em The mean-field ensemble kalman filter: Near-gaussian setting}, SIAM Journal on Numerical Analysis, 62 (2024), pp.~2549--2587.

\bibitem{chen2021solving}
{\sc Y.~Chen, B.~Hosseini, H.~Owhadi, and A.~M. Stuart}, {\em Solving and learning nonlinear {PDE}s with {G}aussian processes}, Journal of Computational Physics, 447 (2021).

\bibitem{chen2025gaussian}
{\sc Y.~Chen, B.~Hosseini, H.~Owhadi, and A.~M. Stuart}, {\em {Gaussian measures conditioned on nonlinear observations: consistency, MAP estimators, and simulation}}, Statistics and Computing, 35 (2025), p.~10.

\bibitem{chen2025sparse}
{\sc Y.~Chen, H.~Owhadi, and F.~Sch{\"a}fer}, {\em Sparse {C}holesky factorization for solving nonlinear {PDE}s via {G}aussian processes}, Mathematics of Computation, 94 (2025), pp.~1235--1280.

\bibitem{chen2021consistency}
{\sc Y.~Chen, H.~Owhadi, and A.~M. Stuart}, {\em {Consistency of empirical Bayes and kernel flow for hierarchical parameter estimation}}, Mathematics of Computation, 90 (2021), pp.~2527--2578.

\bibitem{cleary2021calibrate}
{\sc E.~Cleary, A.~Garbuno-Inigo, S.~Lan, T.~Schneider, and A.~M. Stuart}, {\em Calibrate, emulate, sample}, Journal of Computational Physics, 424 (2021).

\bibitem{constantine2015active}
{\sc P.~G. Constantine}, {\em Active subspaces: Emerging ideas for dimension reduction in parameter studies}, SIAM, 2015.

\bibitem{cyr2020robust}
{\sc E.~C. Cyr, M.~A. Gulian, R.~G. Patel, M.~Perego, and N.~A. Trask}, {\em Robust training and initialization of deep neural networks: An adaptive basis viewpoint}, in Mathematical and Scientific Machine Learning, PMLR, 2020, pp.~512--536.

\bibitem{de2017bilevel}
{\sc J.~C. De~los Reyes, C.~Sch{\"o}nlieb, and T.~Valkonen}, {\em Bilevel parameter learning for higher-order total variation regularisation models}, Journal of Mathematical Imaging and Vision, 57 (2017), pp.~1--25.

\bibitem{dunbar2025hyperparameter}
{\sc O.~R. Dunbar, N.~H. Nelsen, and M.~Mutic}, {\em Hyperparameter optimization for randomized algorithms: A case study on random features}, Statistics and Computing, 35 (2025), pp.~1--28.

\bibitem{engl2015regularization}
{\sc H.~W. Engl and R.~Ramlau}, {\em Regularization of inverse problems}, in Encyclopedia of applied and computational mathematics, Springer, 2015, pp.~1233--1241.

\bibitem{franceschi2018bilevel}
{\sc L.~Franceschi, P.~Frasconi, S.~Salzo, R.~Grazzi, and M.~Pontil}, {\em Bilevel programming for hyperparameter optimization and meta-learning}, in International conference on machine learning, PMLR, 2018, pp.~1568--1577.

\bibitem{frazier2018tutorial}
{\sc P.~I. Frazier}, {\em A tutorial on {B}ayesian optimization}, preprint arXiv:1807.02811,  (2018).

\bibitem{ge2023maximum}
{\sc J.~Ge, S.~Tang, J.~Fan, C.~Ma, and C.~Jin}, {\em Maximum likelihood estimation is all you need for well-specified covariate shift}, preprint arXiv:2311.15961,  (2023).

\bibitem{gibbs1998bayesian}
{\sc M.~N. Gibbs}, {\em Bayesian {G}aussian processes for regression and classification}, PhD thesis, University of Cambridge, 1998.

\bibitem{hamzi2021learning}
{\sc B.~Hamzi and H.~Owhadi}, {\em Learning dynamical systems from data: a simple cross-validation perspective, part {I}: parametric kernel flows}, Physica D: Nonlinear Phenomena, 421 (2021), p.~132817.

\bibitem{jalalian2025data}
{\sc Y.~Jalalian, J.~F.~O. Ramirez, A.~Hsu, B.~Hosseini, and H.~Owhadi}, {\em Data-efficient kernel methods for learning differential equations and their solution operators: Algorithms and error analysis}, preprint arXiv:2503.01036,  (2025).

\bibitem{jones1998efficient}
{\sc D.~R. Jones, M.~Schonlau, and W.~J. Welch}, {\em Efficient global optimization of expensive black-box functions}, Journal of Global optimization, 13 (1998), pp.~455--492.

\bibitem{jorgensen2025bayesian}
{\sc F.~J.~N. Jorgensen and Y.~M. Marzouk}, {\em A {Bayesian} characterization of ensemble {Kalman} updates}, preprint arXiv:2510.00158,  (2025).

\bibitem{kaipio2005statistical}
{\sc J.~P. Kaipio and E.~Somersalo}, {\em Statistical and computational inverse problems}, Springer, 2005.

\bibitem{kingma2014adam}
{\sc D.~P. Kingma and J.~Ba}, {\em Adam: A method for stochastic optimization}, preprint arXiv:1412.6980,  (2014).

\bibitem{kovachki2023neural}
{\sc N.~Kovachki, Z.~Li, B.~Liu, K.~Azizzadenesheli, K.~Bhattacharya, A.~Stuart, and A.~Anandkumar}, {\em Neural operator: Learning maps between function spaces with applications to pdes}, Journal of Machine Learning Research, 24 (2023), pp.~1--97.

\bibitem{li2020fourier}
{\sc Z.~Li, N.~Kovachki, K.~Azizzadenesheli, B.~Liu, K.~Bhattacharya, A.~Stuart, and A.~Anandkumar}, {\em Fourier neural operator for parametric partial differential equations}, preprint arXiv:2010.08895,  (2020).

\bibitem{lorraine2020optimizing}
{\sc J.~Lorraine, P.~Vicol, and D.~Duvenaud}, {\em Optimizing millions of hyperparameters by implicit differentiation}, in International conference on artificial intelligence and statistics, PMLR, 2020, pp.~1540--1552.

\bibitem{lu2021learning}
{\sc L.~Lu, P.~Jin, G.~Pang, Z.~Zhang, and G.~E. Karniadakis}, {\em {Learning nonlinear operators via DeepONet based on the universal approximation theorem of operators}}, Nature Machine Intelligence, 3 (2021), pp.~218--229.

\bibitem{maclaurin2015gradient}
{\sc D.~Maclaurin, D.~Duvenaud, and R.~Adams}, {\em Gradient-based hyperparameter optimization through reversible learning}, in International conference on machine learning, PMLR, 2015, pp.~2113--2122.

\bibitem{meng2023sparse}
{\sc R.~Meng and X.~Yang}, {\em Sparse {G}aussian processes for solving nonlinear {PDE}s}, Journal of Computational Physics, 490 (2023), p.~112340.

\bibitem{movckus1974bayesian}
{\sc J.~Mo{\v{c}}kus}, {\em On {B}ayesian methods for seeking the extremum}, in IFIP Technical Conference on Optimization Techniques, Springer, 1974, pp.~400--404.

\bibitem{mora2025gaussian}
{\sc C.~Mora, A.~Yousefpour, S.~Hosseinmardi, and R.~Bostanabad}, {\em A {G}aussian process framework for solving forward and inverse problems involving nonlinear partial differential equations}, Computational Mechanics, 75 (2025), pp.~1213--1239.

\bibitem{naslidnyk2025comparing}
{\sc M.~Naslidnyk, M.~Kanagawa, T.~Karvonen, and M.~Mahsereci}, {\em Comparing scale parameter estimators for {G}aussian process interpolation with the brownian motion prior: Leave-one-out cross validation and maximum likelihood}, SIAM/ASA Journal on Uncertainty Quantification, 13 (2025), pp.~679--717.

\bibitem{nelsen2024operator}
{\sc N.~H. Nelsen and A.~M. Stuart}, {\em Operator learning using random features: A tool for scientific computing}, SIAM Review, 66 (2024), pp.~535--571.

\bibitem{owhadi2022computational}
{\sc H.~Owhadi}, {\em Computational graph completion}, Research in the Mathematical Sciences, 9 (2022), p.~27.

\bibitem{owhadi2023ideas}
\leavevmode\vrule height 2pt depth -1.6pt width 23pt, {\em {Do ideas have shape? Idea registration as the continuous limit of artificial neural networks}}, Physica D: Nonlinear Phenomena, 444 (2023).

\bibitem{owhadi2019operator}
{\sc H.~Owhadi and C.~Scovel}, {\em Operator-Adapted Wavelets, Fast Solvers, and Numerical Homogenization: From a Game Theoretic Approach to Numerical Approximation and Algorithm Design}, vol.~35, Cambridge University Press, 2019.

\bibitem{owhadi2019kernel}
{\sc H.~Owhadi and G.~R. Yoo}, {\em Kernel flows: From learning kernels from data into the abyss}, Journal of Computational Physics, 389 (2019), pp.~22--47.

\bibitem{pedregosa2016hyperparameter}
{\sc F.~Pedregosa}, {\em Hyperparameter optimization with approximate gradient}, in International conference on machine learning, PMLR, 2016, pp.~737--746.

\bibitem{radhakrishnan2024mechanism}
{\sc A.~Radhakrishnan, D.~Beaglehole, P.~Pandit, and M.~Belkin}, {\em Mechanism for feature learning in neural networks and backpropagation-free machine learning models}, Science, 383 (2024), pp.~1461--1467.

\bibitem{raissi2017machine}
{\sc M.~Raissi, P.~Perdikaris, and G.~E. Karniadakis}, {\em Machine learning of linear differential equations using {G}aussian processes}, Journal of Computational Physics, 348 (2017), pp.~683--693.

\bibitem{raissi2019physics}
\leavevmode\vrule height 2pt depth -1.6pt width 23pt, {\em Physics-informed neural networks: A deep learning framework for solving forward and inverse problems involving nonlinear partial differential equations}, Journal of Computational physics, 378 (2019), pp.~686--707.

\bibitem{rudy2017data}
{\sc S.~H. Rudy, S.~L. Brunton, J.~L. Proctor, and J.~N. Kutz}, {\em Data-driven discovery of partial differential equations}, Science advances, 3 (2017), p.~e1602614.

\bibitem{shaban2019truncated}
{\sc A.~Shaban, C.~A. Cheng, N.~Hatch, and B.~Boots}, {\em Truncated back-propagation for bilevel optimization}, in The 22nd international conference on artificial intelligence and statistics, PMLR, 2019, pp.~1723--1732.

\bibitem{shahriari2015taking}
{\sc B.~Shahriari, K.~Swersky, Z.~Wang, R.~P. Adams, and N.~De~Freitas}, {\em Taking the human out of the loop: A review of {B}ayesian optimization}, Proceedings of the IEEE, 104 (2015), pp.~148--175.

\bibitem{shao2025solving}
{\sc Z.~Shao, K.~Pieper, and X.~Tian}, {\em Solving nonlinear {PDEs} with sparse radial basis function networks}, preprint arXiv:2505.07765,  (2025).

\bibitem{stein1999interpolation}
{\sc M.~L. Stein}, {\em Interpolation of spatial data: some theory for kriging}, Springer Science \& Business Media, 1999.

\bibitem{stuart2010inverse}
{\sc A.~M. Stuart}, {\em {Inverse problems: A Bayesian perspective}}, Acta Numerica, 19 (2010), pp.~451--559.

\bibitem{tarantola2005inverse}
{\sc A.~Tarantola}, {\em Inverse problem theory and methods for model parameter estimation}, SIAM, 2005.

\bibitem{wang2022and}
{\sc S.~Wang, X.~Yu, and P.~Perdikaris}, {\em {When and why PINNs fail to train: A neural tangent kernel perspective}}, Journal of Computational Physics, 449 (2022), p.~110768.

\bibitem{wang2024pinn}
{\sc Y.~Wang and L.~Zhong}, {\em {NAS-PINN: Neural architecture search-guided physics-informed neural network for solving PDEs}}, Journal of Computational Physics, 496 (2024), p.~112603.

\bibitem{white1982maximum}
{\sc H.~White}, {\em Maximum likelihood estimation of misspecified models}, Econometrica: Journal of the econometric society,  (1982), pp.~1--25.

\bibitem{rasmussen2006gaussian}
{\sc C.~K. Williams and C.~E. Rasmussen}, {\em Gaussian processes for machine learning}, MIT press Cambridge, MA.,  (2006).

\bibitem{wilson2016deep}
{\sc A.~G. Wilson, Z.~Hu, R.~Salakhutdinov, and E.~P. Xing}, {\em Deep kernel learning}, in Artificial intelligence and statistics, PMLR, 2016, pp.~370--378.

\bibitem{zou2025multi}
{\sc Z.~Zou and G.~E. Karniadakis}, {\em Multi-head physics-informed neural networks for learning functional priors and uncertainty quantification}, Journal of Computational Physics, 531 (2025), p.~113947.

\bibitem{zou2024correcting}
{\sc Z.~Zou, X.~Meng, and G.~E. Karniadakis}, {\em {Correcting model misspecification in physics-informed neural networks (PINNs)}}, Journal of Computational Physics, 505 (2024), p.~112918.

\bibitem{zou2025uncertainty}
{\sc Z.~Zou, X.~Meng, and G.~E. Karniadakis}, {\em Uncertainty quantification for noisy inputs--outputs in physics-informed neural networks and neural operators}, Computer Methods in Applied Mechanics and Engineering, 433 (2025).

\bibitem{zou2024neuraluq}
{\sc Z.~Zou, X.~Meng, A.~F. Psaros, and G.~E. Karniadakis}, {\em {NeuralUQ: A comprehensive library for uncertainty quantification in neural differential equations and operators}}, SIAM Review, 66 (2024), pp.~161--190.

\bibitem{zou2025learning}
{\sc Z.~Zou, Z.~Wang, and G.~E. Karniadakis}, {\em Learning and discovering multiple solutions using physics-informed neural networks with random initialization and deep ensemble}, preprint arXiv:2503.06320,  (2025).

\end{thebibliography}

\end{document}